\pdfoutput=1

\documentclass[11pt]{article}

\usepackage[]{acl}

\usepackage{times}
\usepackage{latexsym}

\usepackage[T1]{fontenc}

\usepackage[utf8]{inputenc}

\usepackage{microtype}

\usepackage{inconsolata}

\usepackage{multirow}
\usepackage{graphicx}
\usepackage{caption}
\usepackage{subcaption}
\usepackage{amsmath}
\usepackage{booktabs}       

\usepackage{comment}
\usepackage{amsmath}
\usepackage{amssymb}
\usepackage{url}
\usepackage{hyperref}
\usepackage{enumitem}
\usepackage{amsmath}

\title{DA$^3$: A Distribution-Aware Adversarial Attack against Language Models}

\author{Yibo Wang$^1$\thanks{\hspace{0.2cm}Equal Contribution}\quad Xiangjue Dong$^2$\footnotemark[1]\quad James Caverlee$^2$\quad Philip S. Yu$^1$\\
$^1$ University of Illinois Chicago, $^2$ Texas A\&M University \\ \small\texttt{\{ywang633, psyu\}@uic.edu, \{xj.dong, caverlee\}@tamu.edu}}

\begin{document}
\maketitle
\begin{abstract}
Language models can be manipulated by adversarial attacks, which introduce subtle perturbations to input data. While recent attack methods can achieve a relatively high attack success rate (ASR), we've observed that the generated adversarial examples have a different data distribution compared with the original examples. Specifically, these adversarial examples exhibit reduced confidence levels and greater divergence from the training data distribution. Consequently, they are easy to detect using straightforward detection methods, diminishing the efficacy of such attacks. To address this issue, we propose a Distribution-Aware Adversarial Attack (DA$^3$) method. DA$^3$ considers the distribution shifts of adversarial examples to improve attacks' effectiveness under detection methods. We further design a novel evaluation metric, the Non-detectable Attack Success Rate (NASR), which integrates both ASR and detectability for the attack task. We conduct experiments on four widely used datasets to validate the attack effectiveness and transferability of adversarial examples generated by DA$^3$ against both the white-box \textsc{BERT-base} and \textsc{RoBERTa-base} models and the black-box \textsc{LLaMA2-7b} model\footnote{Our codes are available at \url{https://github.com/YiboWANG214/DALA}.}.
\end{abstract}

\section{Introduction}
Language models (LMs), despite their remarkable accuracy and human-like capabilities in many applications, face vulnerability to adversarial attacks and exhibit high sensitivity to subtle input perturbations, which can potentially cause failures~\cite{jia-liang-2017-adversarial,belinkov2018synthetic,wallace-etal-2019-universal}. Recently, an increasing number of adversarial attacks have been proposed, employing techniques such as insertion, deletion, swapping, and substitution at character, word, or sentence levels~\cite{ren-etal-2019-generating,textfooler-2020, garg-ramakrishnan-2020-bae,ribeiro-etal-2020-beyond}. These thoroughly crafted adversarial examples are imperceptible to humans yet can deceive victim models, raising concerns regarding the robustness and security of LMs. For example, chatbots may misunderstand user intent or sentiment, resulting in inappropriate responses~\cite {perez-etal-2022-red}.

\begin{figure}[t]
    \centering
    \includegraphics[width=0.5\textwidth]{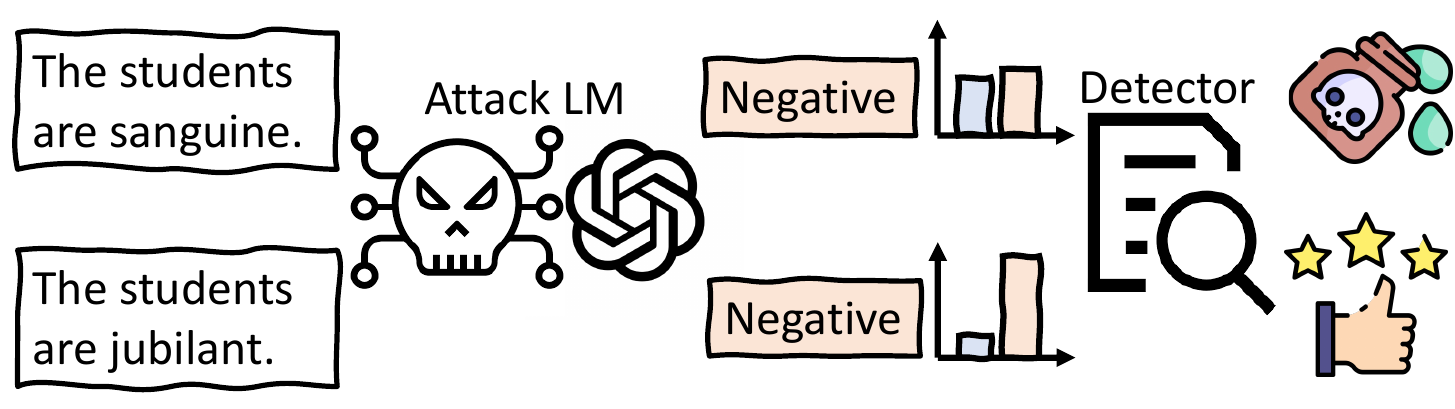}
    \caption{Toy examples of two adversarial sentences in a sentiment analysis task. Although both sentences successfully attack the victim model, the top one is flagged by the detector, while the bottom one is not detected. In our task, we aim to generate adversarial examples that are hard to detect.}
    \label{fig: intro}
\end{figure}

However, while existing adversarial attacks can achieve a relatively high attack success rate~\cite{deepwordbug-2018,belinkov2018synthetic,li-etal-2020-bert-attack}, our experimental observations detailed in \S\ref{observation expers} reveal notable distribution shifts between adversarial examples and original examples, rendering high detectability of adversarial examples.
On one hand, adversarial examples exhibit different confidence levels compared to their original counterparts. Typically, the Maximum Softmax Probability (MSP), a metric indicating prediction confidence, is higher for original examples than for adversarial examples.
On the other hand, there is a disparity in the distance to the training data distribution between adversarial and original examples. Specifically, the Mahalanobis Distance (MD) to training data distribution for original examples is shorter than that for adversarial examples.
Based on these two observations, we conclude that adversarial examples generated by previous attack methods, such as BERT-Attack~\cite{li-etal-2020-bert-attack}, can be easily detected through score-based detection techniques like MSP detection~\cite{hendrycksbaseline} and embedding-based detection methods like MD detection~\cite{lee2018simple}. Thus, the efficacy of previous attack methods is diminished when considering Out-of-distribution (OOD) detection, as shown in Figure~\ref{fig: intro}.

To address the aforementioned problems, we propose a \textbf{D}istribution-\textbf{A}ware \textbf{A}dversarial \textbf{A}ttack (DA$^3$) method with Data Alignment Loss (DAL), which is a novel attack method that can generate hard-to-detect adversarial examples. 
The DA$^3$ framework comprises two phases. Firstly, DA$^3$ finetunes a LoRA-based LM by combining the Masked Language Modeling task and the downstream classification task using DAL. This fine-tuning phase enables the LoRA-based LM to generate adversarial examples closely resembling original examples in terms of MSP and MD. Subsequently, the LoRA-based LM is used during inference to generate adversarial examples.

To measure the detectability of adversarial examples, we propose a new evaluation metric: \textbf{N}on-detectable \textbf{A}ttack \textbf{S}uccess \textbf{R}ate (NASR),
which combines Attack Success Rate (ASR) with OOD detection. We conduct experiments on four datasets to assess whether DA$^3$ can effectively attack white-box LMs using ASR and NASR. Furthermore, given the widespread use of Large Language Models (LLMs) and their costly fine-tuning process, coupled with the limited availability of open-source models, we also evaluate the attack transferability of adversarial examples on black-box LLMs. The results show that DA$^3$ achieves competitive attack performance on the white-box \textsc{BERT-base}~\cite{devlin-etal-2019-bert} and \textsc{RoBERTa-base}~\cite{liu2019roberta} models and superior transferability on the black-box \textsc{LLaMA2-7b}~\cite{touvron2023llama}. 

Our work has the following contributions:
\begin{itemize}
[noitemsep,topsep=0pt,itemsep=2pt,leftmargin=0.4cm]
    \item We analyze the distribution of adversarial and original examples, revealing the existence of distribution shifts in terms of MSP and MD.
    \item We propose a novel Distribution-Aware Adversarial Attack method with Data Alignment Loss, which is capable of generating adversarial examples that effectively undermine victim models while remaining difficult to detect.
    \item We design a new evaluation metric -- NASR -- for the attack task, which considers the detectability of adversarial examples.
    \item We conduct comprehensive experiments to compare DA$^3$ with baselines on four datasets, demonstrating that DA$^3$ achieves competitive attack capabilities and better transferability.
\end{itemize}

\section{Related Work}
\subsection{Adversarial Attacks in NLP}

Adversarial attacks have been extensively studied to explore the robustness of LMs. Current methods fall into character-level, word-level, sentence-level, and multi-level~\cite{goyal-etal-survey-2023}.
Character-level methods manipulate texts by incorporating typos or errors into words, such as deleting, repeating, replacing, swapping, flipping, inserting, and allowing variations in characters for specific words~\cite{deepwordbug-2018,belinkov2018synthetic}. 
Word-level attacks alter entire words rather than individual characters within words.
Common manipulation includes addition, deletion, and substitution with synonyms to mislead language models while the manipulated words are selected based on gradients or importance scores~\cite{ren-etal-2019-generating,textfooler-2020,li-etal-2020-bert-attack,garg-ramakrishnan-2020-bae}. Sentence-level attacks typically involve inserting or rewriting sentences within a text, all while preserving the original meaning~\cite{zhao2018generating,iyyer-etal-2018-adversarial,ribeiro-etal-2020-beyond}. Multi-level attacks combine multiple perturbation techniques to achieve both imperceptibility and a high success rate in the attack~\cite{song-etal-2021-universal}.

\subsection{Out-of-distribution Detection in NLP}
Out-of-distribution (OOD) detection methods have been widely explored in NLP, like machine translation~\cite{arora2021types, ren2022out, adila2022understanding}. OOD detection methods in NLP can be roughly categorized into two types: (1) score-based methods and (2) embedding-based methods.
Score-based methods use maximum softmax probability~\cite{hendrycksbaseline}, perplexity score~\cite{arora2021types}, beam score~\cite{wang-etal-2019-improving-back}, sequence probability~\cite{wang-etal-2019-improving-back}, BLEU variance~\cite{xiao2020wat}, or energy-based scores~\cite{liu2020energy}. Embedding-based methods measure the distance to in-distribution data in the embedding space for OOD detection. For example, \citet{lee2018simple} uses Mahalanobis distance; \citet{ren2021simple} proposes to use relative Mahalanobis distance; \citet{sun2022out} proposes a nearest-neighbor-based OOD detection method.

We select the simple, representative, and widely-used OOD detection methods of these two categories: MSP detection~\cite{hendrycksbaseline} and MD detection~\cite{lee2018simple}, respectively. This selection serves to highlight a significant issue within the community -- the ability to detect adversarial examples using such basic and commonly employed OOD detection methods underscores the criticality of detectability.
These two methods are then incorporated with the ASR to assess the robustness and detectability of adversarial examples generated by different attack models.

\begin{figure}[t]
     \centering
     \begin{subfigure}[t]{0.48\columnwidth}
         \centering
         \includegraphics[width=\textwidth]{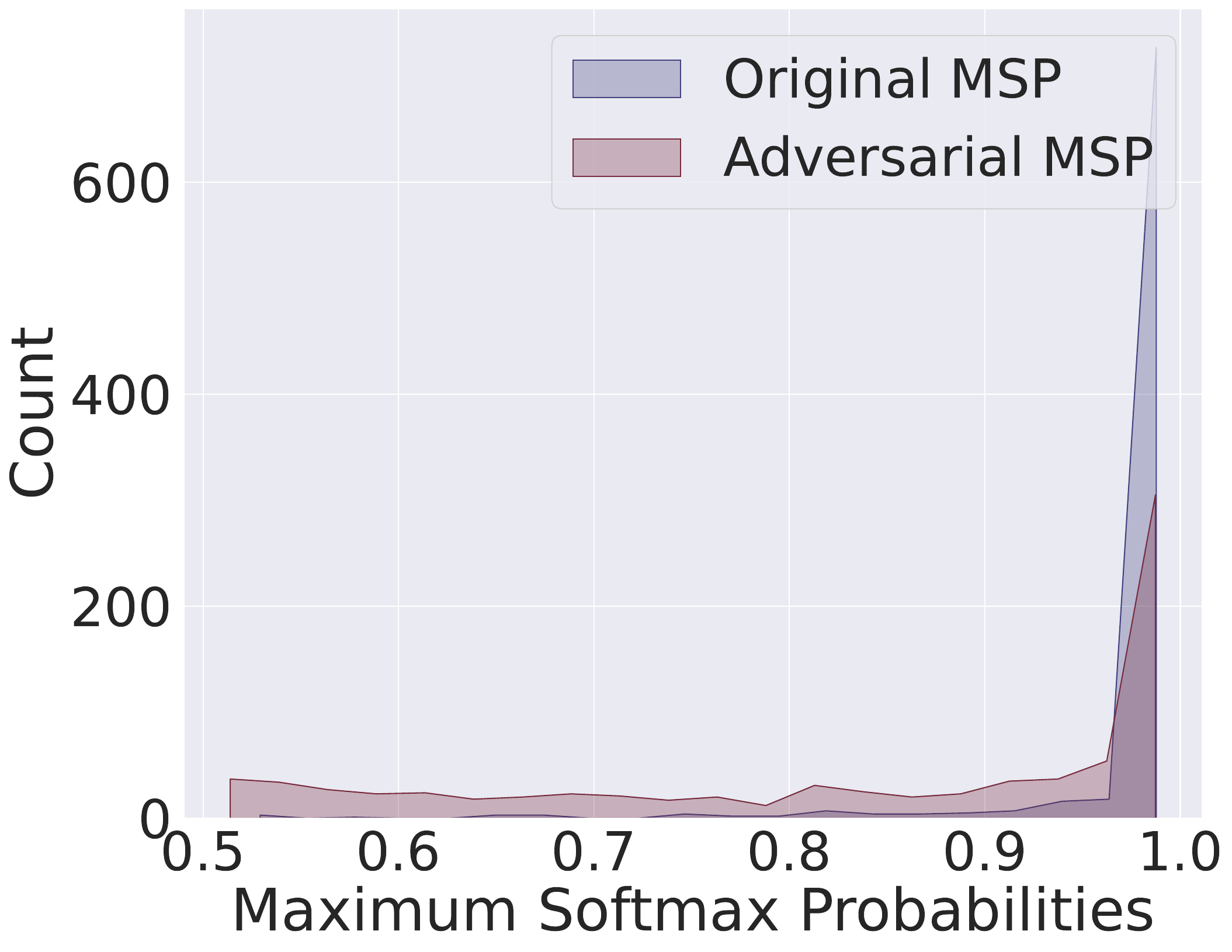}
         \caption{MSP on SST-2 dataset.}
     \end{subfigure}    
     \hfill
     \begin{subfigure}[t]{0.48\columnwidth}
         \centering
         \includegraphics[width=\textwidth]{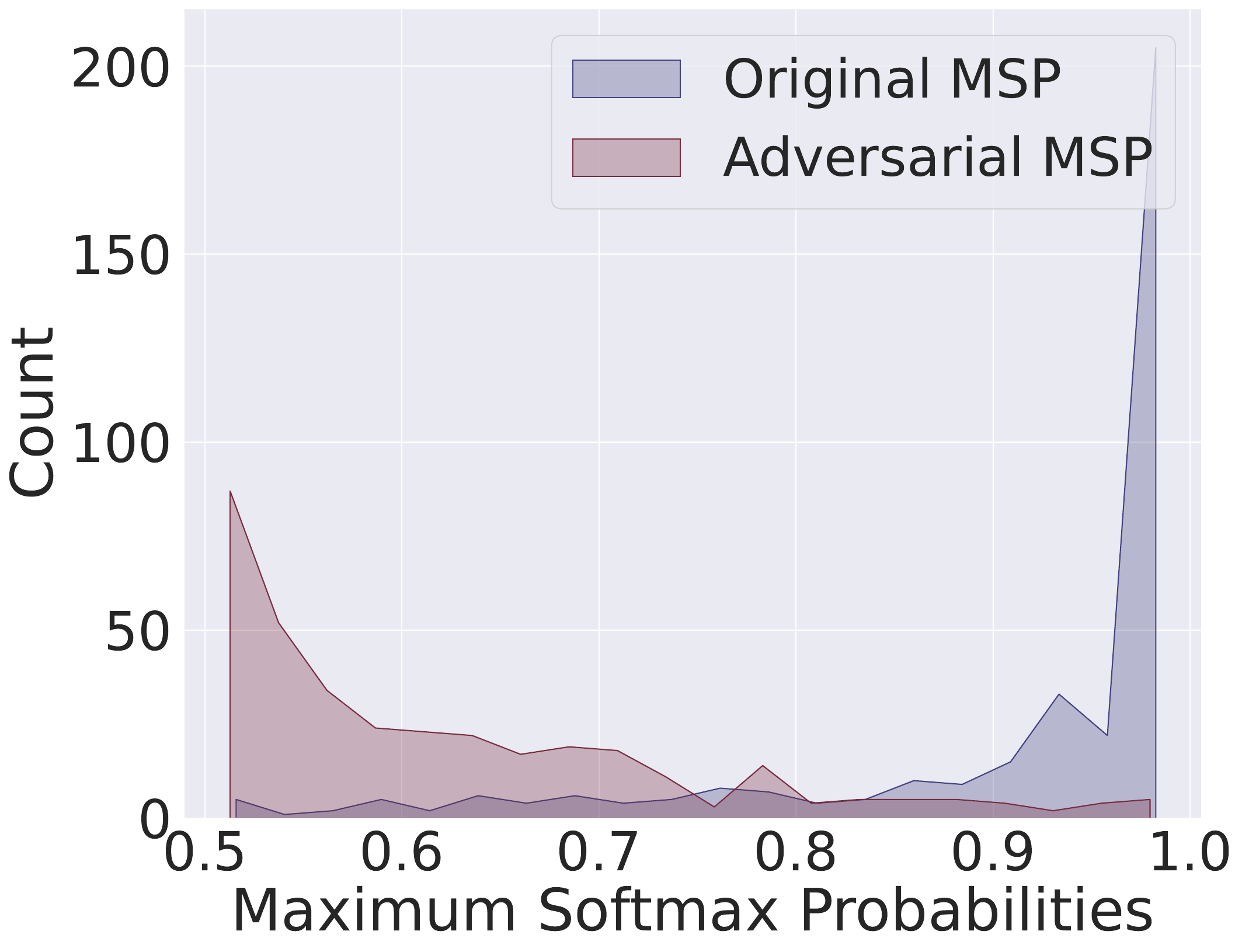}
         \caption{MSP on MRPC dataset.}
     \end{subfigure}    
    \caption{Visualization of the distribution shift between original data and adversarial data generated by BERT-Attack when attacking \textsc{BERT-base} regarding MSP.}
    \label{fig: observe_MSP}
\end{figure}

\begin{figure}[t]
     \centering
     \begin{subfigure}[t]{0.48\columnwidth}
         \centering
         \includegraphics[width=\textwidth]{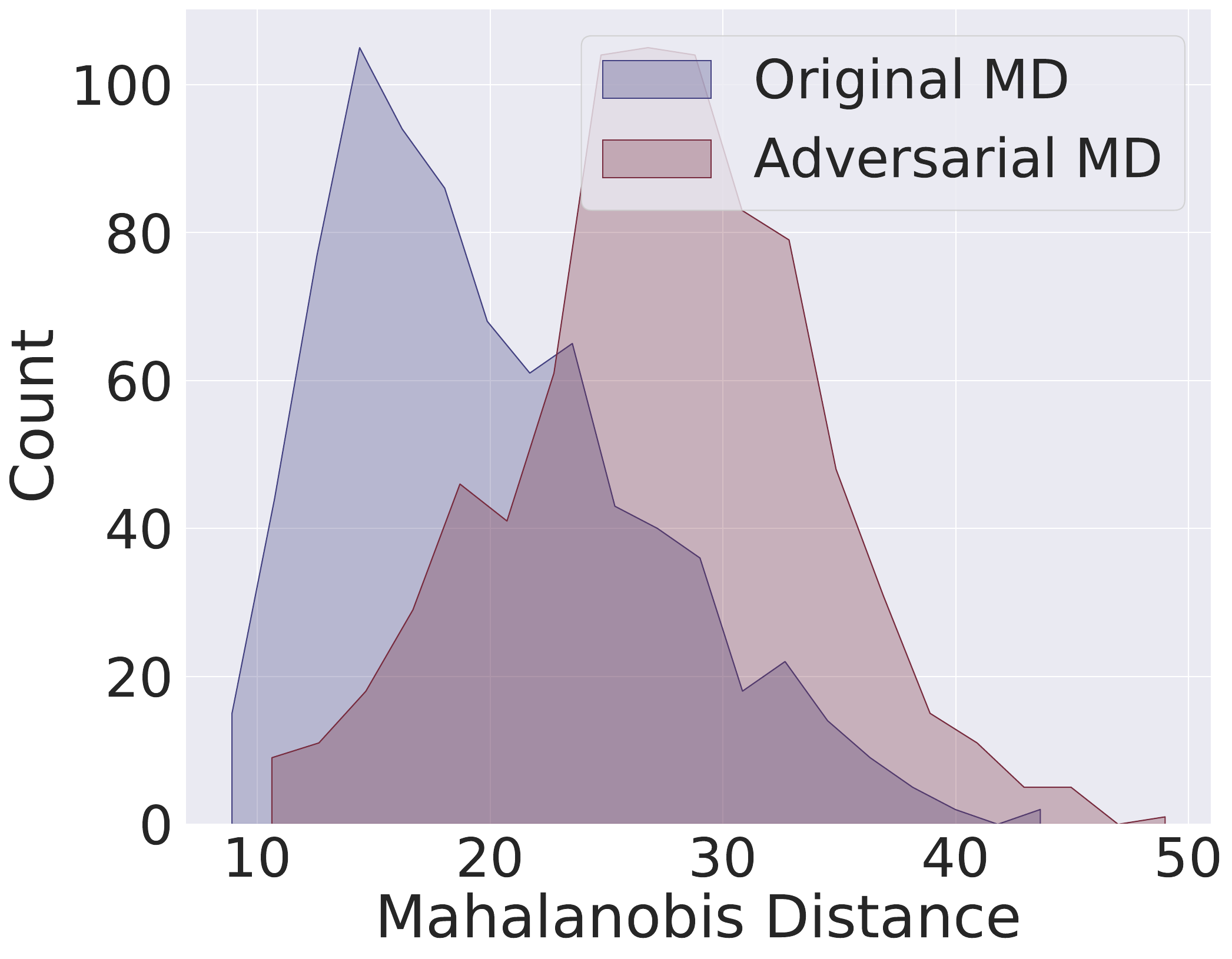}
         \caption{MD on SST-2 dataset.}
     \end{subfigure}
     \hfill
     \begin{subfigure}[t]{0.48\columnwidth}
         \centering
         \includegraphics[width=\textwidth]{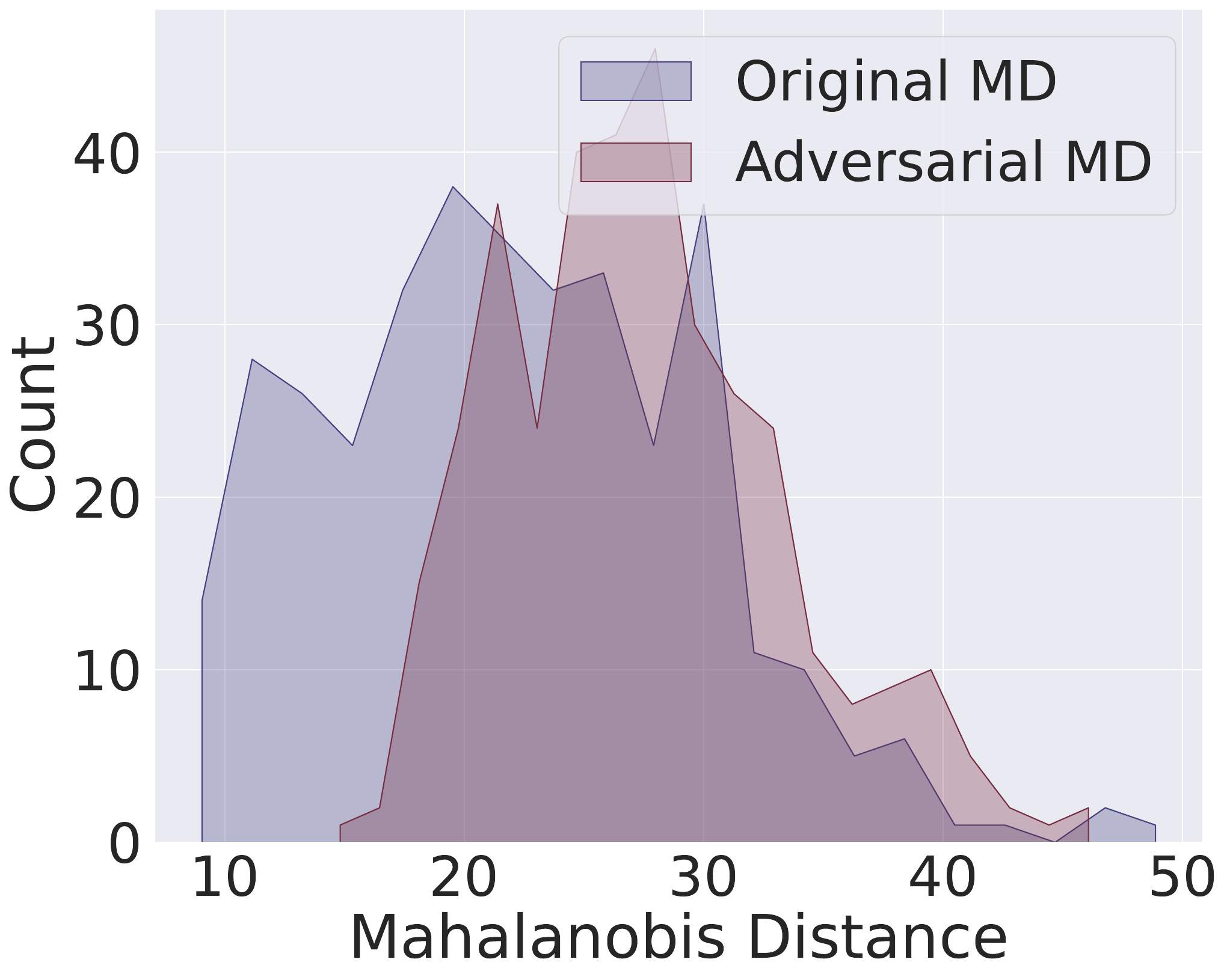}
         \caption{MD on MRPC dataset.}
     \end{subfigure}
    \caption{Visualization of the distribution shift between original data and adversarial data generated by BERT-Attack when attacking \textsc{BERT-base} regarding MD.}
    \label{fig: observe_MD}
\end{figure}

\section{Understanding Distribution Shifts of Adversarial Examples}
\label{observation expers}
This section showcases distribution shifts between adversarial and original examples, suggesting that the original examples are in-distribution examples while adversarial examples are Out-of-Distribution (OOD) examples.
Due to space constraints, we focus our analysis on adversarial examples generated by BERT-Attack on SST-2~\cite{socher-etal-2013-recursive} and MRPC~\cite{dolan-brockett-2005-automatically}; the complete results are available in Appendix~\ref{observation}.

\paragraph{Maximum Softmax Probability (MSP).}
Maximum Softmax Probability (MSP) is a metric to evaluate prediction confidence, rendering it a widely used score-based method for OOD detection, where lower confidence values often signify OOD examples.
To assess MSP, we visualize the MSP distribution of adversarial examples generated by BERT-Attack and original examples from SST-2 and MRPC datasets in Figure~\ref{fig: observe_MSP}. Our observation reveals that in both datasets, the majority of original examples have an MSP exceeding 0.9, indicating a significantly higher MSP compared to adversarial examples overall. This distribution shift is particularly notable in the MRPC dataset, whereby most adversarial examples exhibit MSP below 0.6, highlighting a clear distinction from the original examples.

\paragraph{Mahalanobis Distance (MD).}
Mahalanobis Distance (MD) is a metric used to measure the distance between a data point and a distribution, making it a highly suitable and widespread method for OOD detection. 
A high MD between an example and the in-distribution data (training data) indicates that the example is probably an OOD instance.
To assess the MD difference between adversarial and original examples, we visualize the MD distribution of adversarial examples generated by BERT-Attack and original examples from the SST-2 and MRPC datasets in Figure~\ref{fig: observe_MD}. 
From Figure~\ref{fig: observe_MD}, we can observe that distribution shifts exist between original and adversarial examples in both datasets. This dissimilarity is more noticeable on the SST-2 dataset and not as conspicuous on the MRPC dataset.

\paragraph{Summary.}
These observations regarding MSP and MD highlight clear distinctions between original and adversarial examples generated by one of the state-of-the-art methods, BERT-Attack.
Compared to the original examples, the adversarial examples exhibit a more pronounced OOD nature in either MSP or MD, meaning that adversarial examples are easy to detect and the practical effectiveness of previous attack methods is diminished.

\begin{figure*}[t]
     \centering
     \includegraphics[width=0.9\linewidth]{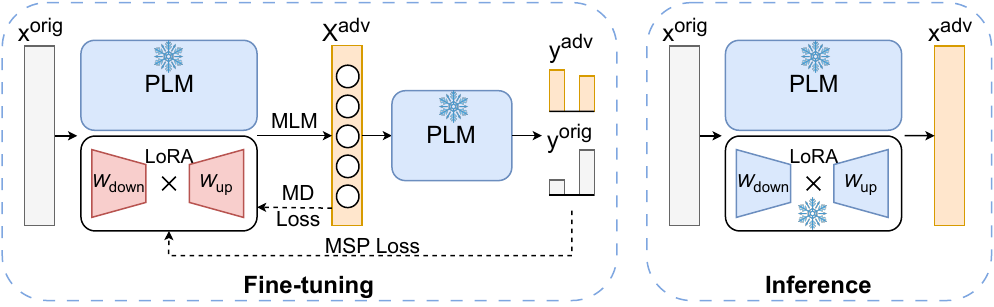}
    \caption{The model architecture of DA$^3$ comprises two phases: fine-tuning and inference. During fine-tuning, a LoRA-based PLM is fine-tuned to develop the ability to generate adversarial examples resembling original examples in terms of MSP and MD. During inference, the LoRA-based PLM is used to generate adversarial examples.
    }
    \label{fig: model}
\end{figure*}

\section{Methodology}
In this section, we define the attack task (\S\ref{define}), propose a novel attack method called Distribution-Aware Adversarial Attack (\S\ref{model}), and introduce the new Data Alignment Loss (\S\ref{loss}).

\subsection{Problem Formulation}
\label{define}
Given an original sentence $x^{orig}\in \mathcal{X}$ and its corresponding original label $y^{orig}\in \mathcal{Y}$, our objective is to generate an adversarial sentence $x^{adv}$ such that the prediction of the victim model corresponds to $y^{adv}\in\mathcal{Y}$ and $y^{adv}\neq y^{orig}$.

\subsection{Distribution-Aware Adversarial Attack}
\label{model}

Motivated by the observed distribution shifts of adversarial examples, we propose a Distribution-Aware Adversarial Attack (DA$^3$) method.
The key idea of DA$^3$ is to consider the distribution of the generated adversarial examples and attempt to achieve a closer alignment between distributions of adversarial and original examples in terms of MSP and MD.
DA$^3$ is composed of two phases: fine-tuning and inference, as shown in Figure~\ref{fig: model}. 

\paragraph{Fine-tuning Phase.}
The fine-tuning phase aims to fine-tune a LoRA-based Pre-trained Language Model (PLM) to make it capable of generating adversarial examples through the Masked Language Modeling (MLM) task. We employ LoRA-based PLM because it is efficient to finetune and the frozen PLM can serve in both MLM and downstream classification tasks.
First, the original sentence $x^{orig}$ undergoes the MLM task through a LoRA-based PLM to generate the adversarial embedding $X^{adv}$, during which the parameters of the PLM are frozen, and the parameters of \textsc{LoRA}~\cite{hu2021lora} are tunable.
Then, the generated adversarial embedding $X^{adv}$ is fed into the frozen PLM to perform the corresponding downstream classification task, producing logits of original ground truth label $y^{orig}$ and adversarial label $y^{adv}$.
The loss is computed based on $X^{adv}$, $P(y^{orig}|X^{adv}, \theta)$, and $P(y^{adv}|X^{adv}, \theta)$ to update the parameters of \textsc{LoRA}, where $\theta$ is the model parameters. Details are discussed in \S\ref{loss}.

\paragraph{Inference Phase.}
The inference phase aims to generate adversarial examples with minimal perturbation.
The original sentence $x^{orig}$ is first tokenized, and a ranked token list is obtained through token importance~\cite{li-etal-2020-bert-attack}. 
Then, a token is selected from the token list to be masked. Subsequently, the MLM task of the frozen LoRA-based PLM is employed to generate a candidate list for the masked token. A word is then chosen from the list to replace the masked token until a successful attack on the victim model is achieved or the candidate list is exhausted. 
If the attack is unsuccessful, another token is chosen from the token list until a successful attack is achieved or the termination condition is met. The termination condition is set as the percentage of the tokens.

\subsection{Model Learning}
\label{loss}
The Data Alignment Loss, denoted as $\mathcal{L}_{DAL}$, is used to minimize the discrepancy between distributions of adversarial examples and original examples in terms of MSP and MD. $\mathcal{L}_{DAL}$ is composed of two losses: $\mathcal{L}_{MSP}$ and $\mathcal{L}_{MD}$.

$\mathcal{L}_{MSP}$ aims to increase the difference between $P(y^{adv}|X^{adv}, \theta)$ and $P(y^{orig}|X^{adv}, \theta)$.
$\mathcal{L}_{MSP}$ is formulated as
\begin{equation}
    \resizebox{\hsize}{!}{$\mathcal{L}_{MSP} = \sum\limits_{X^{adv}} \frac{exp(P(y^{orig}|X^{adv}, \theta))}{exp(P(y^{orig}|X^{adv}, \theta))+exp(P(y^{adv}|X^{adv}, \theta))}$}.
    \notag
\end{equation}
According to our observation experiments in Figure~\ref{fig: observe_MSP}, original examples have higher MSP than adversarial examples. Minimizing $\mathcal{L}_{MSP}$ increases MSP of adversarial examples.
Thus, minimizing $\mathcal{L}_{MSP}$ makes generated adversarial examples more similar to original examples concerning MSP.

$\mathcal{L}_{MD}$ aims to reduce MD between adversarial input and the training data distribution.
$\mathcal{L}_{MD}$ is formulated as:
\begin{equation}
    \resizebox{\hsize}{!}{$\mathcal{L}_{MD} = \sum\limits_{X^{adv}} log\sqrt{(X^{adv}-\mu)\sum\nolimits^{-1}(X^{adv}-\mu)^\intercal},
    $}
    \notag
\end{equation}
where $\mu$ and $\sum^{-1}$ are the mean and covariance embedding of the in-distribution (training) data respectively.
MD is a robust metric for OOD detection and adversarial data detection. In general, adversarial data has higher MD than original data, as shown in Figure~\ref{fig: observe_MD}. Therefore, minimizing $\mathcal{L}_{MD}$ encourages the generated adversarial examples to resemble original examples in terms of MD. $\mathcal{L}_{MD}$ is constrained to the logarithmic space for consistency with the scale of $\mathcal{L}_{MSP}$.

Thus, Data Alignment Loss is represented as
\begin{equation}
    \mathcal{L}_{DAL} = \mathcal{L}_{MSP} + \mathcal{L}_{MD},
\end{equation}
and DA$^3$ is trained by optimizing $\mathcal{L}_{DAL}$.

\section{Automatic Evaluation Metrics}

Given the observations of distribution shifts analyzed in Section~\ref{observation expers}, we adopt a widely-used metric -- Attack Success Rate (ASR) -- and design a new metric -- Non-detectable Attack Success Rate (NASR) -- to evaluate attack performance.
We also report the Percentage of Perturbed Words (\%Words) and Semantic Similarity (SS) to evaluate the impact of text perturbation.
Detailed explanations of ASR, \%Words, and SS are shown in Appendix~\ref{ssec:metrics}.

\paragraph{Non-detectable Attack Success Rate (NASR).}
Considering the detectability of adversarial examples generated by attack methods, we define a new evaluation metric -- Non-Detectable Attack Success Rate (NASR). This metric considers both ASR and OOD detection. 
Specifically, NASR posits that a successful adversarial example is characterized by its ability to deceive the victim model while simultaneously evading OOD detection methods.

We utilize two established and commonly employed OOD detection techniques -- MSP detection~\cite{hendrycksbaseline} and MD detection~\cite{lee2018simple}. MSP detection relies on logits and utilizes a probability distribution-based approach, while MD detection is a distance-based approach. For MSP detection, we use Negative MSPs, calculated as $-\max\limits_{y\in \mathcal{Y}}P(y\mid X, \theta)$. For MD detection, we compute $\sqrt{(X-\mu)\sum^{-1}(X-\mu)^\intercal}$.
NASRs under MSP detection and MD detection are denoted as $\text{\textbf{NASR}}_{MSP}$ and $\text{\textbf{NASR}}_{MD}$.

Thus, NASR is formulated as:
\begin{equation}
    \resizebox{\hsize}{!}{$\text{NASR}_{k} = 1-\frac{|\{x^{orig}\mid y^{adv} = y^{orig}, x^{orig}\in\mathcal{X}\}|+|\mathcal{D}_k|}{|\mathcal{X}|}$},
    \notag
\end{equation}
where $\mathcal{D}_k$ denotes the set of examples that successfully attack the victim model but are detected by the detection method $k\in\{MSP, MD\}$. 

In this context, adversarial examples are considered as OOD examples (positive), while original examples are considered as in-distribution examples (negative).
To avoid misdetecting original examples as adversarial examples from a defender's view, we use the negative MSP and MD value at 99\% False Positive Rate of the training data as thresholds. Values exceeding these thresholds are considered positive, while those falling below are classified as negative.

\section{Experimental Settings}


\begin{table*}[t]
    \centering
    \caption{Evaluation results on the white-box victim models. \textsc{BERT-base} and \textsc{RoBERTa-base} models are finetuned on the corresponding dataset. ACC represents model accuracy. We highlight the \textbf{best} and the \fbox{second-best} results.}
    \resizebox{\linewidth}{!}{
    \begin{tabular}{c|c|cccc|cccc}
    \hline
    \multicolumn{1}{c|}{\multirow{2}{*}{\textbf{Dataset}}} & \multicolumn{1}{c|}{\multirow{2}{*}{\textbf{Model}}} & \multicolumn{4}{c|}{\textbf{\textsc{BERT-base}}} & \multicolumn{4}{c}{\textbf{\textsc{RoBERTa-base}}} \\
    & & ACC$\downarrow$ & ASR$\uparrow$ & $\text{NASR}_{MSP}$ $\uparrow$ & $\text{NASR}_{MD}$ $\uparrow$ & ACC$\downarrow$ & ASR$\uparrow$ & $\text{NASR}_{MSP}$ $\uparrow$ & $\text{NASR}_{MD}$ $\uparrow$ \\
    \hline
    \multirow{7}{*}{SST-2} 
    & Original & 92.43 &&&& 94.04  \\
    & TextFooler & 4.47 & 95.16 & 53.47 & \textbf{91.94} & 4.7 & 95.0 & 73.29 & 92.93 \\
    & TextBugger & 29.01 & 68.61 & 37.34 & 66.87 & 36.70 & 60.98 & 44.02 & 60.37\\
    & DeepWordBug & 16.74 & 81.89 & \textbf{57.57} & \fbox{80.77} & 16.97 & 81.95 & 68.17 & 81.10 \\
    & BERT-Attack & 38.42 & 58.44 & 33.62 & 54.96 & 2.06 & 97.80 & \fbox{74.02} & \textbf{94.76} \\
    & A2T & 55.16 & 40.32 & 20.72 & 11.79 & 59.63 & 36.59 & 26.10 & 35.73 \\
    & DA$^3$ (ours)& 21.10 & 77.17 & \fbox{54.22} & 75.06 & 4.82 & 94.88 & \textbf{75.98} & \fbox{94.27} \\
    \hline
    \multirow{7}{*}{CoLA} 
    & Original & 81.21 &&&& 85.04 \\
    & TextFooler & 1.92 & 97.64 & \textbf{95.63} & \textbf{94.92} & 5.56 & 93.46 & \fbox{90.98} & \fbox{89.18}\\
    & TextBugger & 12.18 & 85.01 & 81.23 & 77.69 & 15.63 & 81.62 & 75.87 & 73.28\\
    & DeepWordBug & 7.09 & 91.26 & 88.78 & 86.19 & 11.02 & 87.03 & 84.10 & 74.18\\
    & BERT-Attack & 12.46 & 84.65 & 79.22 & 79.93 & 2.21 & 97.41 & \textbf{91.43} & \textbf{90.98}\\
    & A2T & 20.44 & 74.82 & 71.63 & 48.82 & 19.75 & 76.78 & 72.72 & 71.82\\
    & DA$^3$ (ours) & 2.78 & 96.58 & \fbox{93.74} & \fbox{93.27} & 6.33 & 92.56 & 87.60 & 85.91 \\
    \hline
    \multirow{7}{*}{RTE} 
    & Original & 72.56 &&&& 78.34 \\
    & TextFooler & 1.44 & 98.01 & 68.66 & 79.60 & 5.05 & 93.55 & 67.74 & 87.56 \\
    & TextBugger & 2.53 & 96.52 & 68.66 & 83.08 & 9.75 & 87.56 & 70.05 & 81.57 \\
    & DeepWordBug & 4.33 & 94.03 & \textbf{79.60} & \textbf{88.06} & 16.25 & 79.26 & 69.59 & 76.04 \\
    & BERT-Attack & 3.61 & 95.02 & 67.16 & 72.64 & 1.44 & 98.16 & \fbox{70.51} & \textbf{90.32} \\
    & A2T & 8.66 & 88.06 & 62.69 & 25.87 & 16.97 & 78.34 & 67.28 & 77.88 \\
    & DA$^3$ (ours) & 1.08 & 98.51 & \fbox{72.14} & \fbox{86.07} & 7.22 & 90.78 & \textbf{71.43} & \fbox{88.94} \\
    \hline
    \multirow{7}{*}{MRPC} 
    & Original & 87.75 &&&& 91.18 \\
    & TextFooler & 2.94 & 96.65 & 58.38 & \fbox{91.62} & 4.90 & 94.62 & 35.48 & 94.62 \\
    & TextBugger & 7.35 & 91.60  & 62.85 & 87.15 & 9.80 & 89.25 & 34.68 & 89.25 \\
    & DeepWordBug & 10.05 & 88.55 & \fbox{72.35} & 86.31 & 12.01 & 86.83 & \fbox{47.31} & 86.83 \\
    & BERT-Attack & 9.56 & 89.11 & 55.31 & 61.39 & 2.45 & 97.31 & 34.95 & \fbox{97.04} \\
    & A2T & 30.88 & 64.80 & 46.65 & 26.54 & 49.51 & 45.70 & 21.51 & 45.43 \\
    & DA$^3$ (ours) & 0.74 & 99.16 & \textbf{74.86} & \textbf{93.29} & 0.49 & 99.46 & \textbf{50.27} & \textbf{99.46} \\
    \hline
    \end{tabular}}
    \label{tab: main results}
\end{table*}

\begin{table}[t]
    \centering
    \caption{Evaluation results on the black-box \textsc{LLaMA2-7b} model. Results of \textsc{LLaMA2-7b} are the average of zero-shot prompting with five different prompts.}
    \resizebox{\linewidth}{!}{
    \begin{tabular}{c|c|cccc}
    \hline
    \multicolumn{1}{c|}{\multirow{2}{*}{\textbf{Dataset}}} & \multicolumn{1}{c|}{\multirow{2}{*}{\textbf{Model}}} & \multicolumn{4}{c}{\textbf{\textsc{LLaMA2-7b}}} \\
    & & ACC$\downarrow$ & ASR$\uparrow$ & $\text{NASR}_{MSP}$ $\uparrow$ & $\text{NASR}_{MD}$ $\uparrow$ \\
    \hline
    \multirow{7}{*}{SST-2} 
    & Original & 89.91 \\
    & TextFooler & 68.97 & 23.81 & 22.97 & 23.58 \\
    & TextBugger & 84.50 & 6.89 & 6.51 & 6.69\\
    & DeepWordBug & 81.97 & 9.49 & 9.01 & 9.39 \\
    & BERT-Attack & 66.42 & 26.61 & \fbox{25.81} & \fbox{26.38} \\
    & A2T & 81.33 & 10.63 & 10.14 & 10.15 \\
    & DA$^3$ (ours)& 64.19 & 29.42 & \textbf{28.68} & \textbf{29.14}\\
    \hline
    \multirow{7}{*}{CoLA} 
    & Original & 70.97 \\
    & TextFooler & 31.95 & 57.65 & 52.13 & 57.09 \\
    & TextBugger & 39.41 & 48.22 & 42.49 & 47.22 \\
    & DeepWordBug & 31.93 & 61.23 & \textbf{56.67} & \textbf{60.58} \\
    & BERT-Attack & 39.98 & 46.07 & 40.97 & 45.68 \\
    & A2T & 40.38 & 45.09 & 39.81 & 37.75 \\
    & DA$^3$ (ours) & 33.06 & 58.51 & \fbox{53.39} & \fbox{57.69} \\
    \hline
    \multirow{7}{*}{RTE} 
    & Original & 57.76 && \\
    & TextFooler & 53.29 & 12.62 & 10.54 & 12.11 \\
    & TextBugger & 56.39 & 5.62 & 3.77 & 5.10 \\
    & DeepWordBug & 51.05 & 12.78 & 9.76 & 12.39 \\
    & BERT-Attack & 44.33 & 24.96 & \fbox{20.30} & \fbox{24.05} \\
    & A2T & 48.52 & 21.40 & 17.45 & 19.72 \\
    & DA$^3$ (ours) & 42.81 & 28.95 & \textbf{24.26} & \textbf{26.87} \\
    \hline
    \multirow{7}{*}{MRPC} 
    & Original & 67.94 && \\
    & TextFooler & 61.96 & 14.32 & 9.69 & 7.74 \\
    & TextBugger & 65.25 & 8.60 & 6.71 & 7.21 \\
    & DeepWordBug & 63.97 & 9.59 & 6.77 & 8.87 \\
    & BERT-Attack & 60.64 & 15.47 & 10.99 & \fbox{14.82} \\
    & A2T & 60.19 & 15.40 & \fbox{11.06} & 14.17 \\
    & DA$^3$ (ours) & 59.85 & 17.92 & \textbf{12.22} & \textbf{16.84} \\
    \hline
    \end{tabular}}
    \label{tab: transfer_llama}
\end{table}

\paragraph{Attack Baselines.}
We use two character-level attack methods, DeepWordBug~\cite{deepwordbug-2018} and TextBugger~\cite{textbugger-2019}, and three word-level attack methods, TextFooler~\cite{textfooler-2020}, BERT-Attack~\cite{li-etal-2020-bert-attack} and A2T~\cite{yoo2021towards}. Detailed descriptions are listed in Appendix~\ref{ssec:baselines}.

\paragraph{Datasets.}
We evaluate DA$^3$ on four different types of tasks: sentiment analysis task -- SST-2~\cite{socher-etal-2013-recursive}, grammar correctness task -- CoLA~\cite{warstadt-etal-2019-neural}, textual entailment task -- RTE~\cite{wang2019glue}, and textual similarity task -- MRPC~\cite{dolan-brockett-2005-automatically}. Detailed descriptions and statistics of each dataset are shown in Appendix~\ref{ssec:datasets}.

\paragraph{Implementation Details}
The backbone models of DA$^3$ are \textsc{BERT-base} or \textsc{RoBERTa-base} models fine-tuned on corresponding downstream datasets.
We use \textsc{BERT-base} and \textsc{RoBERTa-base} as white-box victim models and \textsc{LLaMA2-7b} as the black-box victim model.
More detailed information about hyperparameters and settings is in Appendix~\ref{sec: hyperparameters}. 
The prompts used for the black-box \textsc{LLaMA2-7b} are listed in Appendix~\ref{sec: prompts}

\section{Experimental Results and Analysis}

In this section, we conduct experiments and analysis to answer five research questions: 
\begin{itemize}[leftmargin=0.27cm]
    \setlength{\itemsep}{-0.4em}
    \item \textbf{RQ1} Will DA$^3$ effectively attack the white-box language models?
    \item \textbf{RQ2} Are generated adversarial examples transferable to the black-box \textsc{LLaMA2-7b} model?
    \item \textbf{RQ3} Will human judges find the quality of the generated adversarial examples reasonable?
    \item \textbf{RQ4} How do $\mathcal{L}_{DAL}$ components impact DA$^3$?
    \item \textbf{RQ5} Does $\mathcal{L}_{DAL}$ outperform other attack losses?
\end{itemize}

\subsection{Automatic Evaluation Results}
We use the adversarial examples generated by DA$^3$ with \textsc{BERT-base} or \textsc{RoBERTa-base} as the backbone to attack the white-box \textsc{BERT-base} and \textsc{RoBERTa-base} models, respectively.
White-box models have been fine-tuned on the corresponding datasets and are accessible during our fine-tuning phase.
Besides, considering that LLMs are widely used, expensive to fine-tune, and often not open source, we evaluate the attack transferability of the adversarial examples, which are generated by DA$^3$ with \textsc{BERT-base} as the backbone, on the black-box \textsc{LLaMA2-7b} model, which is not available during DA$^3$ fine-tuning.
The experimental results on ACC, ASR, and NASR are shown in Table~\ref{tab: main results}.

\paragraph{Attack Performance (RQ1).} 

When attacking white-box models, DA$^3$ obtains the best or second-to-best performance regarding NASR on most datasets. 
Aside from DA$^3$, some baseline methods perform well on one of the victim models. For example, TextFooler works well on \textsc{BERT-base}, while its $\text{NASR}_{MSP}$ decreases drastically compared to ASR on SST-2, RTE, and MRPC. Similarly, BERT-Attack shows good performance on \textsc{RoBERTa-base}, while its $\text{NASR}_{MSP}$ is notably lower than its ASR, especially on SST-2, RTE, and MRPC. This phenomenon indicates these adversarial examples are relatively easy to detect using MSP detection.
Considering the results of both victim models, DA$^3$ consistently produces reasonable and favorable outcomes when attacking white-box models, which proves the effectiveness of DA$^3$.

We also report \%Words and SS in Appendix~\ref{sec: more_eval}. DA$^3$ achieves best or second-to-best \%Words and comparable SS compared to baselines across datasets on both victim models.

\paragraph{Transferability to LLMs (RQ2).}\footnote{We also present results on \textsc{Mistral-7b} and the analysis on why the generated samples can be transferred to another LLMs in Appendix~\ref{sec: more_eval}. The results show DA$^3$ achieves the best performance in most cases when attacking \textsc{Mistral-7b}.} When attacking the black-box \textsc{LLaMA2-7b} model, DA$^3$ performs the best on SST-2, RTE, and MRPC, outperforming baselines in all evaluation metrics. On CoLA, DA$^3$ achieves second-to-best results on NASR.
Further analysis and visualization of attack performance on \textsc{LLaMA2-7b} across five different prompts are displayed in Appendix~\ref{sec: prompts_vis}. 
DA$^3$ consistently surpasses all baselines across five prompts.

The experimental results underscore the substantial advantage of our model when generalizing generated adversarial examples to the black-box \textsc{LLaMA2-7b} model, compared to baselines.


\begin{table}[t]
    \centering
    \caption{Grammar correctness, prediction accuracy and semantic preservation of original examples (denoted as Orig.) and adversarial examples generated by DA$^3$.}
    \resizebox{\linewidth}{!}{
    \begin{tabular}{c|cc|cc|cc}
    \hline
        \multirow{2}{*}{\textbf{Dataset}} & \multicolumn{2}{c|}{\textbf{Grammar}} & \multicolumn{2}{c|}{\textbf{Accuracy}} & \multicolumn{2}{c}{\textbf{Semantic}} \\
         & DA$^3$ & Orig. & DA$^3$ & Orig. & DA$^3$ & TextFooler \\
        \hline
        SST-2 & 4.12 & 4.37 & 0.68 & 0.74 & 0.71 & 0.66 \\
        MRPC & 4.62	& 4.86 & 0.68 & 0.76 & 0.88 & 0.84 \\
    \hline
    \end{tabular}
    }
    \label{tab: human_eval}
\end{table}

\subsection{Human Evaluation (RQ3)}
Given that our goal is to generate high-quality adversarial examples that preserve the original semantics and remain imperceptible to humans, we perform human evaluations to assess the adversarial examples generated by DA$^3$ using \textsc{BERT-base} as the backbone. These evaluations focus on grammar, prediction accuracy, and semantic preservation on SST-2 and MRPC datasets. 
For this purpose, three human judges evaluate 50 randomly selected original-adversarial pairs from each dataset. Detailed annotation guidelines are in Appendix~\ref{sec:annotation}.

First, human raters are tasked with evaluating the grammar correctness and making predictions of a shuffled mix of the sampled original and adversarial examples. Grammar correctness is scored from 1-5~\citep{li-etal-2020-bert-attack, textfooler-2020}. 
Then, human judges assess the semantic preservation of adversarial examples, determining whether they maintain the original semantics. We follow \citet{textfooler-2020} and ask human judges to classify adversarial examples as similar (1), ambiguous (0.5), or dissimilar (0) to the original examples. We compare DA$^3$ with the best baseline model, TextFooler, on semantic preservation for better evaluation.
We take the average scores among human raters for grammar correctness and semantic preservation and take the majority class as the predicted label.

As shown in Table~\ref{tab: human_eval}, grammar correctness scores of adversarial examples generated by DA$^3$ are similar to those of original examples. 
While word perturbations make predictions more challenging, adversarial examples generated by DA$^3$ still show decent accuracy.
Compared to TextFooler, DA$^3$ can better preserve semantic similarity to original examples.
Some generated adversarial examples are displayed in Appendix~\ref{case study}.

\begin{table}[t]
    \centering
    \caption{Ablation study on \textsc{BERT-base} regarding MSP.}
    \resizebox{\linewidth}{!}{
    \begin{tabular}{c|c|c|c|cc}
    \hline
    \textbf{Dataset} & \textbf{Model} & \textbf{ACC}$\downarrow$ & \textbf{ASR}$\uparrow$ & $\text{\textbf{NASR}}_{MSP}$$\uparrow$ & $\text{\textbf{DR}}_{MSP}$$\downarrow$ \\
    \hline
    \multirow{2}{*}{SST-2} 
    & DA$^3$ & 21.10 & 77.17 & \textbf{54.22} & \textbf{29.74} \\ 
    & (w/o MSP) & 1.61 & 98.26 & 47.27 & 51.89 \\ 
    \hline
    \multirow{2}{*}{CoLA} 
    & DA$^3$ & 2.78 & 96.58 & \textbf{93.74} & \textbf{2.93} \\ 
    & (w/o MSP) & 2.11 & 97.40 & 93.15 & 4.36 \\ 
    \hline
    \multirow{2}{*}{RTE} 
    & DA$^3$ & 1.08 & 98.51 & \textbf{72.14} & \textbf{26.77} \\
    & (w/o MSP) & 1.08 & 98.51 & 70.65 & 28.28 \\ 
    \hline
    \multirow{2}{*}{MRPC} 
    & DA$^3$ & 0.74 & 99.16 & \textbf{74.86} & \textbf{24.51} \\ 
    & (w/o MSP) & 0.74 & 99.16 & 73.18 & 26.20\\ 
    \hline
    \end{tabular}}
    \label{tab: ablation_MSP}
\end{table}

\begin{table}[t]
    \centering
    \caption{Ablation study on \textsc{BERT-base} regarding MD.}
    \resizebox{\linewidth}{!}{
    \begin{tabular}{c|c|c|c|cc}
    \hline
    \textbf{Dataset} & \textbf{Model} & \textbf{ACC}$\downarrow$ & \textbf{ASR}$\uparrow$ & $\text{\textbf{NASR}}_{MD}$$\uparrow$ & $\text{\textbf{DR}}_{MD}$$\downarrow$ \\
    \hline
    \multirow{2}{*}{SST-2} 
    & DA$^3$ & 21.10 & 77.17 & 75.06 & \textbf{2.73} \\ 
    & (w/o MD) & 15.60 & 83.13 & \textbf{80.77} & 2.84 \\ 
    \hline
    \multirow{2}{*}{CoLA} 
    & DA$^3$ & 2.78 & 96.58  & \textbf{93.27} & \textbf{3.42}  \\ 
    & (w/o MD) & 2.30 & 97.17  & 90.55 & 6.80 \\ 
    \hline
    \multirow{2}{*}{RTE} 
    & DA$^3$ & 1.08 & 98.51 & \textbf{86.07} & \textbf{12.63} \\
    & (w/o MD) & 1.08 & 98.51 & 85.57 & 13.13 \\ 
    \hline
    \multirow{2}{*}{MRPC} 
    & DA$^3$ & 0.74 & 99.16 & \textbf{93.29} & \textbf{5.90} \\ 
    & (w/o MD) & 1.72 & 98.04 & 90.22 & 7.98 \\ 
    \hline
    \end{tabular}}
    \label{tab: ablation_MD}
\end{table}

\subsection{Ablation Study (RQ4)}
To analyze the effectiveness of different components of $\mathcal{L}_{DAL}$, we conduct an ablation study on \textsc{BERT-base}. The results of the ablation study are shown in Table~\ref{tab: ablation_MSP} and Table~\ref{tab: ablation_MD}.

\paragraph{MSP Loss.}
We ablate $\mathcal{L}_{MSP}$ during fine-tuning to assess the efficacy of $\mathcal{L}_{MSP}$. $\mathcal{L}_{MSP}$ helps improve $\text{NASR}_{MSP}$ and MSP Detection Rate ($\text{DR}_{MSP}$), which is the ratio of $|\mathcal{D}_{MSP}|$ to the total number of successful adversarial examples, across all datasets.
An interesting finding is that on SST-2 and CoLA, although models without $\mathcal{L}_{MSP}$ perform better in terms of ASR, the situation deteriorates when considering detectability, leading to lower $\text{NASR}_{MSP}$ and higher $\text{DR}_{MSP}$ compared to the model with $\mathcal{L}_{DAL}$.

\paragraph{MD Loss.}
We ablate $\mathcal{L}_{MD}$ during fine-tuning to assess the efficacy of $\mathcal{L}_{MD}$. $\mathcal{L}_{MD}$ helps improve MD Detection Rate ($\text{DR}_{MD}$), which is the ratio of $|\mathcal{D}_{MD}|$ to the number of successful adversarial examples, across all datasets. $\mathcal{L}_{MD}$ also improves $\text{NASR}_{MD}$ on all datasets except SST-2. 
A similar finding on CoLA exists that although models without $\mathcal{L}_{MD}$ perform better on ASR, the performance worsens when considering detectability.

The ablation study shows that both $\mathcal{L}_{MSP}$ and $\mathcal{L}_{MD}$ are effective on most datasets.

\begin{figure}[t]
     \centering
     \includegraphics[width=0.9\columnwidth]{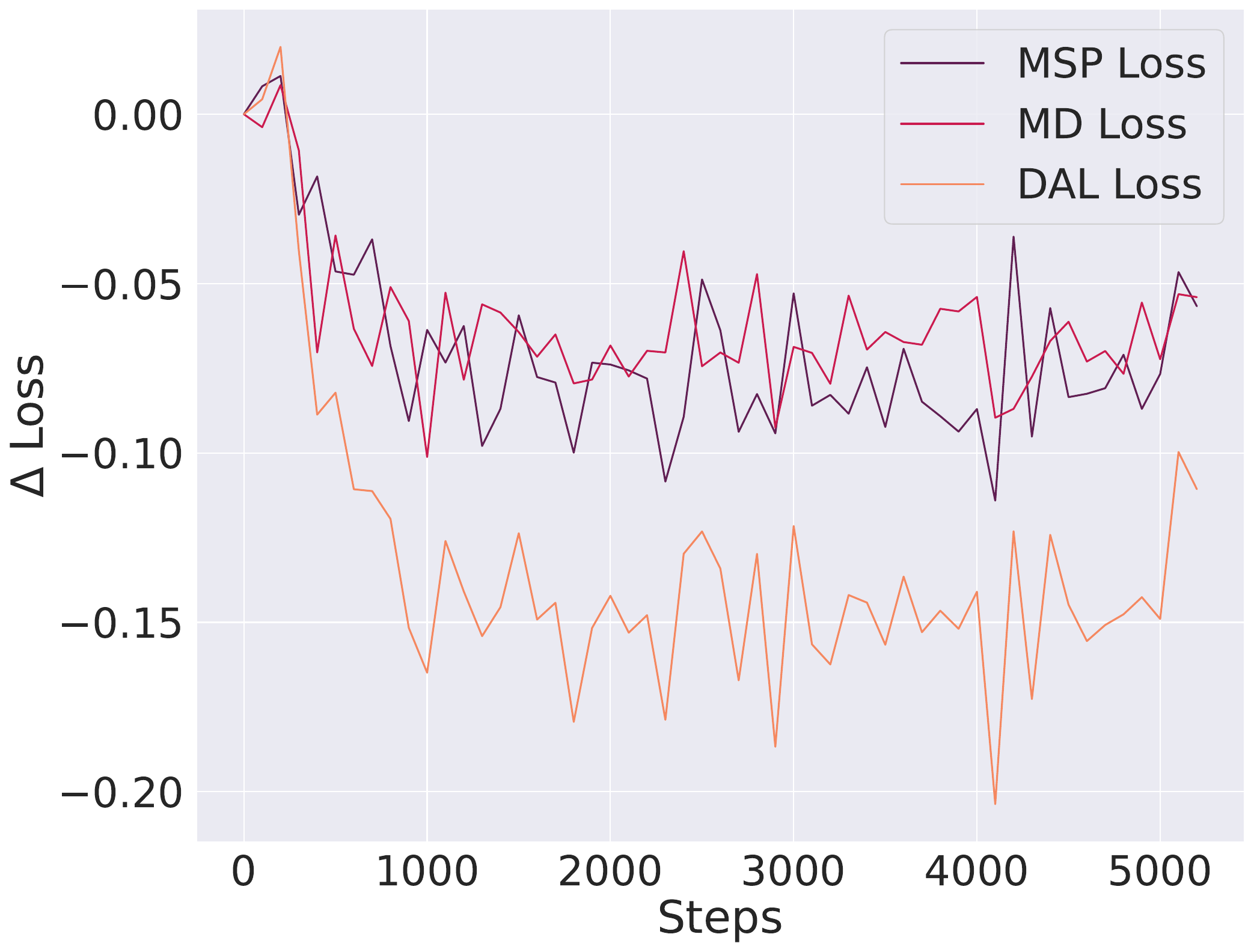}
    \caption{The change of $\mathcal{L}_{MSP}$, $\mathcal{L}_{MD}$, and $\mathcal{L}_{DAL}$ throughout the fine-tuning phase of DA$^3$ with \textsc{BERT-base} as backbone on SST-2. The x-axis represents fine-tuning steps; the y-axis represents the change of loss compared to the initial loss.}
    \label{fig: loss_viz}
\end{figure}

\subsection{Loss Visualization and Analysis (\textbf{RQ4})}
To better understand how different loss components contribute to DA$^3$, we visualize the changes of $\mathcal{L}_{MSP}$, $\mathcal{L}_{MD}$, and $\mathcal{L}_{DAL}$ throughout the fine-tuning phase of DA$^3$ with \textsc{BERT-base} as backbone on SST-2 dataset, as illustrated in Figure~\ref{fig: loss_viz}. 

We observe that all three losses exhibit oscillating descent and eventual convergence. 
Although the overall trends of $\mathcal{L}_{MSP}$ and $\mathcal{L}_{MD}$ are consistent, a closer examination reveals that they often exhibit opposite trends at each step, especially in the initial stages. 
Despite both losses sharing a common goal of reducing distribution shifts between adversarial examples and original examples, this observation reveals a potential trade-off relationship between them. 
One possible interpretation is that, on the one hand, minimizing $\mathcal{L}_{MSP}$ increases the confidence of wrong predictions, aligning with the objective of the adversarial attack task to induce incorrect predictions.
On the other hand, minimizing $\mathcal{L}_{MD}$ encourages the generated adversarial sentences to resemble the original ones more closely, loosely akin to the objective of the masked language modeling task to restore masked tokens to their original values. 
While these two objectives are not inherently conflicting, an extreme standpoint reveals that when the latter objective is fully satisfied -- meaning the model generates identical examples to the original ones -- the former objective naturally becomes untenable.

\subsection{Loss Comparison (RQ5)}
Other than using our $\mathcal{L}_{DAL}$, we also explore other loss variants: $\mathcal{L}_{NCE}$ and $\mathcal{L}_{FCE}$.

\begin{table}[t]
    \centering
    \caption{Comparison of DA$^3$ using \textsc{BERT-base} as backbone with loss variants.}
    \resizebox{\linewidth}{!}{
    \begin{tabular}{c|c|c|c|cc|cc}
    \hline
    \multirow{2}{*}{\textbf{Dataset}} & \multirow{2}{*}{\textbf{Model}} & \multirow{2}{*}{\textbf{ACC}$\downarrow$} & \multirow{2}{*}{\textbf{ASR}$\uparrow$} & \multicolumn{2}{c|}{\textbf{MSP}} & \multicolumn{2}{c}{\textbf{MD}} \\
    & & & & NASR$\uparrow$ & DR$\downarrow$ & NASR$\uparrow$ & DR$\downarrow$ \\
    \hline
    \multirow{3}{*}{SST-2} 
    & w/ $\mathcal{L}_{NCE}$ & 18.23 & 80.27 & 55.71 & 30.60 & 76.30 & 4.95 \\ 
    & w/ $\mathcal{L}_{FCE}$ & 17.66 & 80.89 & \textbf{63.03} & \textbf{22.09} & \textbf{78.04} & 3.53 \\
    & ours & 21.10 & 77.17 & 54.22 & 29.74 & 75.06 & \textbf{2.73} \\ 
    \hline
    \multirow{3}{*}{CoLA} 
    & w/ $\mathcal{L}_{NCE}$ & 2.03 & 97.52 & \textbf{94.10} & 3.51 & 92.80 & 4.84 \\ 
    & w/ $\mathcal{L}_{FCE}$ & 3.07 & 96.22 & 93.98 & \textbf{2.33} & 91.97 & 4.42 \\ 
    & ours & 2.78 & 96.58 & 93.74 & 2.93 & \textbf{93.27} & \textbf{3.42} \\ 
    \hline
    \multirow{3}{*}{RTE} 
    & w/ $\mathcal{L}_{NCE}$ & 1.08 & 98.51 & 71.14 & 27.78 & 85.57 & 13.13 \\
    & w/ $\mathcal{L}_{FCE}$ & 1.44 & 98.01 & 69.65 & 28.93 & 85.07 & 13.20 \\
    & ours & 1.08 & 98.51 & \textbf{72.14} & \textbf{26.77} & \textbf{86.07} & \textbf{12.63} \\
    \hline
    \multirow{3}{*}{MRPC} 
    & w/ $\mathcal{L}_{NCE}$ & 2.45 & 97.21 & 71.79 & 26.15 & 89.39 & 8.05 \\
    & w/ $\mathcal{L}_{FCE}$ & 0.74 & 99.16 & 68.99 & 30.42 & 91.34 & 7.89 \\
    & ours & 0.74 & 99.16 & \textbf{74.86} & \textbf{24.51}  & \textbf{93.29} & \textbf{5.90} \\ 
    \hline
    \end{tabular}}
    \label{tab: loss_comparison}
\end{table}

Minimizing the negative of regular cross-entropy loss (denoted as $\mathcal{L}_{NCE}$) or minimizing the cross-entropy loss of flipped adversarial labels (denoted as $\mathcal{L}_{FCE}$) are two simple ideas as baseline attack methods.
We replace $\mathcal{L}_{DAL}$ with $\mathcal{L}_{NCE}$ or $\mathcal{L}_{FCE}$ during the fine-tuning phase to assess the efficacy of our loss $\mathcal{L}_{DAL}$. The results in Table~\ref{tab: loss_comparison} show that $\mathcal{L}_{DAL}$ outperforms the other two losses across all evaluation metrics on RTE and MRPC datasets. On CoLA dataset, $\mathcal{L}_{DAL}$ achieves better or similar performance compared to $\mathcal{L}_{NCE}$ and $\mathcal{L}_{FCE}$. 
While $\mathcal{L}_{DAL}$ may not perform as well as $\mathcal{L}_{NCE}$ and $\mathcal{L}_{FCE}$ on SST-2, given its superior performance on the majority of datasets, we believe $\mathcal{L}_{DAL}$ is more effective than $\mathcal{L}_{NCE}$ and $\mathcal{L}_{FCE}$ generally.

\section{Conclusion}
We analyze the adversarial examples generated by previous attack methods and identify distribution shifts between adversarial examples and original examples in terms of MSP and MD.
To address this, we propose a Distribution-Aware Adversarial Attack (DA$^3$) method with the Data Alignment Loss and introduce a novel evaluation metric, NASR, which integrates out-of-distribution detection into the assessment of successful attacks.
Our experiments validate the attack effectiveness of DA$^3$ on \textsc{BERT-base} and \textsc{RoBERTa-base} and the transferability of adversarial examples generated by DA$^3$ on the black-box \textsc{LLaMA2-7b}.

\section*{Limitations}
We analyze the distribution shifts between adversarial examples and original examples in terms of MSP and MD, which exist in most datasets. Nevertheless, the MD distribution shift is not very obvious in some datasets like MRPC. This indicates that MD detection may not always effectively identify adversarial examples.
However, we believe that since such a distribution shift is present in many datasets, we still need to consider MD detection. Furthermore, our experiments demonstrate that considering distribution shift is not only effective for NASR but also enhances the performance of the model in ASR.

\section*{Ethics Statement}

There exists a potential risk associated with our proposed attack methods -- they could be used maliciously to launch adversarial attacks against off-the-shelf systems. Despite this risk, we emphasize the necessity of conducting studies on adversarial attacks. Understanding these attack models is crucial for the research community to develop effective defenses against such attacks.

\section*{Acknowledgements}
This work is supported in part by NSF under grants III-2106758, and POSE-2346158.

\bibliography{anthology,custom}

\clearpage
\appendix
\section*{\centering Appendix}
\section{Evaluation Metrics}
\label{ssec:metrics}

\paragraph{Percentage of Perturbed Words (\%Words).}
Percentage of Perturbed Words (\%Words) is used to measure how much a text has been altered or perturbed from its original form. \%Words is formulated as
\begin{equation}
    \text{\%Words} = \frac{\text{Number of Perturbed Words}}{\text{Total Number of Words}} \times 100.
    \notag
\end{equation}

\paragraph{Semantic Similarity (SS).}
We calculate Semantic Similarity (SS) using sentence semantic similarity between $x^{orig}$ and $x^{adv}$. 
Specifically, we transform the two sentences into high-dimensional sentence embeddings using the Universal Sentence Encoder (USE)~\cite{cer2018universal}. We then approximate their semantic similarity by calculating the cosine similarity score between these vectors.

\paragraph{Attack Success Rate (ASR).}
Attack Success Rate (ASR) is defined as the percentage of generated adversarial examples that successfully deceive model predictions. 
Thus, ASR is formulated as
\begin{equation}
    \text{ASR} = \frac{|\{x^{orig}\mid y^{adv} \neq y^{orig}, x^{orig}\in\mathcal{X}\}|}{|\mathcal{X}|}.
    \notag
\end{equation}

These definitions are consistent with prior work.

\section{More Implementation Details}
\subsection{Baselines}
\label{ssec:baselines}

\noindent \textbf{DeepWordBug}~\cite{deepwordbug-2018} uses two scoring functions to determine the most important words and then adds perturbations through random substation, deletion, insertion, and swapping letters in the word while constrained by the edit distance.

\smallskip
\noindent \textbf{TextBugger}~\cite{textbugger-2019} finds important words through the Jacobian matrix or scoring function and then uses insertion, deletion, swapping, substitution with visually similar words, and substitution with semantically similar words.

\smallskip
\noindent \textbf{TextFooler}~\cite{textfooler-2020} uses the prediction change before and after deleting the word as the word importance score and then replaces each word in the sentence with synonyms until the prediction label of the target model changes.

\smallskip
\noindent \textbf{BERT-Attack}~\cite{li-etal-2020-bert-attack} finds the vulnerable words through logits from the target model and then uses BERT to generate perturbations based on the top-K predictions.

\smallskip
\noindent \textbf{A2T}~\cite{yoo2021towards} employs a gradient-based method for ranking word importance, iteratively replacing each word with top synonyms generated from counter-fitting word embeddings~\cite{mrkvsic2016counter}.

For the implementation of baselines, we use the TextAttack\footnote{\url{https://github.com/QData/TextAttack} (MIT License).} package with its default parameters.

\subsection{Datasets}
\label{ssec:datasets}

\noindent \textbf{SST-2.} The Stanford Sentiment Treebank~\cite{socher-etal-2013-recursive} is a binary sentiment classification task. It consists of sentences extracted from movie reviews with human-annotated sentiment labels. 

\smallskip
\noindent \textbf{CoLA.} The Corpus of Linguistic Acceptability~\cite{warstadt-etal-2019-neural} contains English sentences extracted from published linguistics literature, aiming to check grammar correctness.

\smallskip
\noindent \textbf{RTE.} The Recognizing Textual Entailment dataset~\cite{wang2019glue} is derived from a combination of news and Wikipedia sources, aiming to determine whether the given pair of sentences entail each other. 

\smallskip
\noindent \textbf{MRPC.} The Microsoft Research Paraphrase Corpus~\cite{dolan-brockett-2005-automatically} comprises sentence pairs sourced from online news articles. These pairs are annotated to indicate whether the sentences are semantically equivalent.

Data statistics for each dataset are shown in Table~\ref{tab:data-statistics}.

\begin{table}[t]
\centering
\caption{Dataset statistics.}
\resizebox{\linewidth}{!}{
\begin{tabular}{cccc} \hline
\textbf{Dataset} & \textbf{Train} & \textbf{Validation} & \textbf{Description} \\ \hline
SST-2 & 67,300 & 872 & Sentiment analysis \\
CoLA & 8,550 & 1,043 & Grammar correctness \\
RTE & 2,490 & 277 & Textual entailment \\
MRPC & 3,670 & 408 & Textual similarity \\ \hline
\end{tabular}}
\label{tab:data-statistics}
\end{table}

\begin{table}[t]
\centering
\caption{Hyperparameters of different datasets.}
\resizebox{\linewidth}{!}{
\begin{tabular}{c|c|cccc}
\hline
 \textbf{Backbone} & \textbf{Hyperparameter} & \textbf{SST-2} & \textbf{CoLA} & \textbf{RTE} & \textbf{MRPC} \\
 \hline
\multirow{3}{*}{\textsc{BERT-base}} 
& batch size & 128 & 128 & 32 & 128 \\
& learning rate & 1e-4 & 5e-5 & 1e-5 & 1e-3 \\
& \% masked tokens & 30 & 30 & 30 & 30 \\
\hline
\multirow{3}{*}{\textsc{RoBERTa-base}} 
& batch size & 128 & 128 & 32 & 128 \\
& learning rate & 5e-5 & 1e-4 & 1e-5 & 1e-3 \\
& \% masked tokens & 30 & 30 & 30 & 30 \\
\hline
\end{tabular}
}
\label{tab: hyperparameters}
\end{table}

\begin{table*}[t]
\centering
\caption{Prompt template for different datasets. \{instruct\} is replaced by different instructions in Table~\ref{tab: instructs}, while \{text\} is replaced with input sentence.}
\resizebox{\linewidth}{!}{
\begin{tabular} {p{0.06\textwidth}p{\textwidth}}
\hline
\textbf{Dataset} & \textbf{Prompt} \\
 \hline
SST-2 & ``\{instruct\} Respond with `positive' or `negative' in lowercase, only one word. \textbackslash nInput: \{text\}\textbackslash nAnswer:'' \\
\hline
CoLA &  ``\{instruct\} Respond with `acceptable' or `unacceptable' in lowercase, only one word.\textbackslash nInput: \{text\}\textbackslash nAnswer:'',
\\
\hline
RTE & ``\{instruct\} Respond with `entailment' or `not\_entailment' in lowercase, only one word.\textbackslash nInput: \{text\}\textbackslash nAnswer:
\\
\hline
MRPC &  ``\{instruct\} Respond with `equivalent' or `not\_equivalent' in lowercase, only one word.\textbackslash nInput: \{text\} \textbackslash nAnswer:
\\
\hline
\end{tabular}
}
\label{tab: prompts}
\end{table*}

\begin{table*}[t]
\centering
\caption{Different instructions used for different runs.}
\resizebox{\linewidth}{!}{
\begin{tabular} {p{0.06\textwidth}p{\textwidth}}
\hline
\textbf{Dataset} & \textbf{Prompt} \\
 \hline
SST-2 & ``Evaluate the sentiment of the given text.'' \\
& ``Please identify the emotional tone of this passage.'' \\
& ``Determine the overall sentiment of this sentence.'' \\
& ``After examining the following expression, label its emotion.''\\
& ``Assess the mood of the following quote.'' \\
\hline
CoLA & ``Assess the grammatical structure of the given text.''\\
& ``Assess the following sentence and determine if it is grammatically correct.''\\
& ``Examine the given sentence and decide if it is grammatically sound.''\\
& ``Check the grammar of the following sentence.''\\
& ``Analyze the provided sentence and classify its grammatical correctness.''\\
\hline
RTE & ``Assess the relationship between sentence1 and sentence2.''\\
& ``Review the sentence1 and sentence2 and categorize their relationship.''\\
& ``Considering the sentence1 and sentence2, identify their relationship.''\\
& ``Please classify the relationship between sentence1 and sentence2.''\\
& ``Indicate the connection between sentence1 and sentence2.''\\
\hline
MRPC &  ``Assess whether sentence1 and sentence2 share the same semantic meaning.''\\
& ``Compare sentence1 and sentence2 and determine if they share the same semantic meaning.''\\
& ``Do sentence1 and sentence2 have the same underlying meaning?''\\
& ``Do the meanings of sentence1 and sentence2 align?''\\
& ``Please analyze sentence1 and sentence2 and indicate if their meanings are the same.''\\
\hline
\end{tabular}
}
\label{tab: instructs}
\end{table*}

\begin{figure*}[b]
     \centering
     \begin{subfigure}[t]{0.32\linewidth}
         \centering
         \includegraphics[width=\linewidth]{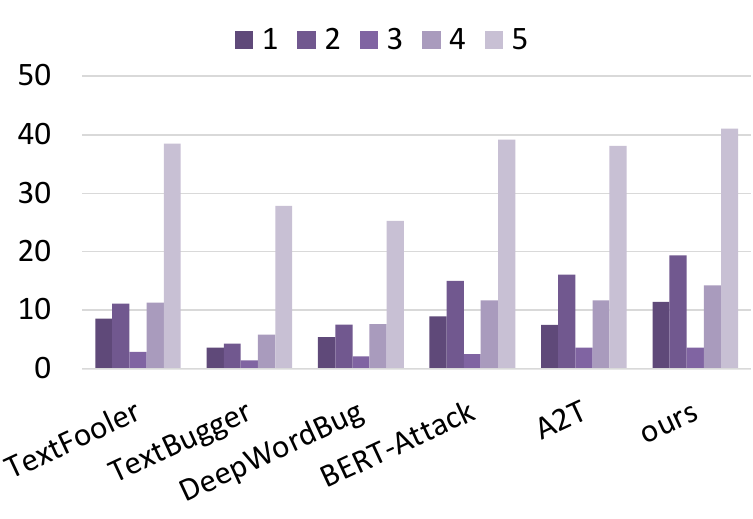}
         \caption{ASR}
     \end{subfigure}    
     \hfill
     \begin{subfigure}[t]{0.32\linewidth}
         \centering
         \includegraphics[width=\linewidth]{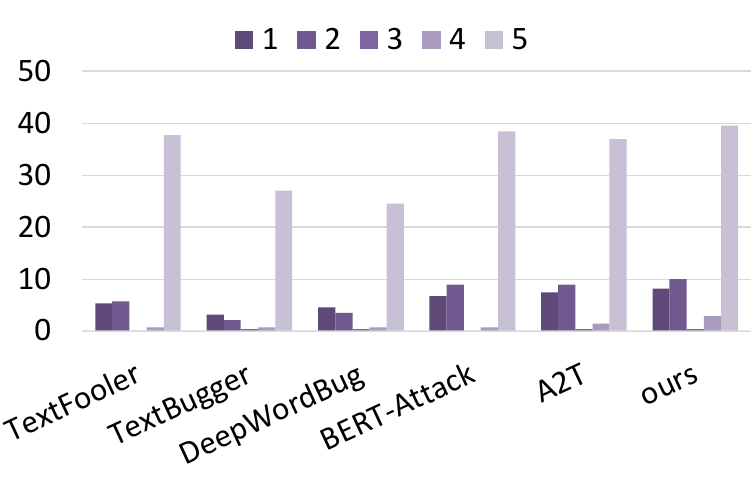}
         \caption{NASR (MSP)}
     \end{subfigure}    
     \hfill
     \begin{subfigure}[t]{0.32\linewidth}
         \centering
         \includegraphics[width=\linewidth]{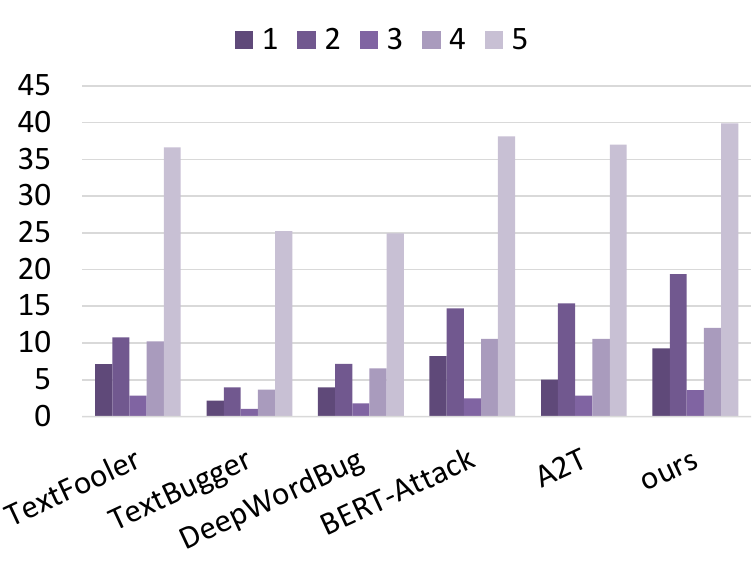}
         \caption{NASR (MD)}
     \end{subfigure}    
    \caption{Results of \textsc{LLaMA2-7b} across five different prompts on MRPC.}
    \label{fig: multi_prompts}
\end{figure*}
\subsection{Hyperparameters and More Settings}
\label{sec: hyperparameters}
For each experiment, the DA$^3$ fine-tuning phrase is executed for a total of 20 epochs. The learning rate is searched from $\left[1e-5, 1e-3\right]$. Up to 30\% of the tokens are masked during the fine-tuning phrase. 
The rank of the update matrices of \textsc{LoRA} is set to 8; \textsc{LoRA} scaling factor is 32; \textsc{LoRA} dropout value is set as 0.1.
The inference termination condition is set as 40\% of the tokens. 

Table~\ref{tab: hyperparameters} shows the hyperparameters used in experiments.

White-box experiments are conducted on two NVIDIA GeForce RTX 3090ti GPUs, and black-box experiments are conducted on two NVIDIA RTX A5000 24GB GPUs.

\subsection{Prompts Used for the Black-box LLM}
\label{sec: prompts}
The constructed prompt templates used for the Black-box LLM (\textsc{LLaMA2-7b}\footnote{LLaMA2 Community License}) are shown in Table~\ref{tab: prompts}. For each run, \{instruct\} in the prompt template is replaced by different instructions in Table~\ref{tab: instructs}, while \{text\} is replaced with the input sentence.

\begin{table}[t]
    \centering
    \caption{Evaluation results on the black-box \textsc{Mistral-7b} models. Results of \textsc{Mistral-7b} are the average of zero-shot prompting with five different prompts.}
    \resizebox{\linewidth}{!}{
    \begin{tabular}{c|c|cccc}
    \hline
    \multicolumn{1}{c|}{\multirow{2}{*}{\textbf{Dataset}}} & \multicolumn{1}{c|}{\multirow{2}{*}{\textbf{Model}}} & \multicolumn{4}{c}{\textbf{\textsc{Mistral-7b}}} \\
    & & ACC$\downarrow$ & ASR$\uparrow$ & $\text{NASR}_{MSP}$ $\uparrow$ & $\text{NASR}_{MD}$ $\uparrow$ \\
    \hline
    \multirow{7}{*}{SST-2} 
    & Original & 89.17 \\
    & TextFooler & 66.56 & 26.15 & 26.15 & 25.61 \\
    & TextBugger & 83.07 & 8.26 & 8.26 & 7.74 \\
    & DeepWordBug & 82.48 & 9.21 & 9.21 & 8.87 \\
    & BERT-Attack & 63.97 & 29.01 & \fbox{28.94} & \fbox{28.50} \\
    & A2T & 77.89 & 14.12 & 14.04 & 13.50 \\
    & DA$^3$ (ours)& 60.18 & 33.71 & \textbf{33.71} & \textbf{33.17} \\
    \hline
    \multirow{7}{*}{CoLA} 
    & Original & 79.35 \\
    & TextFooler & 27.84 & 66.20 & \fbox{57.59} & \textbf{63.57} \\
    & TextBugger & 38.28 & 52.52 & 46.36 & 48.26 \\
    & DeepWordBug & 34.67 & 58.99 & 51.69 & 53.87 \\
    & BERT-Attack & 33.25 & 59.58 & 52.23 & 55.96 \\
    & A2T & 35.70 & 56.36 & 49.26 & 51.86 \\
    & DA$^3$ (ours)& 29.11 & 66.12 & \textbf{63.41} & \fbox{62.49} \\
    \hline
    \multirow{7}{*}{RTE} 
    & Original & 80.94 \\
    & TextFooler & 65.20 & 24.35 & 24.35 & 24.17 \\
    & TextBugger & 77.91 & 6.95 & 6.95 & 6.86 \\
    & DeepWordBug & 77.98 & 6.33 & 6.33 & 6.24 \\
    & BERT-Attack & 56.73 & 33.18 & \fbox{33.18} & \fbox{33.12} \\
    & A2T & 57.69 & 32.11 & 32.11 & 32.11 \\
    & DA$^3$ (ours)& 54.08 & 35.98 & \textbf{35.71} & \textbf{35.45}\\
    \hline
    \multirow{7}{*}{MRPC} 
    & Original & 79.31 \\
    & TextFooler & 63.09 & 25.00 & 24.81 & 22.97 \\
    & TextBugger & 78.68 & 4.52 & 4.52 & 4.52 \\
    & DeepWordBug & 78.33 & 4.46 & 4.46 & 4.40 \\
    & BERT-Attack & 56.22 & 34.58 & \fbox{33.72} & \fbox{34.60} \\
    & A2T & 61.91 & 26.52 & 26.03 & 26.52 \\
    & DA$^3$ (ours)& 56.18 & 35.30 & \textbf{35.07 }& \textbf{35.38} \\
    \hline
    \end{tabular}}
    \label{tab: transfer_mistral}
    \vspace{-10pt}
\end{table}

\section{More Automatic Evaluation Results}
\label{sec: more_eval}
Experimental results of \%Words and SS on the white-box victim models \textsc{BERT-base} and \textsc{RoBERTa-base} are shown in Table~\ref{tab:words_ss_bert} and Table~\ref{tab:words_ss_roberta}.
DA$^3$ achieves best or second-to-best \%Words and comparable SS compared to baselines across datasets on both victim models.

The results of the generated adversarial examples by DA$^3$ with \textsc{BERT-base} as the backbone on attacking the white-box \textsc{Mistral-7b} model on CoLA, RTE, and MRPC are shown in Table~\ref{tab: transfer_mistral}. Our proposed DA$^3$ outperforms all other baselines.

Although \textsc{BERT-base}, \textsc{LLaMA2-7b}, and \textsc{Mistral-7b} have different structures and parameters, they are both trained on large text corpora. Thus, they share similar knowledge. From Table~\ref{tab: transfer_llama} and Table~\ref{tab: transfer_mistral}, we can see that BERT-based models (BERT-Attack and DA$^3$) perform better than other models in most cases, which confirms our explanations. Besides, the best transferability also shows that our proposed DA$^3$ can generate high-quality adversarial examples that are robust to the black-box LLMs.

\begin{table*}[t]
    \centering
    \caption{\%Words and SS results on the \textsc{BERT-base} victim model.}
    \resizebox{\linewidth}{!}{
    \begin{tabular}{c|cccccc|cccccc}
    \hline
    \textbf{Dataset} & \multicolumn{6}{c|}{\textbf{SST-2}} & \multicolumn{6}{c}{\textbf{CoLA}} \\
    \hline
    \textbf{Model} & TextFooler & TextBugger & DeepWordBug & BERT-Attack & A2T & DA$^3$ & TextFooler & TextBugger & DeepWordBug & BERT-Attack & A2T & DA$^3$ \\
    \textbf{\%Words} & 17.58 & 15.35 & 19.11 & 13.42 & \fbox{11.06} & \textbf{10.72} & 19.16 & 19.16 & 18.53 & \fbox{18.34} & 19.04 & \textbf{16.83} \\
    \textbf{SS} & 82.32 & \textbf{90.98} & 80.03 & 89.89 & \fbox{90.25} & 87.78 & 82.09 & \textbf{91.36} & 83.60 & \fbox{90.65} & 88.62 & 86.95 \\
    \hline
    \textbf{Dataset} & \multicolumn{6}{c|}{\textbf{RTE}} & \multicolumn{6}{c}{\textbf{MRPC}} \\
    \hline
    \textbf{Model} & TextFooler & TextBugger & DeepWordBug & BERT-Attack & A2T & DA$^3$ & TextFooler & TextBugger & DeepWordBug & BERT-Attack & A2T & DA$^3$ \\
    \textbf{\%Words} & 6.01 & 12.07 & 6.59 & 6.97 & \textbf{4.41} & \fbox{4.75} & 9.69 & 19.09 & 8.32 & 11.66 & \textbf{6.2} & \fbox{6.64} \\
    \textbf{SS} & 96.80 & \textbf{97.26} & 96.72 & 96.32 & \fbox{97.18} & 96.37 & 94.04 & \fbox{95.60} & 94.56 & 93.07 & \textbf{96.10} & 93.86 \\
    \hline
    \end{tabular}}
    \label{tab:words_ss_bert}
\end{table*}

\begin{table*}[t]
    \centering
    \caption{\%Words and SS results on the \textsc{RoBERTa-base} victim model.}
    \resizebox{\linewidth}{!}{
    \begin{tabular}{c|cccccc|cccccc}
    \hline
    \textbf{Dataset} & \multicolumn{6}{c|}{\textbf{SST-2}} & \multicolumn{6}{c}{\textbf{CoLA}} \\
    \hline
    \textbf{Model} & TextFooler & TextBugger & DeepWordBug & BERT-Attack & A2T & DA$^3$ & TextFooler & TextBugger & DeepWordBug & BERT-Attack & A2T & DA$^3$ \\
    \textbf{\%Words} & 18.73 & 18.03 & 22.70 & 14.33 & \textbf{12.30} & \fbox{12.58} & 19.07 & 18.40 & 19.10 & \fbox{17.31} & 17.60 & \textbf{17.29} \\
    \textbf{SS} & 81.58 & \textbf{90.37} & 75.26 & 86.44 & \fbox{89.48} & 86.98 & 83.31 & \textbf{91.90} & 83.22 & \fbox{90.49} & 90.15 & 85.99 \\
    \hline
    \textbf{Dataset} & \multicolumn{6}{c|}{\textbf{RTE}} & \multicolumn{6}{c}{\textbf{MRPC}} \\
    \hline
    \textbf{Model} & TextFooler & TextBugger & DeepWordBug & BERT-Attack & A2T & DA$^3$ & TextFooler & TextBugger & DeepWordBug & BERT-Attack & A2T & DA$^3$ \\
    \textbf{\%Words} & 6.96 & 7.93 & \fbox{5.27} & 6.59 & \textbf{3.93} & 6.38 & 12.50 & 18.84 & 13.18 & 10.09 & \textbf{7.04} & \fbox{8.10} \\
    \textbf{SS} & 96.35 & \fbox{97.32} & 96.93 & 96.67 & \textbf{97.69} & 94.88 & 92.12 & 93.28 & 90.44 & 93.13 & \textbf{95.96} & \fbox{94.12} \\
    \hline
    \end{tabular}}
    \label{tab:words_ss_roberta}
\end{table*}

\section{Annotation Guidelines}
\label{sec:annotation}

Here we provide the annotation guidelines for annotators:

\paragraph{Grammar.} Rate the grammaticality and fluency of the text between 1-5; the higher the score, the better the grammar of the text.

\paragraph{Prediction.} For SSTS-2 dataset, classify the sentiment of the text into negative (0) or positive (1); For MRPC dataset, classify if the two sentences are equivalent (1) or not\_equivalent (0).

\paragraph{Semantic.} Compare the semantic similarity between text1 and text2, and label with similar (1), ambiguous (0.5), and dissimilar (0).

\section{Examples of Generated Adversarial Sentences}
\label{case study}

Table~\ref{tab: case study} displays some original examples and the corresponding adversarial examples generated by DA$^3$. The table also shows the predicted results of the original or adversarial sentence using \textsc{BERT-base}. Blue words are perturbed into the red words.
Table~\ref{tab: case study} shows that DA$^3$ only perturbs a very small number of words, leading to model prediction failure. Besides, the adversarial examples generally preserve similar semantic meanings to their original inputs.

\begin{table*}[t]
\centering
\caption{Examples of generated adversarial sentences}
\resizebox{\linewidth}{!}{
\begin{tabular}{p{0.03\textwidth}p{0.9\textwidth}p{0.15\textwidth}}
\hline
 & \textbf{Sentence} & \textbf{Prediction}\\
 \hline
Ori & / but daphne , you 're too buff / fred thinks he 's tough / and velma - wow , you 've \textcolor{blue}{lost} weight !  & Negative \\
Adv & / but daphne , you 're too buff / fred thinks he 's tough / and velma - wow , you 've \textcolor{red}{corrected} weight !  & Positive \\
\hline
Ori & The car was \textcolor{blue}{driven} by John to Maine. &  Acceptable \\
Adv & The car was \textcolor{red}{amounted} by John to Maine. & Unacceptable
\\
\hline 
Ori & The sailors \textcolor{blue}{rode} the breeze clear of the rocks. & Acceptable \\
Adv & The sailors \textcolor{red}{wandered} the breeze clear of the rocks. & Unacceptable
\\
\hline
Ori & The more Fred is obnoxious, the less \textcolor{blue}{attention} you should pay to him. & Acceptable \\
Adv & The more Fred is obnoxious, the less \textcolor{red}{noticed} you should pay to him. & Unacceptable
\\
\hline
Ori & Sentence1: And, despite its own suggestions to the contrary, Oracle will sell PeopleSoft and JD Edwards financial software through reseller channels to new customers.<SPLIT>Sentence2: Oracle sells \textcolor{blue}{financial} software. & Not\_entailment \\
Adv & Sentence1: And, despite its own suggestions to the contrary, Oracle will sell PeopleSoft and JD Edwards financial software through reseller channels to new customers.<SPLIT>Sentence2: Oracle sells \textcolor{red}{another} software. & Entailment
\\
\hline
Ori & Sentence1: Ms Stewart , the chief executive , was not expected to \textcolor{blue}{attend} .<SPLIT>Sentence2: Ms Stewart , 61 , its chief executive officer and chairwoman , did not attend . & Equivalent\\
Adv & Sentence1: Ms Stewart , the chief executive , was not expected to \textcolor{red}{visiting} .<SPLIT>Sentence2: Ms Stewart , 61 , its chief executive officer and chairwoman , did not attend . & Not\_equivalent 
\\
\hline
Ori & Sentence1: Sen. Patrick Leahy of Vermont , the committee 's senior Democrat , later said the problem is serious but called Hatch 's suggestion too drastic .<SPLIT>Sentence2: Sen. Patrick Leahy , the committee 's senior Democrat , later said the problem is serious but called Hatch 's idea too drastic a remedy to be \textcolor{blue}{considered} . & Equivalent \\
Adv & Sentence1: Sen. Patrick Leahy of Vermont , the committee 's senior Democrat , later said the problem is serious but called Hatch 's suggestion too drastic .<SPLIT>Sentence2: Sen. Patrick Leahy , the committee 's senior Democrat , later said the problem is serious but called Hatch 's idea too drastic a remedy to be \textcolor{red}{counted} . & Not\_equivalent 
\\
\hline
\end{tabular}
}
\label{tab: case study}
\end{table*}

\section{Results Visualization Across Different Prompts} 
\label{sec: prompts_vis}
We display the individual attack performance of five runs with different prompts on the MRPC dataset in Figure~\ref{fig: multi_prompts}. The figure illustrates that DA$^3$ consistently surpasses other baseline methods for each run.

\section{Observation Experiments} 
\label{observation}
\begin{figure*}[t]
     \centering
     \begin{subfigure}[t]{0.45\columnwidth}
         \centering
         \includegraphics[width=\textwidth]{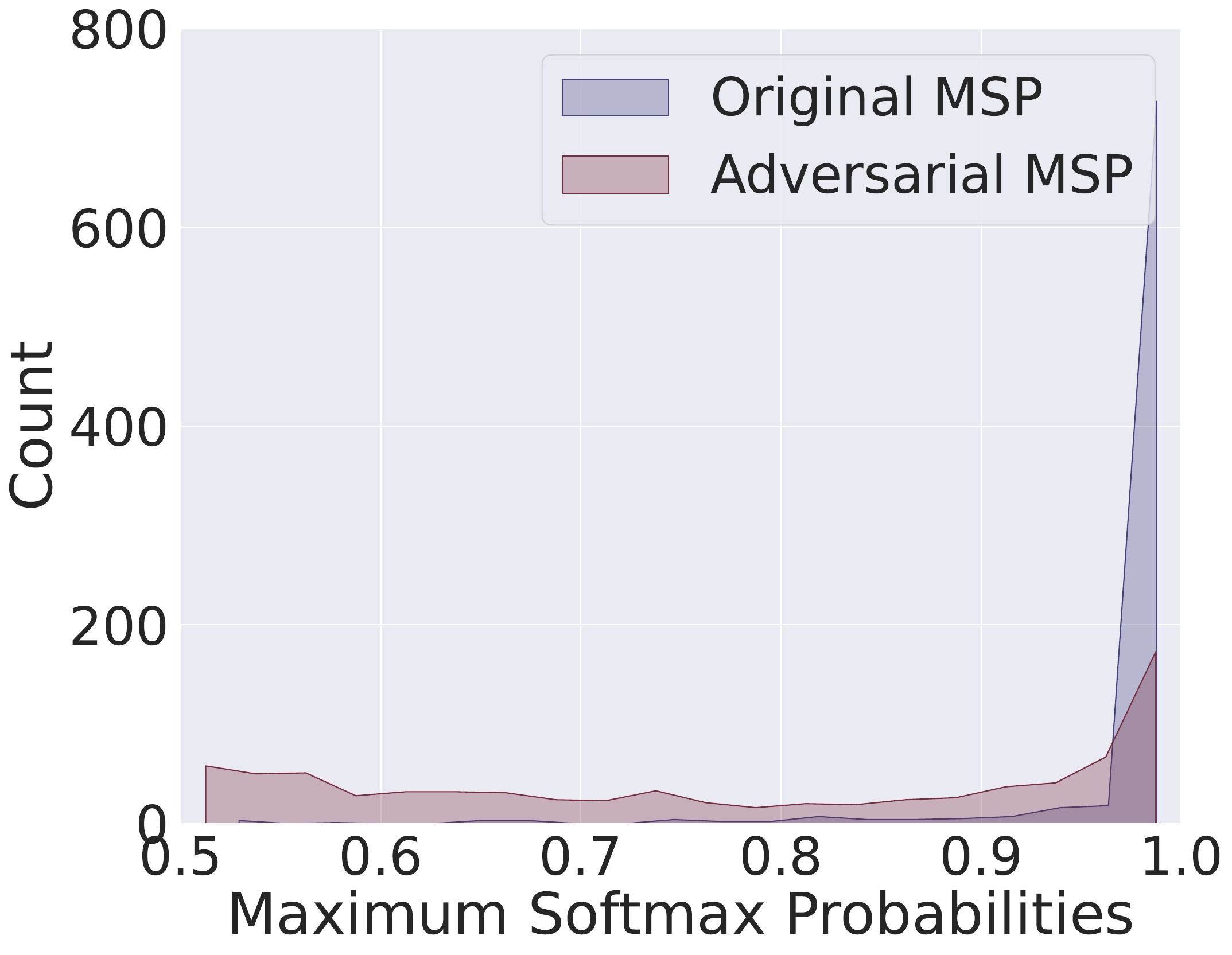}
         \caption{MSP on SST-2 dataset.}
     \end{subfigure}    
     \hfill
     \begin{subfigure}[t]{0.45\columnwidth}
         \centering
         \includegraphics[width=\textwidth]{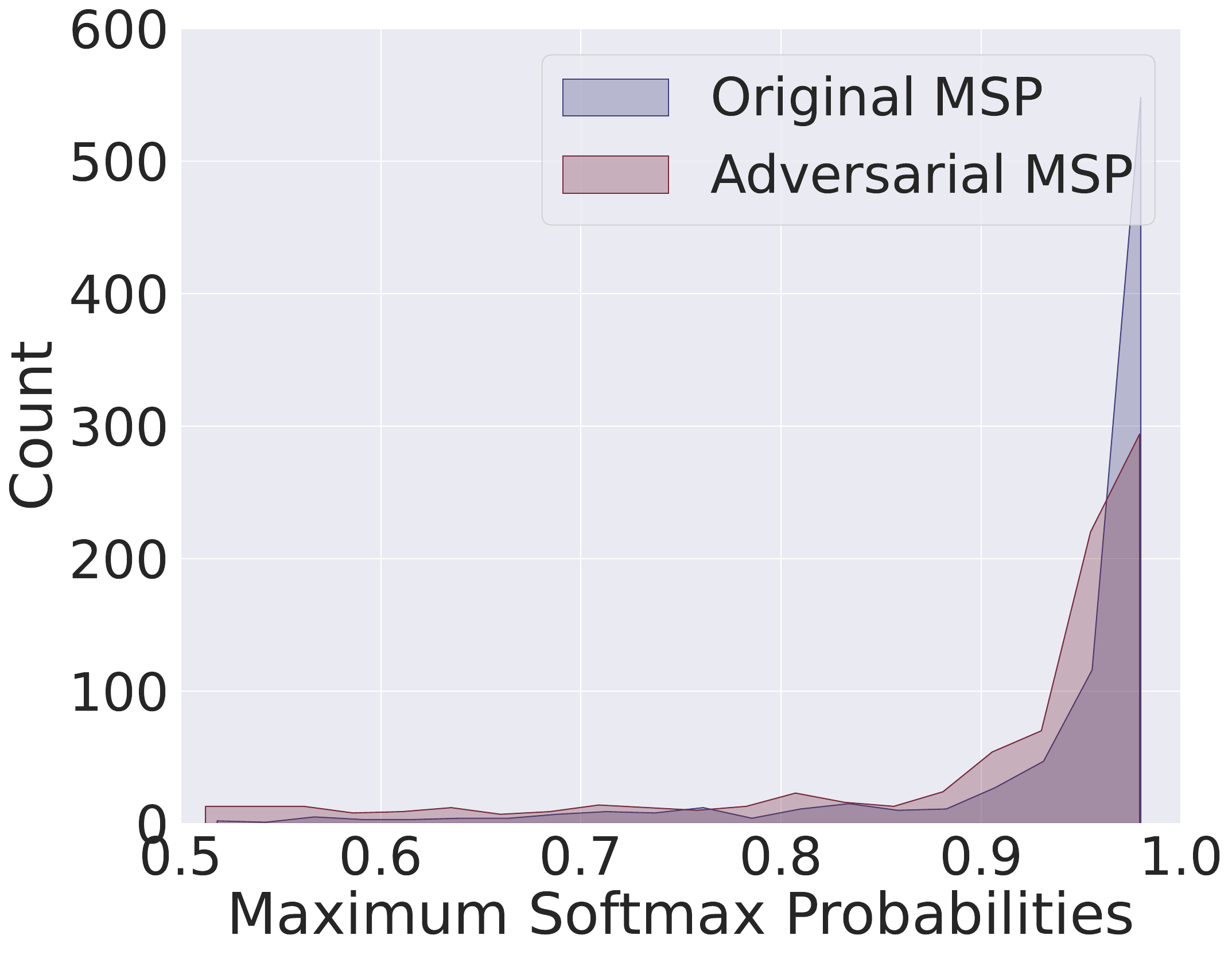}
         \caption{MSP on CoLA dataset.}
     \end{subfigure}    
     \hfill     
     \begin{subfigure}[t]{0.45\columnwidth}
         \centering
         \includegraphics[width=\textwidth]{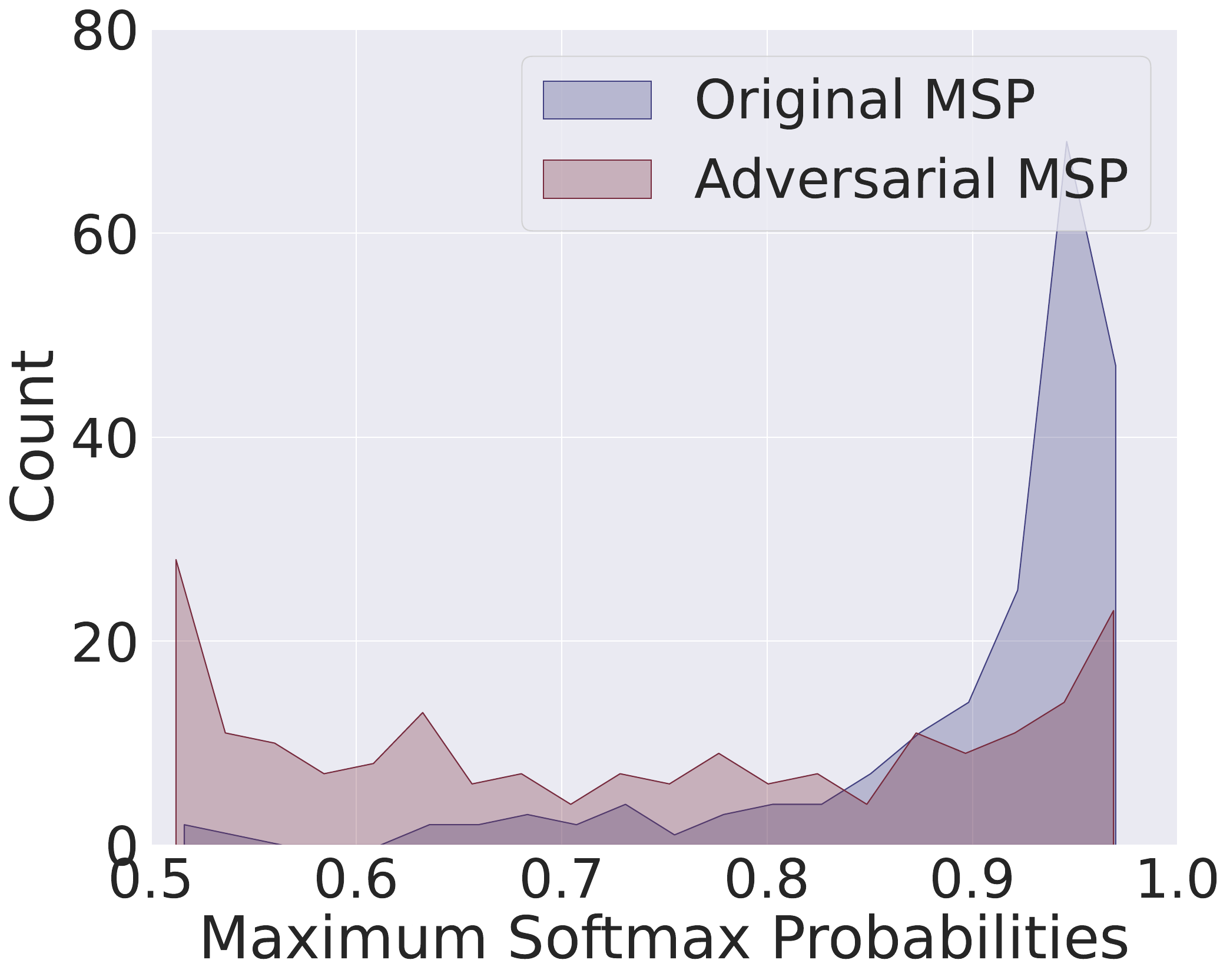}
         \caption{MSP on RTE dataset.}
     \end{subfigure}    
     \hfill
     \begin{subfigure}[t]{0.45\columnwidth}
         \centering
         \includegraphics[width=\textwidth]{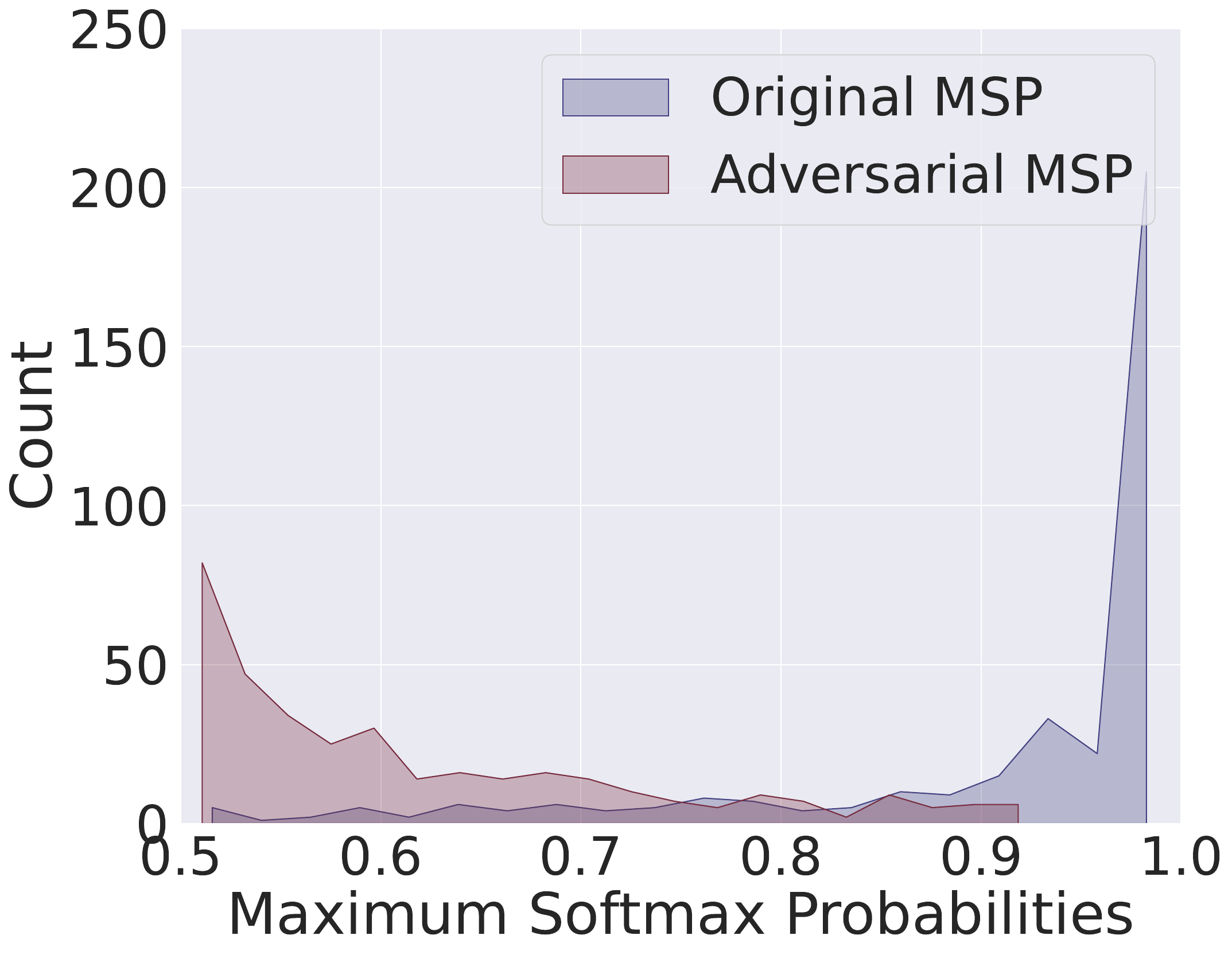}
         \caption{MSP on MRPC dataset.}
     \end{subfigure}   
    \caption{Visualization of the distribution shift between original data and adversarial data generated by TextFooler when attacking \textsc{BERT-base} regarding Maximum Softmax Probability.}
    \label{fig: textfooler_observe_MSP}
\end{figure*}

\begin{figure*}[t]
     \centering
     \begin{subfigure}[t]{0.45\columnwidth}
         \centering
         \includegraphics[width=\textwidth]{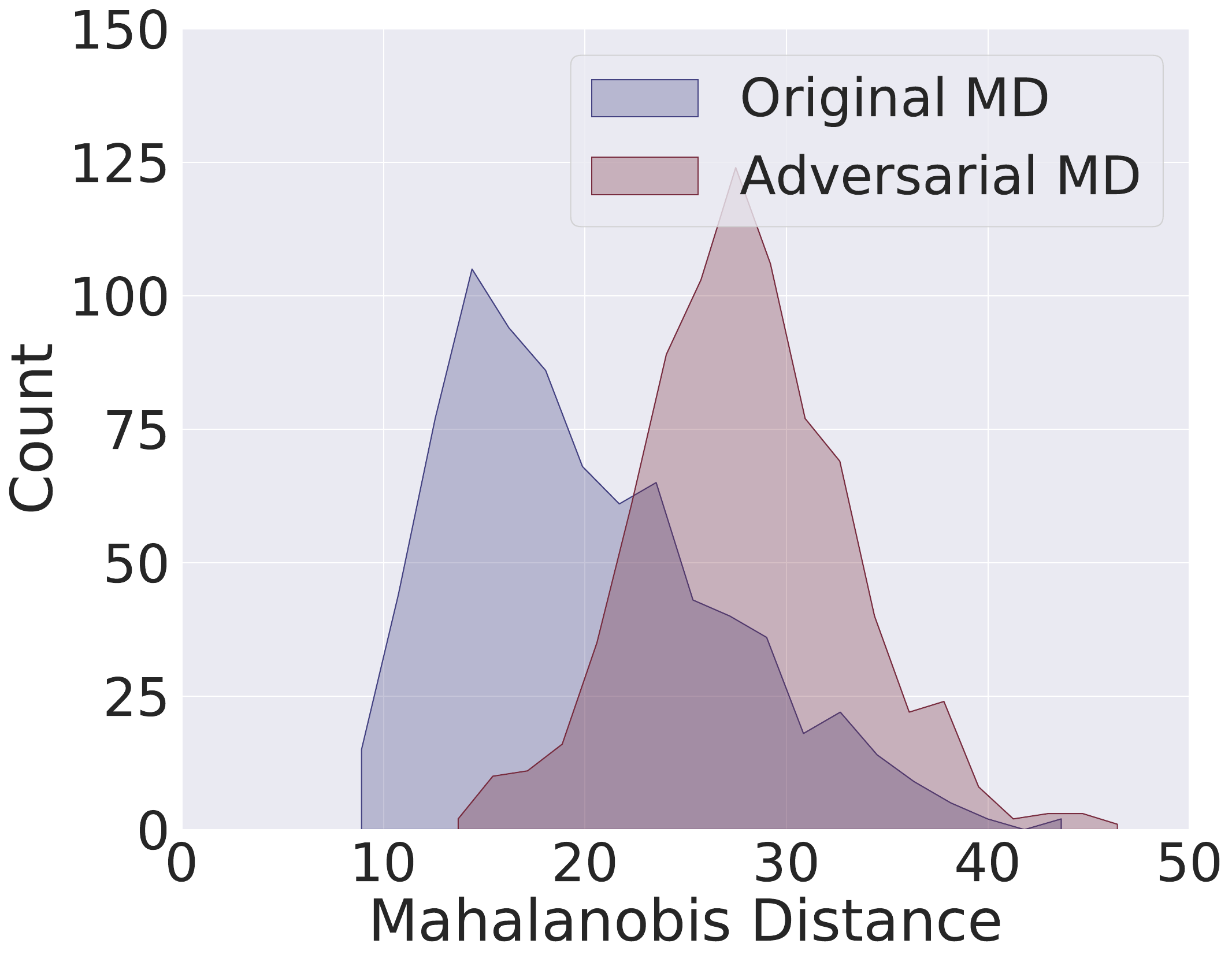}
         \caption{MD on SST-2 dataset.}
     \end{subfigure}
     \hfill
     \begin{subfigure}[t]{0.45\columnwidth}
         \centering
         \includegraphics[width=\textwidth]{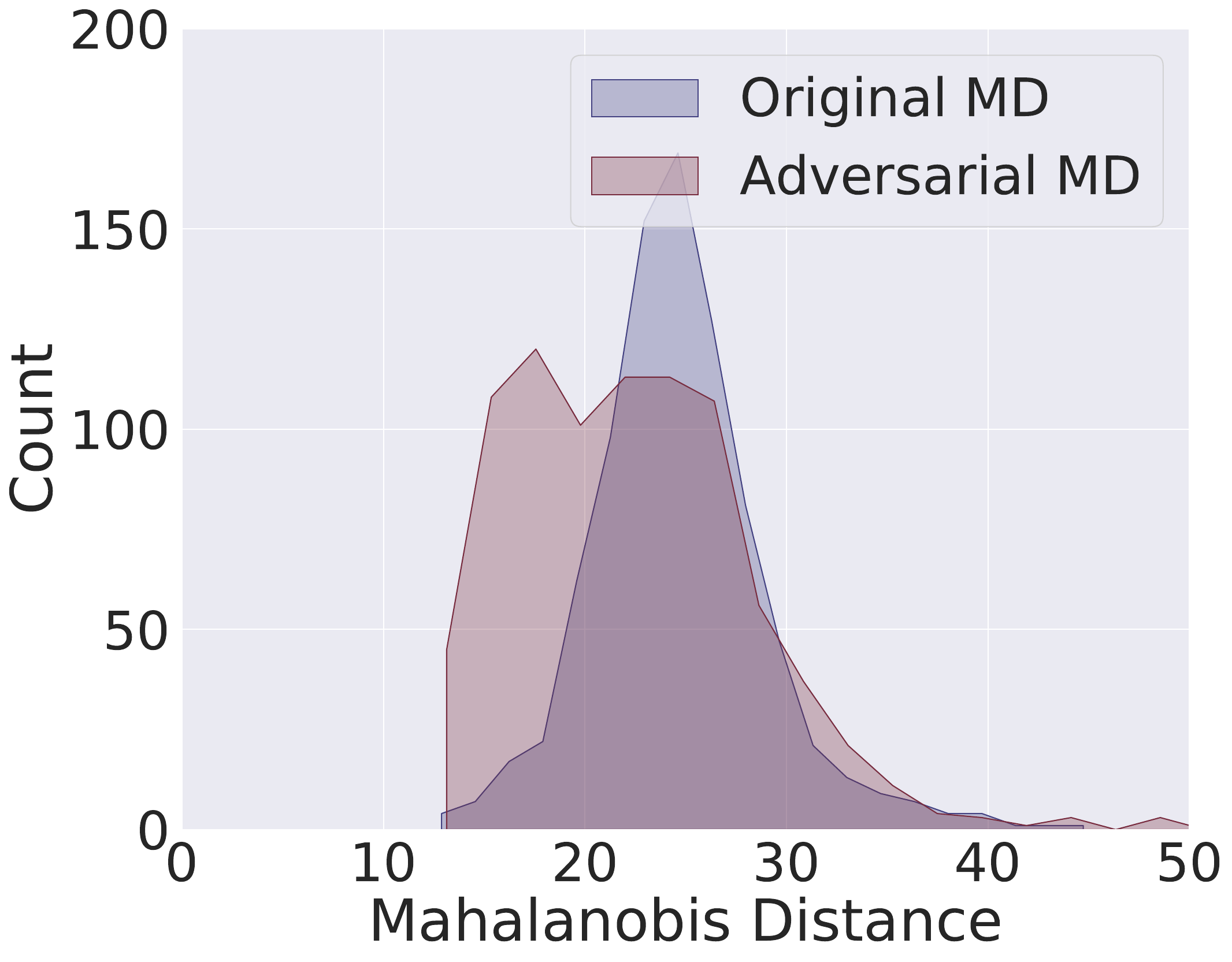}
         \caption{MD on CoLA dataset.}
     \end{subfigure}
     \hfill
     \begin{subfigure}[t]{0.45\columnwidth}
         \centering
         \includegraphics[width=\textwidth]{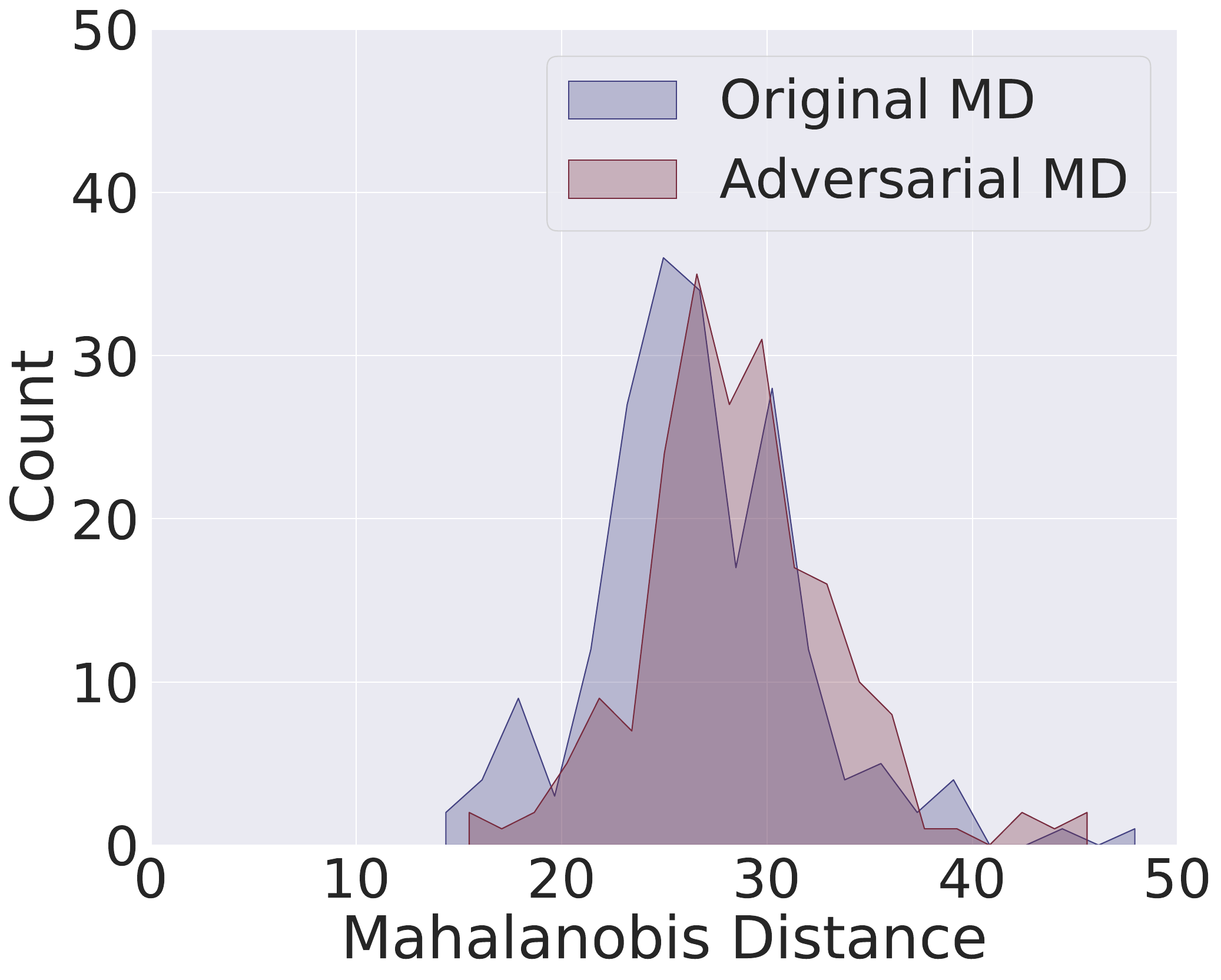}
         \caption{MD on RTE dataset.}
     \end{subfigure}
     \hfill
     \begin{subfigure}[t]{0.45\columnwidth}
         \centering
         \includegraphics[width=\textwidth]{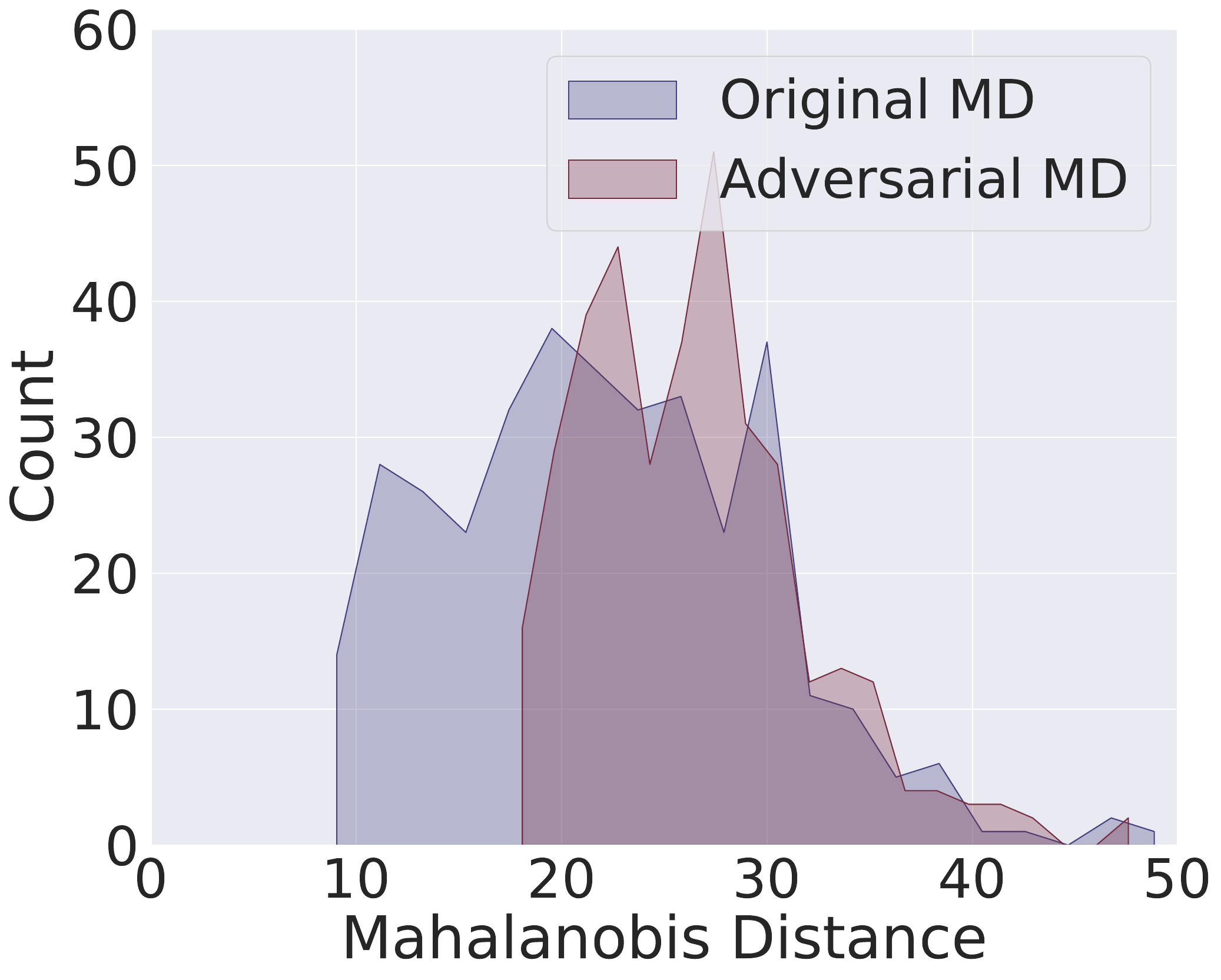}
         \caption{MD on MRPC dataset.}
     \end{subfigure}
    \caption{Visualization of the distribution shift between original data and adversarial data generated by TextFooler when attacking \textsc{BERT-base} regarding Mahalanobis Distance.}
    \label{fig: textfooler_observe_MD}
\end{figure*}

\begin{figure*}[t]
     \centering
     \begin{subfigure}[t]{0.45\columnwidth}
         \centering
         \includegraphics[width=\textwidth]{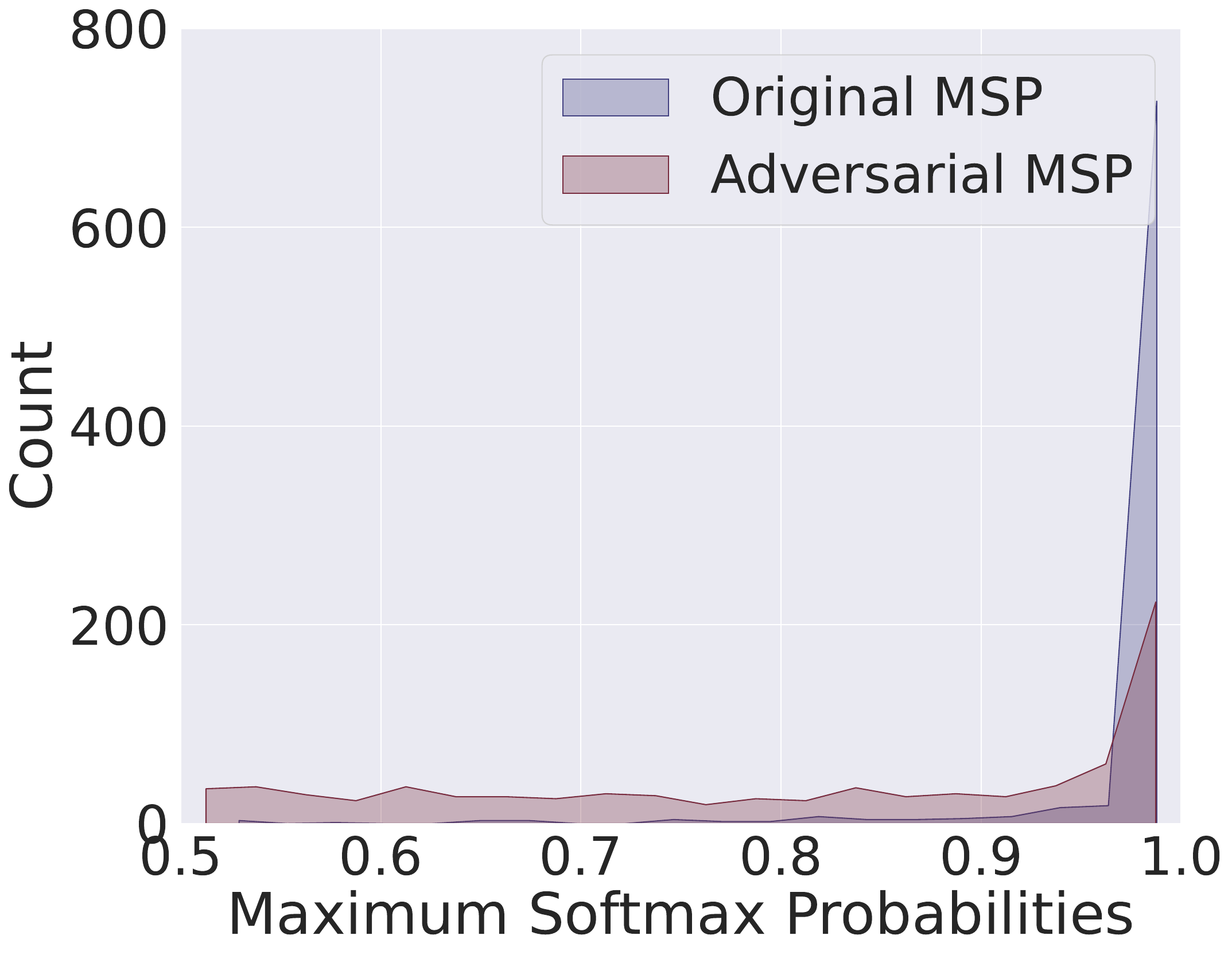}
         \caption{MSP on SST-2 dataset.}
     \end{subfigure}    
     \hfill
     \begin{subfigure}[t]{0.45\columnwidth}
         \centering
         \includegraphics[width=\textwidth]{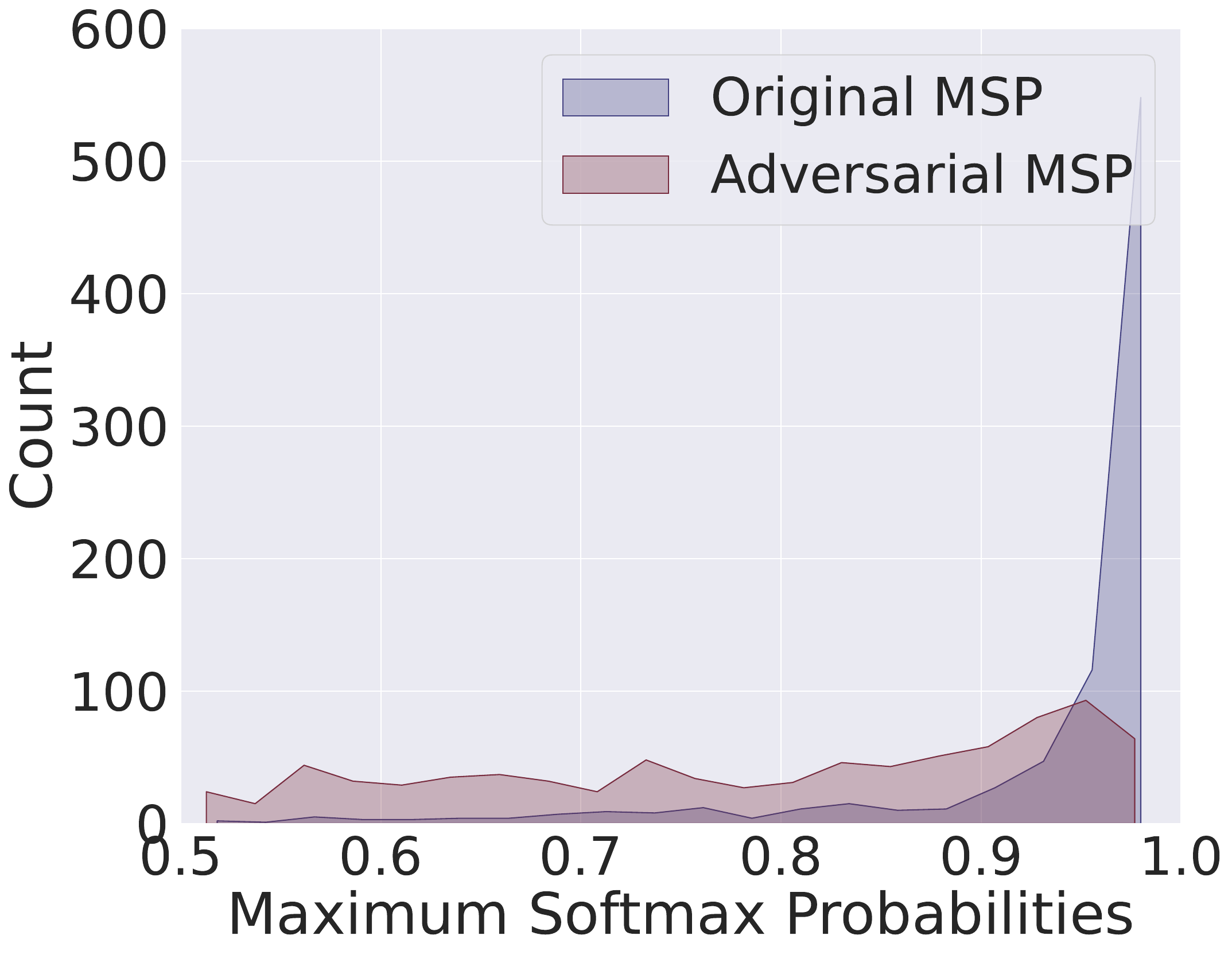}
         \caption{MSP on CoLA dataset.}
     \end{subfigure}    
     \hfill     
     \begin{subfigure}[t]{0.45\columnwidth}
         \centering
         \includegraphics[width=\textwidth]{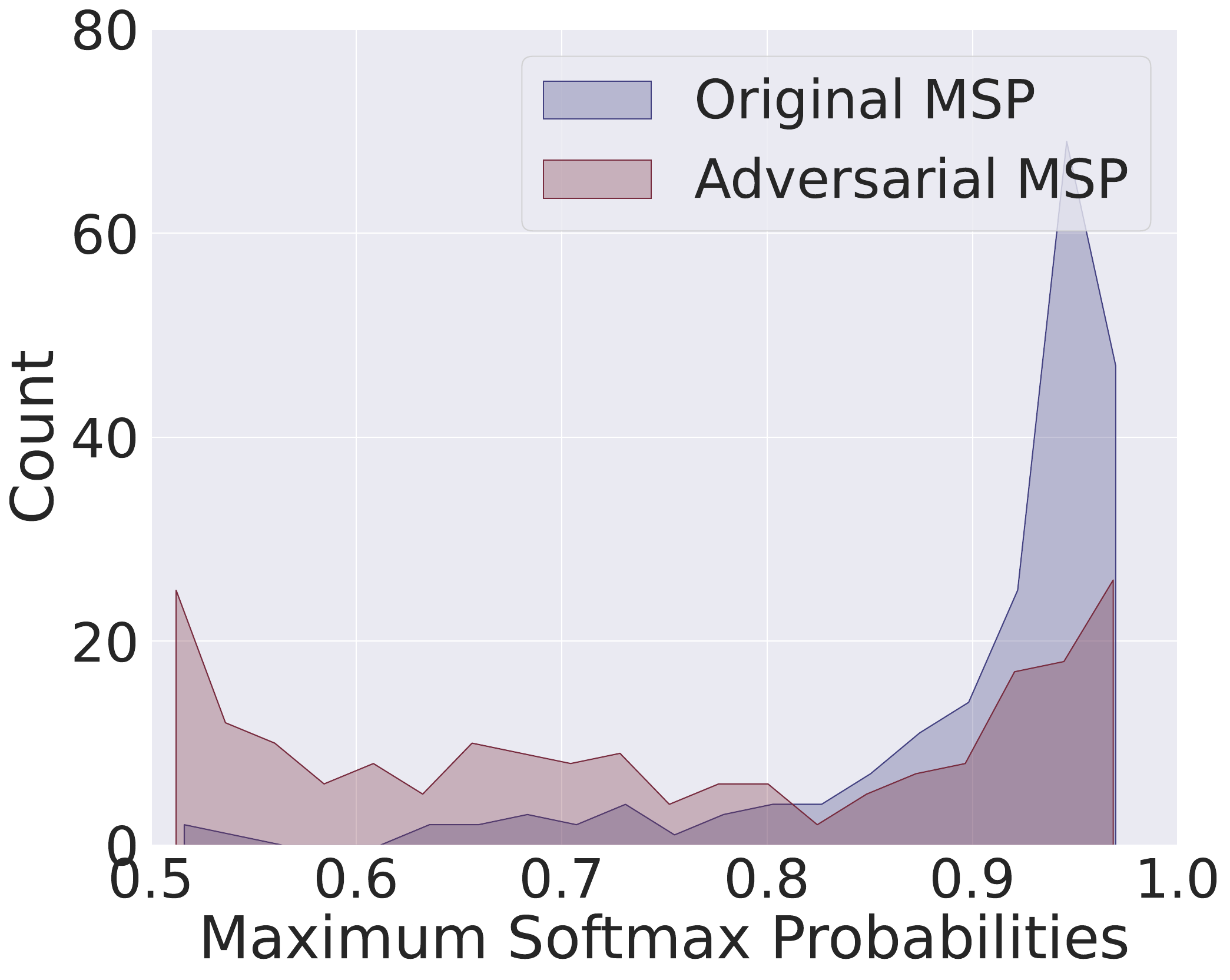}
         \caption{MSP on RTE dataset.}
     \end{subfigure}    
     \hfill
     \begin{subfigure}[t]{0.45\columnwidth}
         \centering
         \includegraphics[width=\textwidth]{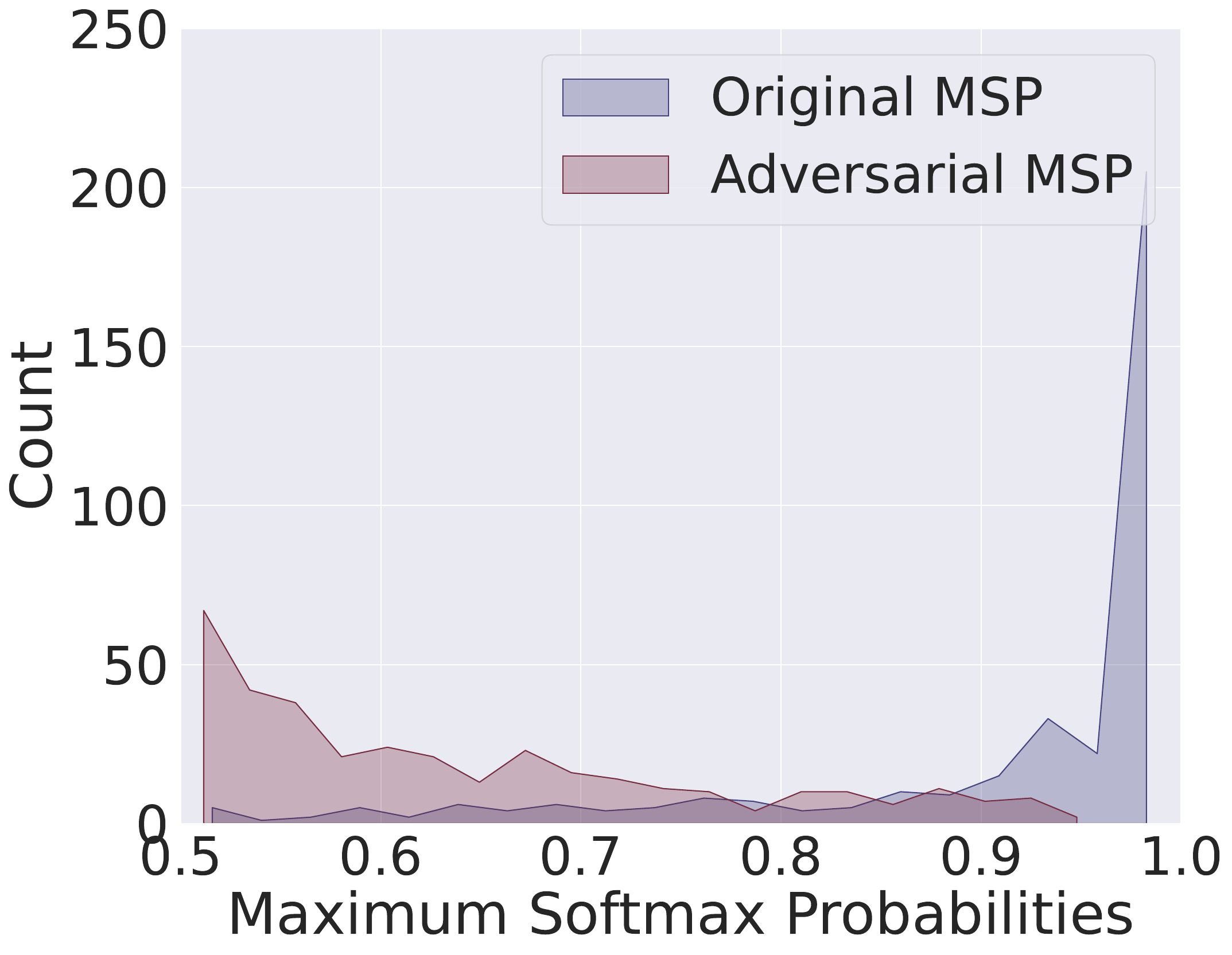}
         \caption{MSP on MRPC dataset.}
     \end{subfigure}   
    \caption{Visualization of the distribution shift between original data and adversarial data generated by TextBugger when attacking \textsc{BERT-base} regarding Maximum Softmax Probability.}
    \label{fig: textbugger_observe_MSP}
\end{figure*}

\begin{figure*}[t]
     \centering
     \begin{subfigure}[t]{0.45\columnwidth}
         \centering
         \includegraphics[width=\textwidth]{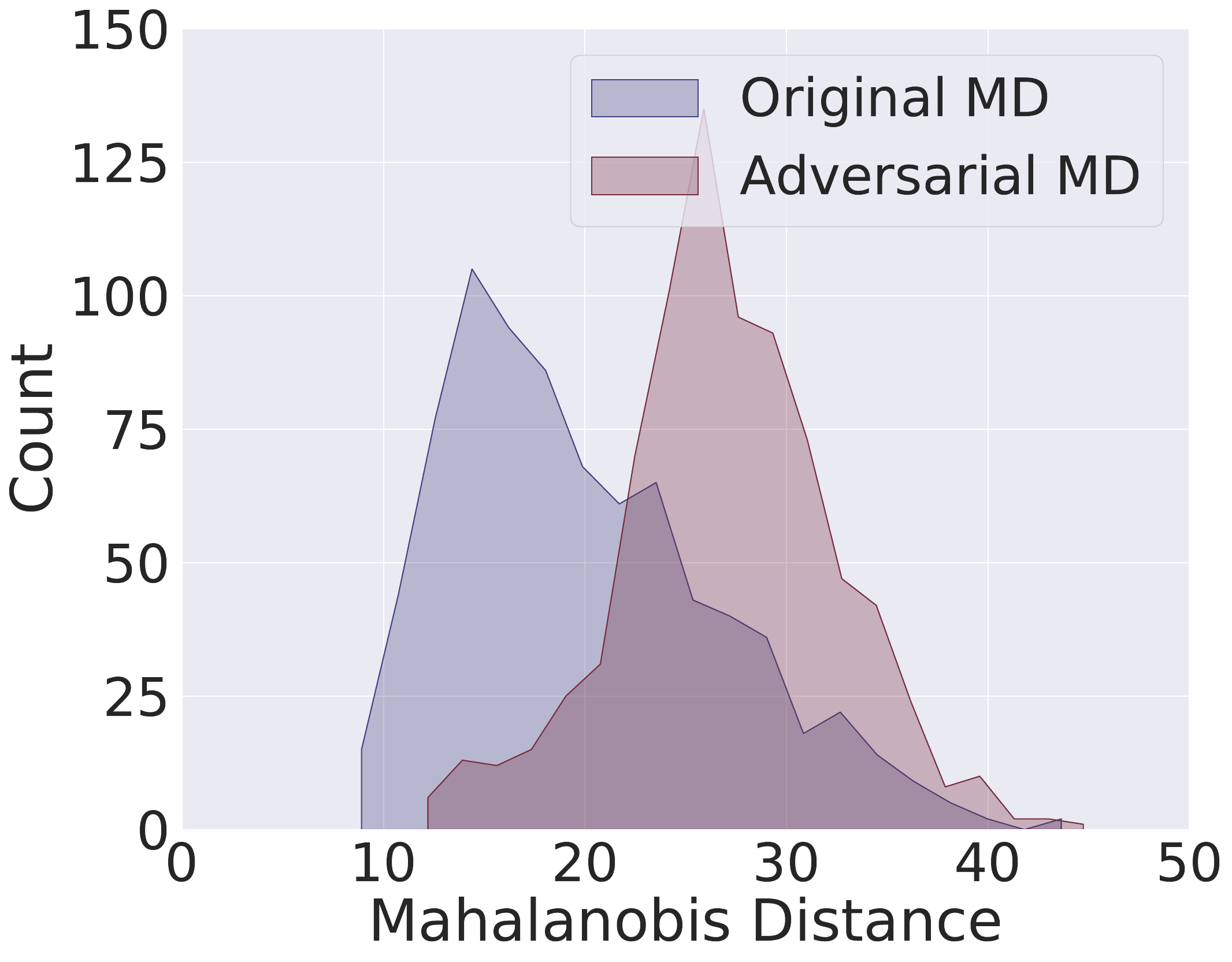}
         \caption{MD on SST-2 dataset.}
     \end{subfigure}
     \hfill
     \begin{subfigure}[t]{0.45\columnwidth}
         \centering
         \includegraphics[width=\textwidth]{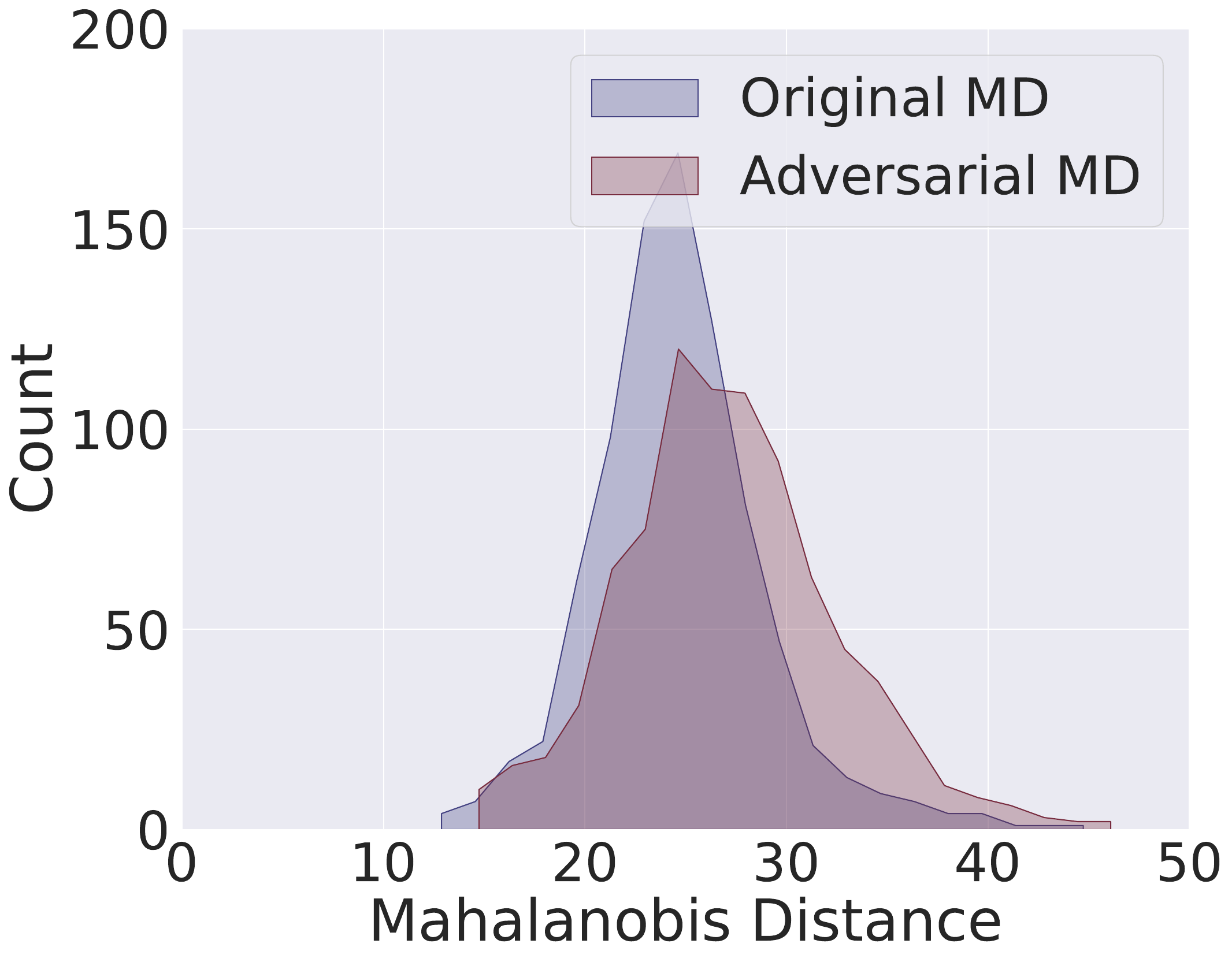}
         \caption{MD on CoLA dataset.}
     \end{subfigure}
     \hfill
     \begin{subfigure}[t]{0.45\columnwidth}
         \centering
         \includegraphics[width=\textwidth]{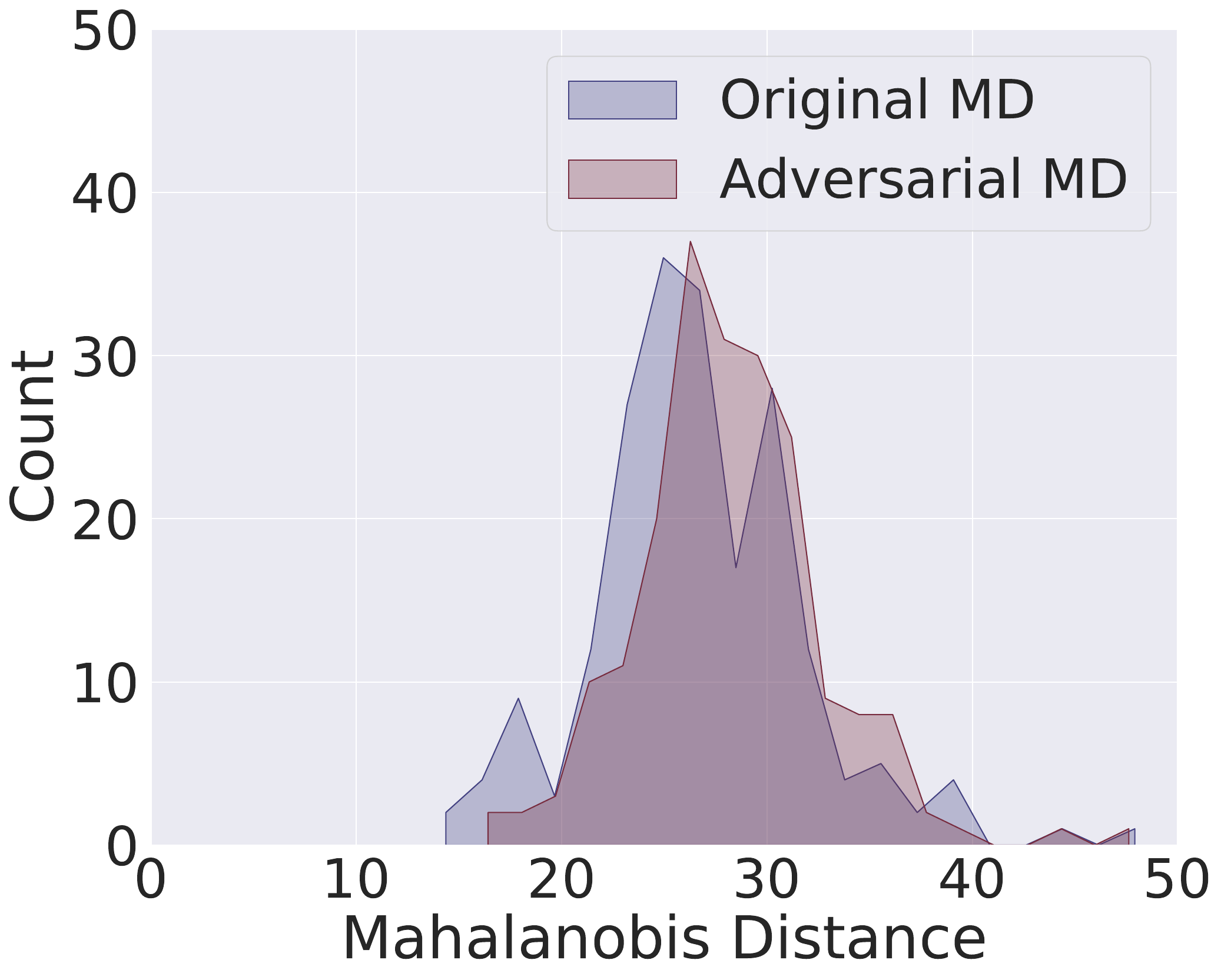}
         \caption{MD on RTE dataset.}
     \end{subfigure}
     \hfill
     \begin{subfigure}[t]{0.45\columnwidth}
         \centering
         \includegraphics[width=\textwidth]{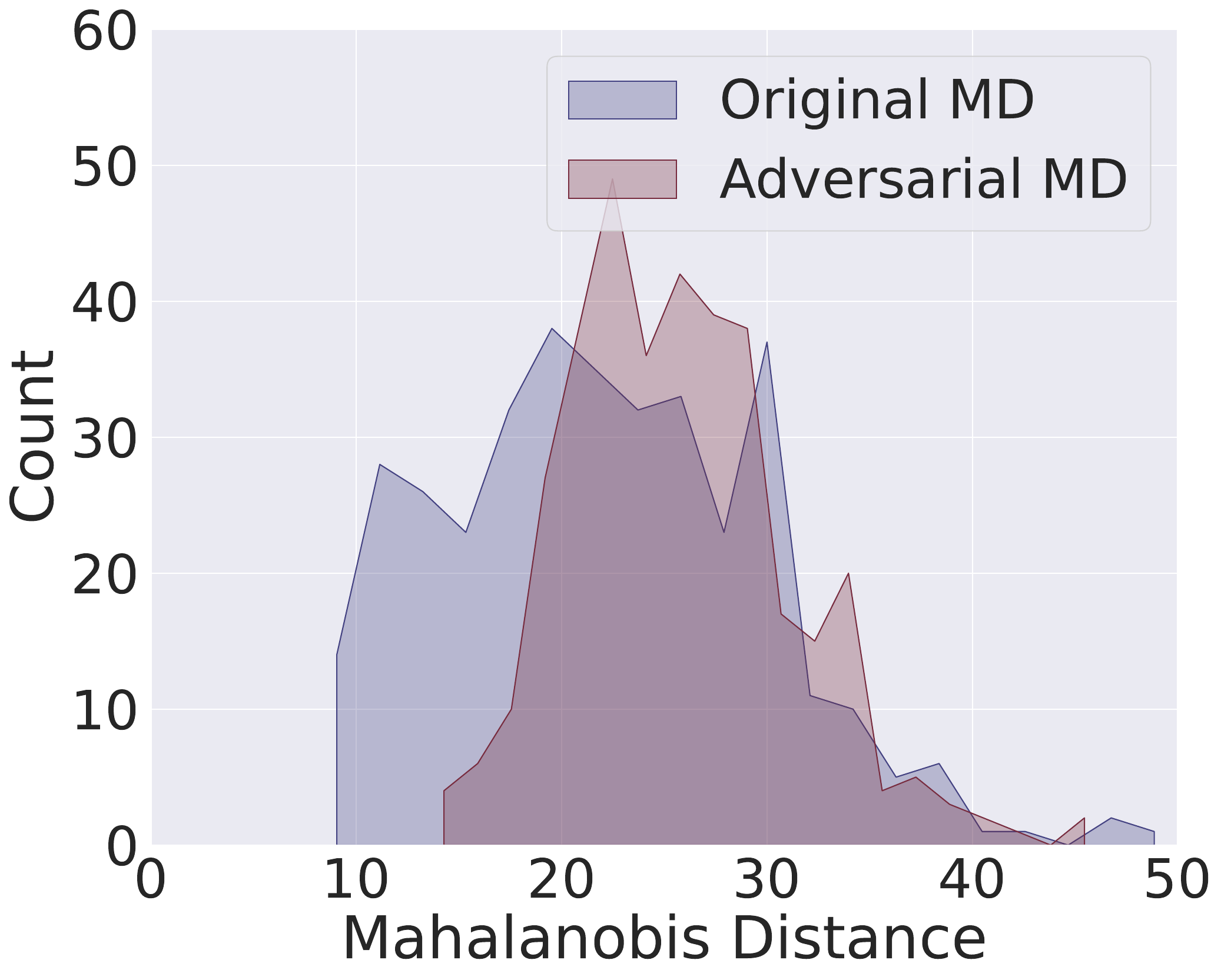}
         \caption{MD on MRPC dataset.}
     \end{subfigure}
    \caption{Visualization of the distribution shift between original data and adversarial data generated by TextBugger when attacking \textsc{BERT-base} regarding Mahalanobis Distance.}
    \label{fig: textbugger_observe_MD}
\end{figure*}

\begin{figure*}[t]
     \centering
     \begin{subfigure}[t]{0.45\columnwidth}
         \centering
         \includegraphics[width=\textwidth]{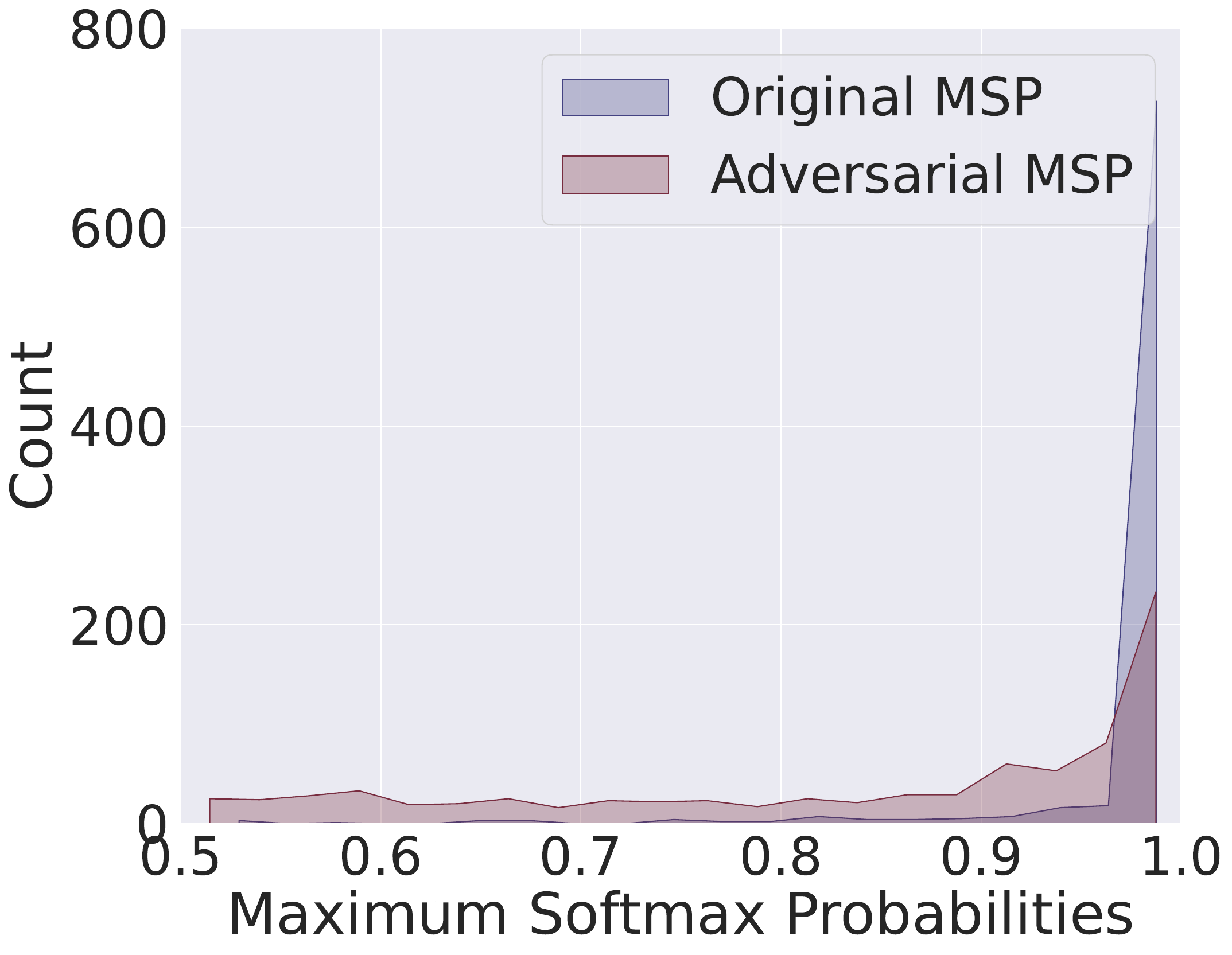}
         \caption{MSP on SST-2 dataset.}
     \end{subfigure}    
     \hfill
     \begin{subfigure}[t]{0.45\columnwidth}
         \centering
         \includegraphics[width=\textwidth]{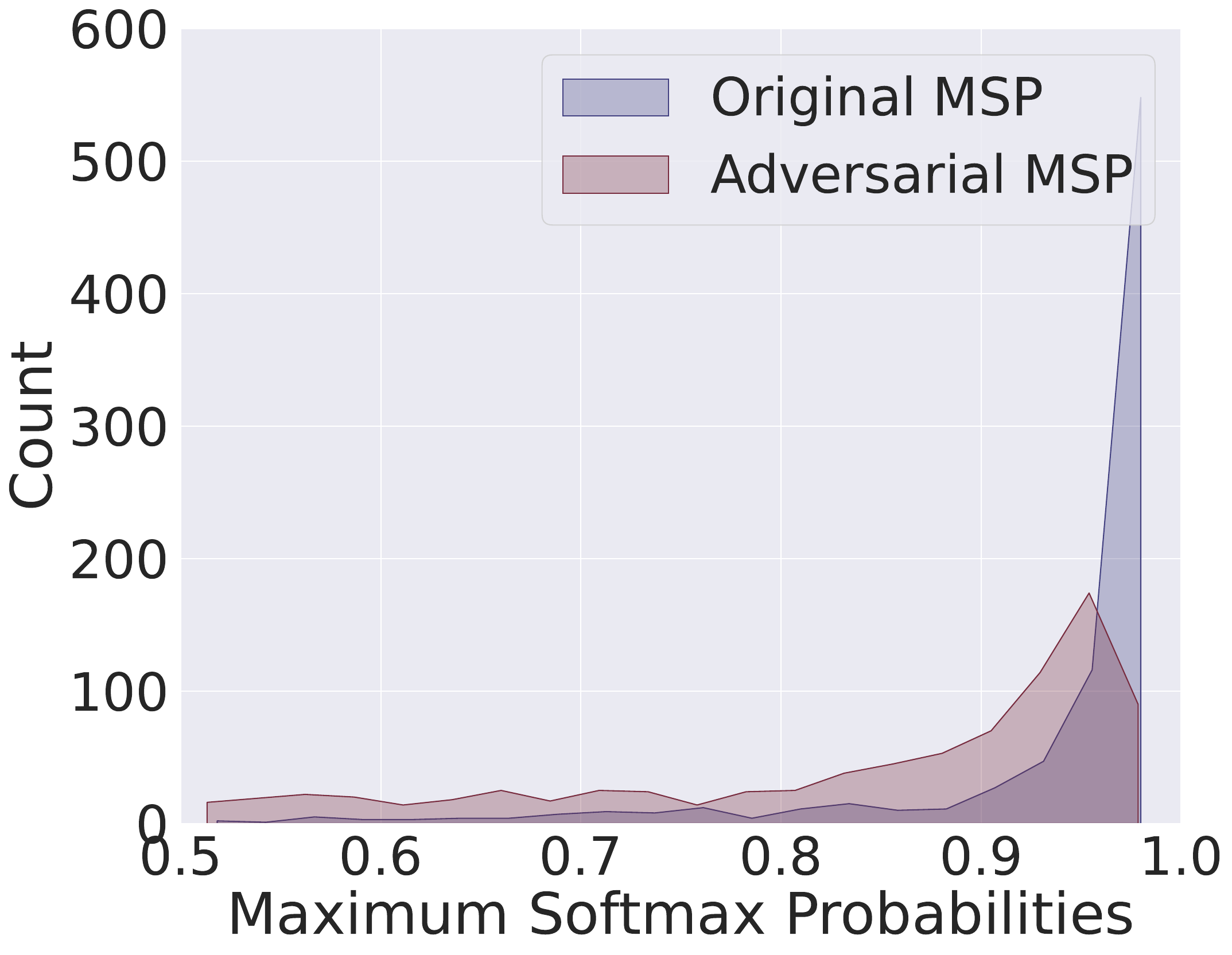}
         \caption{MSP on CoLA dataset.}
     \end{subfigure}    
     \hfill     
     \begin{subfigure}[t]{0.45\columnwidth}
         \centering
         \includegraphics[width=\textwidth]{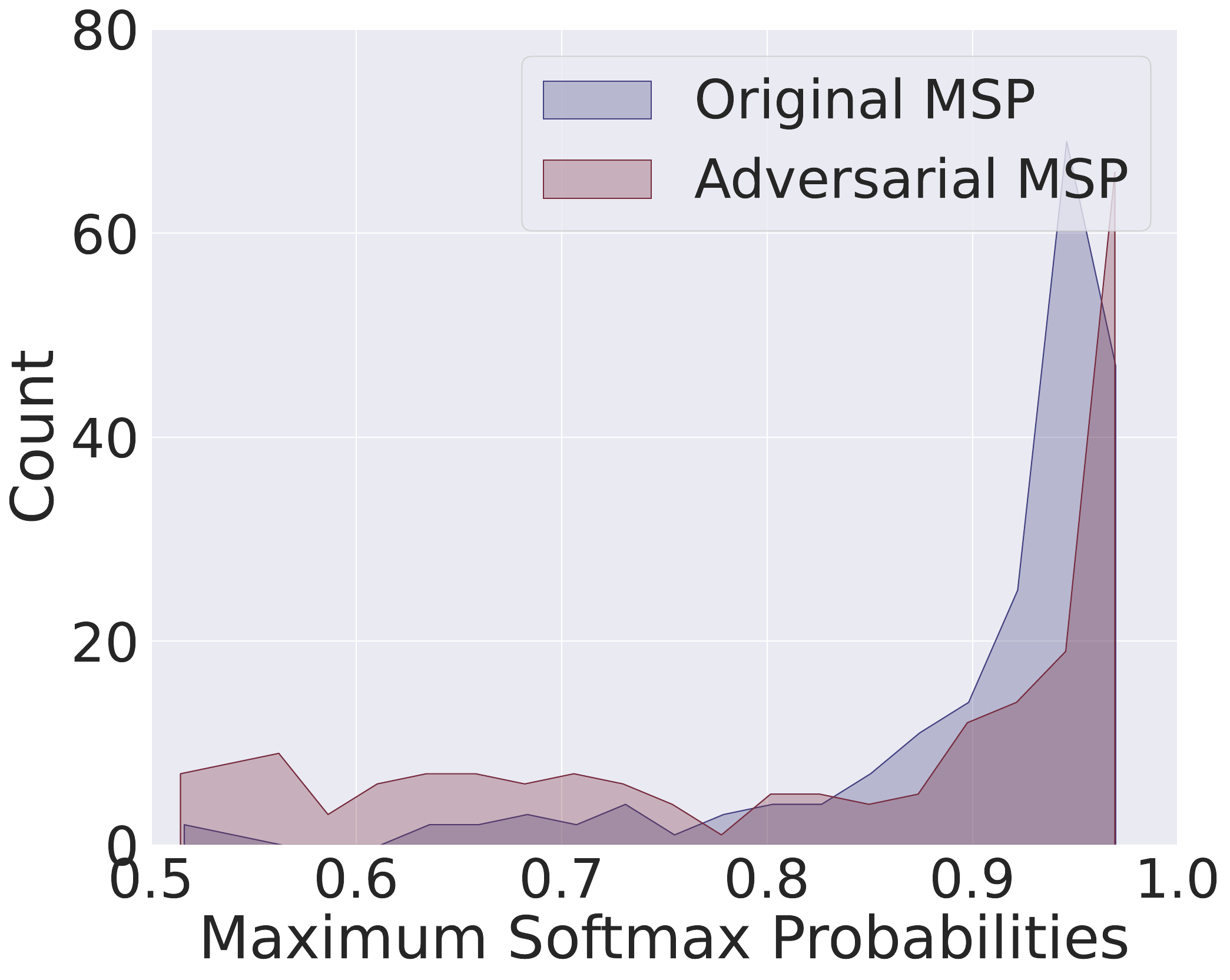}
         \caption{MSP on RTE dataset.}
     \end{subfigure}    
     \hfill
     \begin{subfigure}[t]{0.45\columnwidth}
         \centering
         \includegraphics[width=\textwidth]{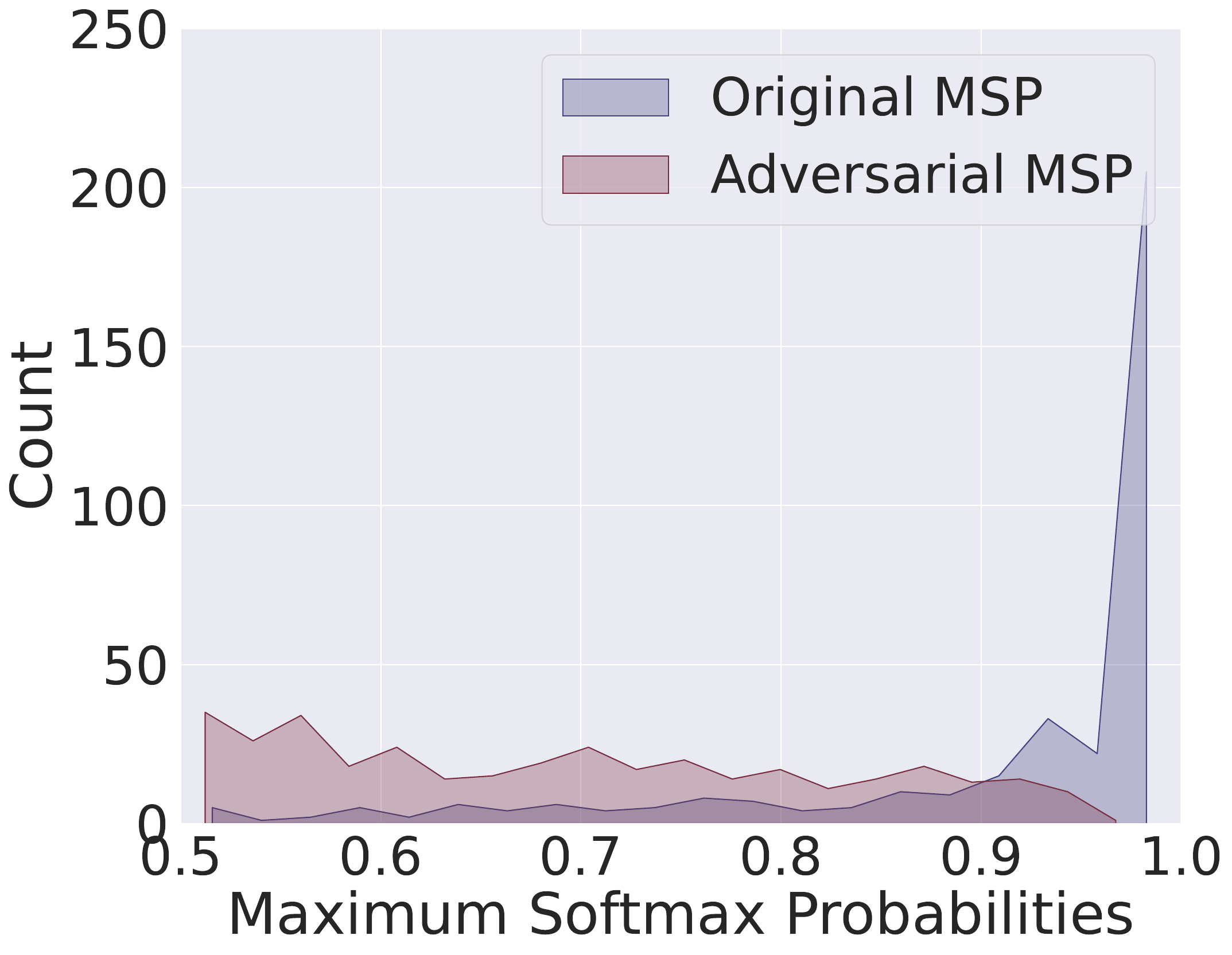}
         \caption{MSP on MRPC dataset.}
     \end{subfigure}   
    \caption{Visualization of the distribution shift between original data and adversarial data generated by DeepWordBug when attacking \textsc{BERT-base} regarding Maximum Softmax Probability.}
    \label{fig: deepwordbug_observe_MSP}
\end{figure*}

\begin{figure*}[t]
     \centering
     \begin{subfigure}[t]{0.45\columnwidth}
         \centering
         \includegraphics[width=\textwidth]{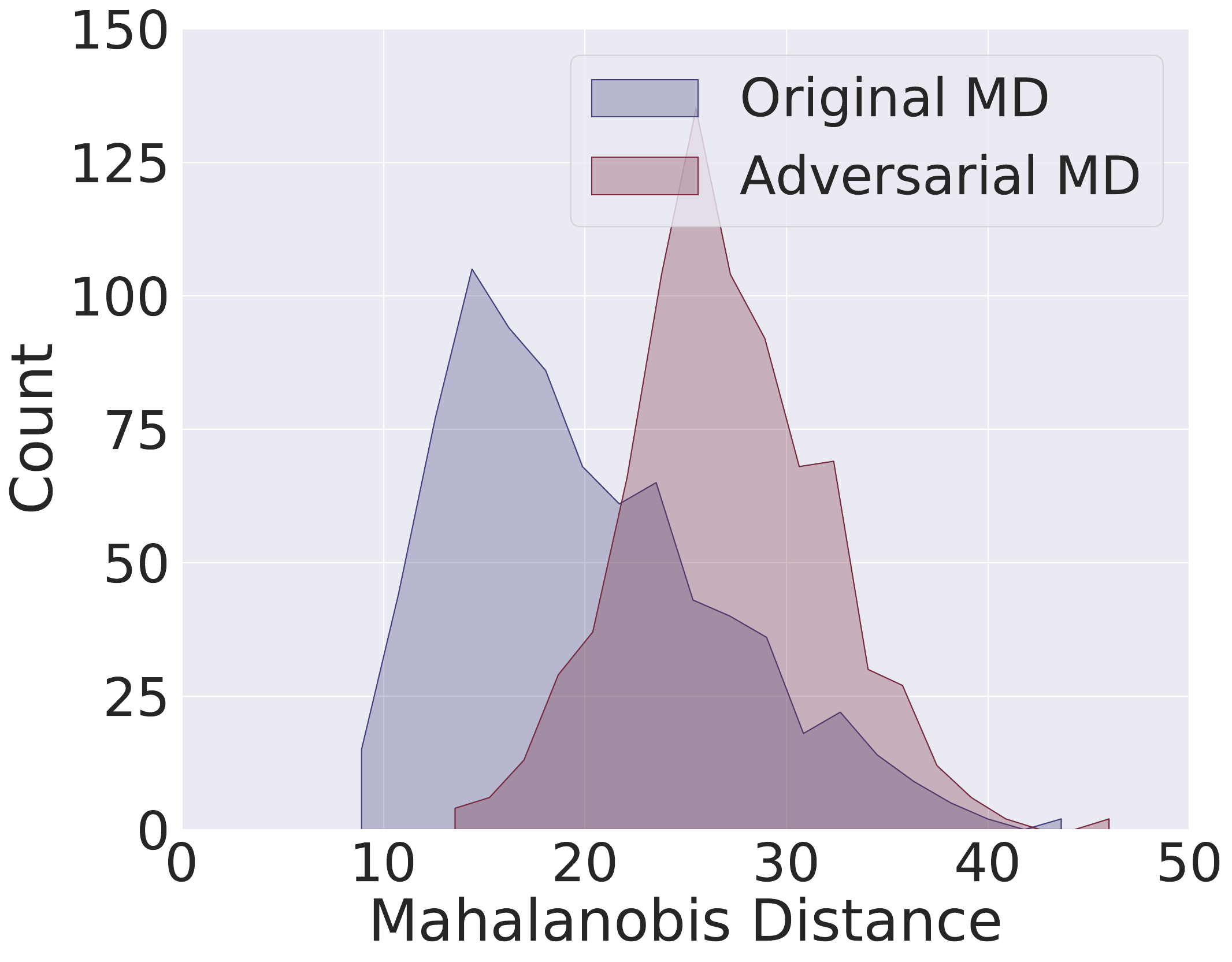}
         \caption{MD on SST-2 dataset.}
     \end{subfigure}
     \hfill
     \begin{subfigure}[t]{0.45\columnwidth}
         \centering
         \includegraphics[width=\textwidth]{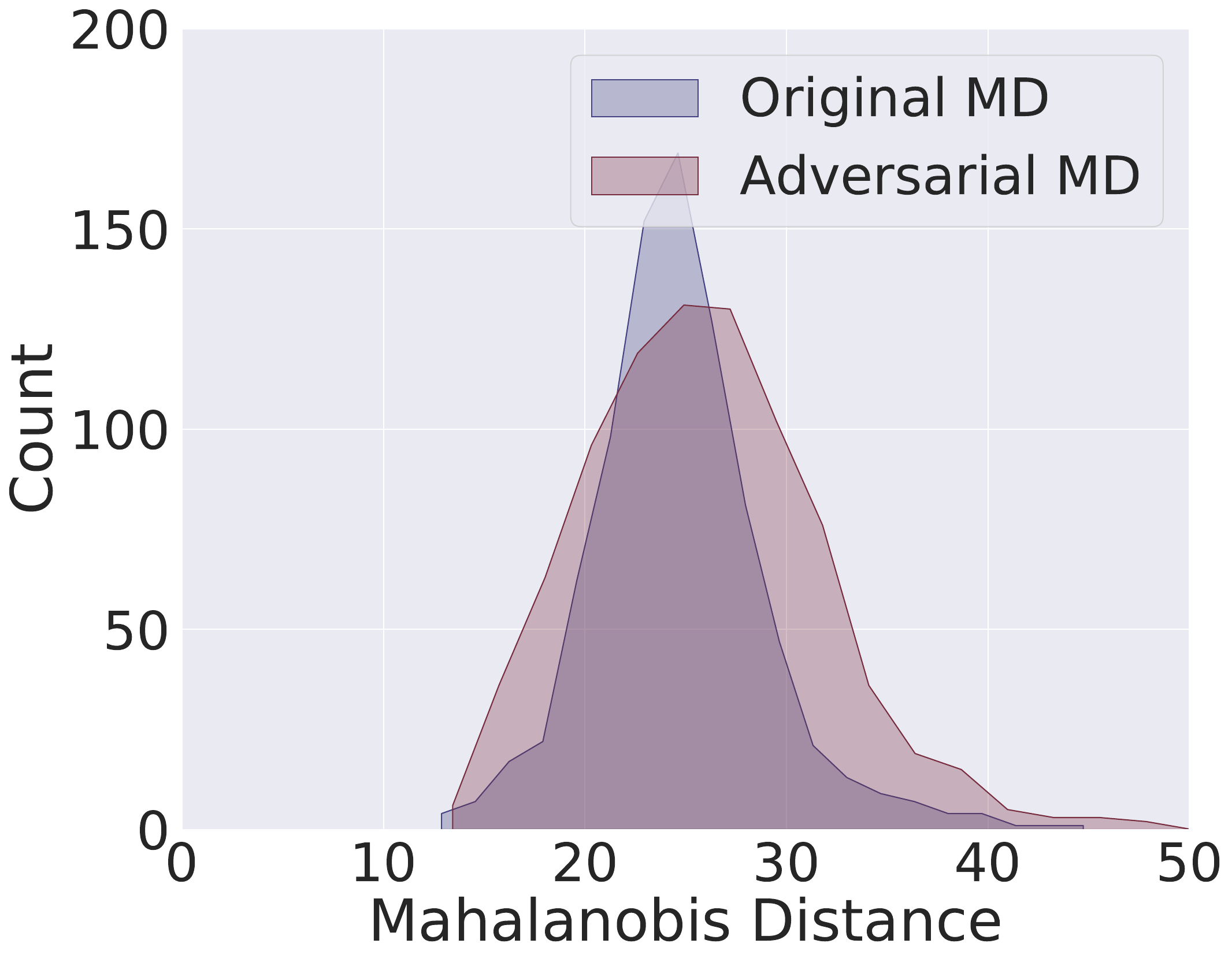}
         \caption{MD on CoLA dataset.}
     \end{subfigure}
     \hfill
     \begin{subfigure}[t]{0.45\columnwidth}
         \centering
         \includegraphics[width=\textwidth]{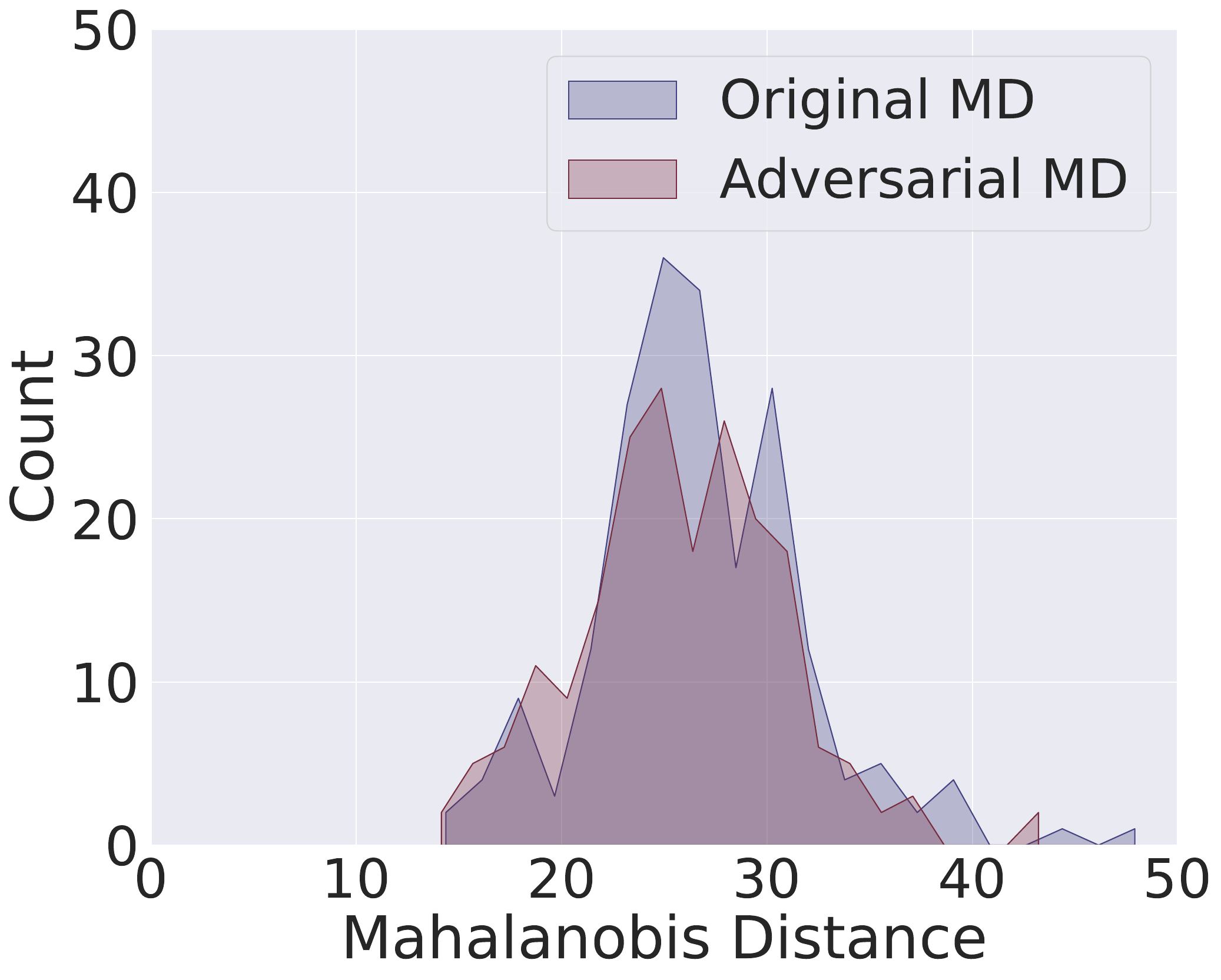}
         \caption{MD on RTE dataset.}
     \end{subfigure}
     \hfill
     \begin{subfigure}[t]{0.45\columnwidth}
         \centering
         \includegraphics[width=\textwidth]{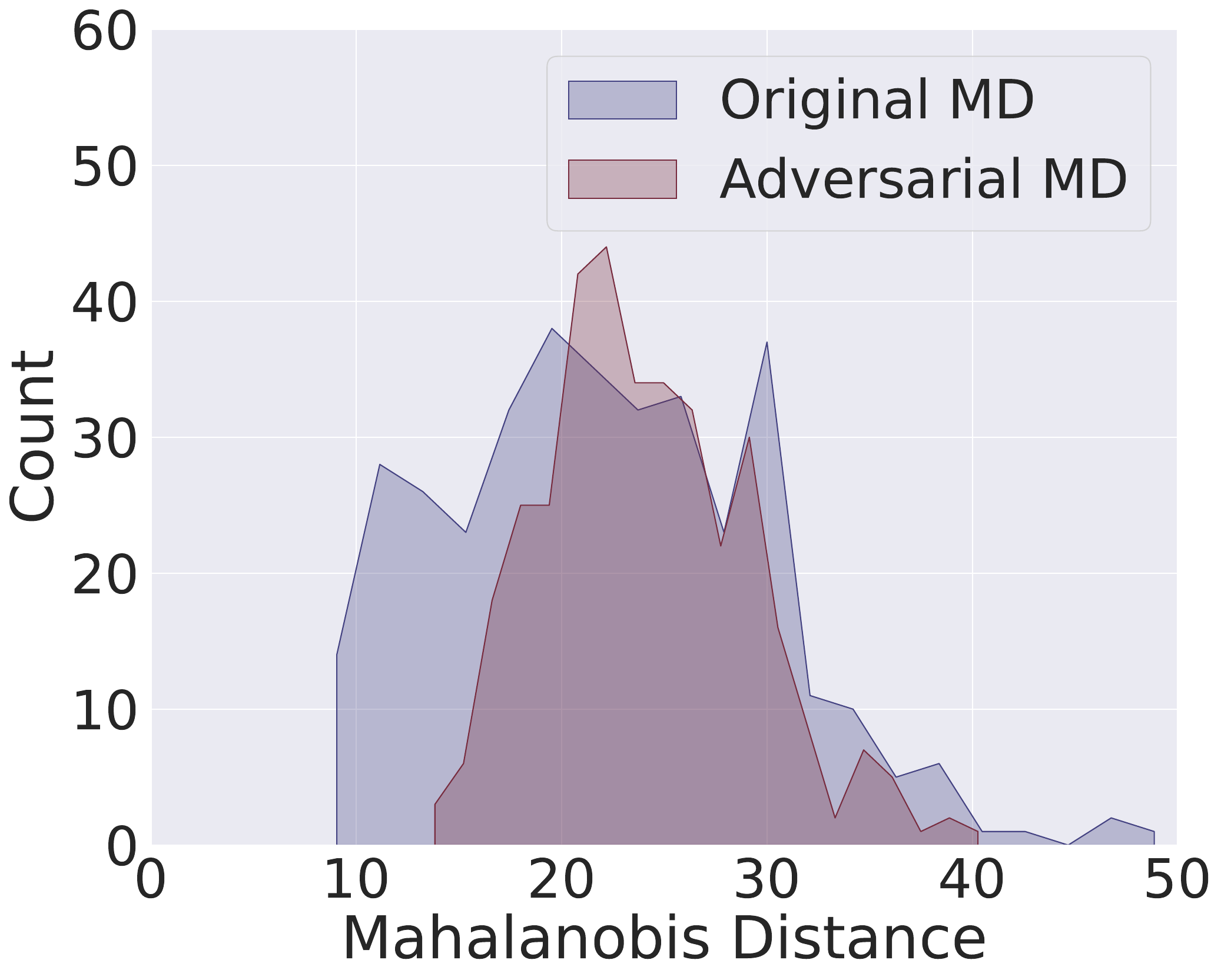}
         \caption{MD on MRPC dataset.}
     \end{subfigure}
    \caption{Visualization of the distribution shift between original data and adversarial data generated by DeepWordBug when attacking \textsc{BERT-base} regarding Mahalanobis Distance.}
    \label{fig: deepwordbug_observe_MD}
\end{figure*}

\begin{figure*}[t]
     \centering
     \begin{subfigure}[t]{0.45\columnwidth}
         \centering
         \includegraphics[width=\textwidth]{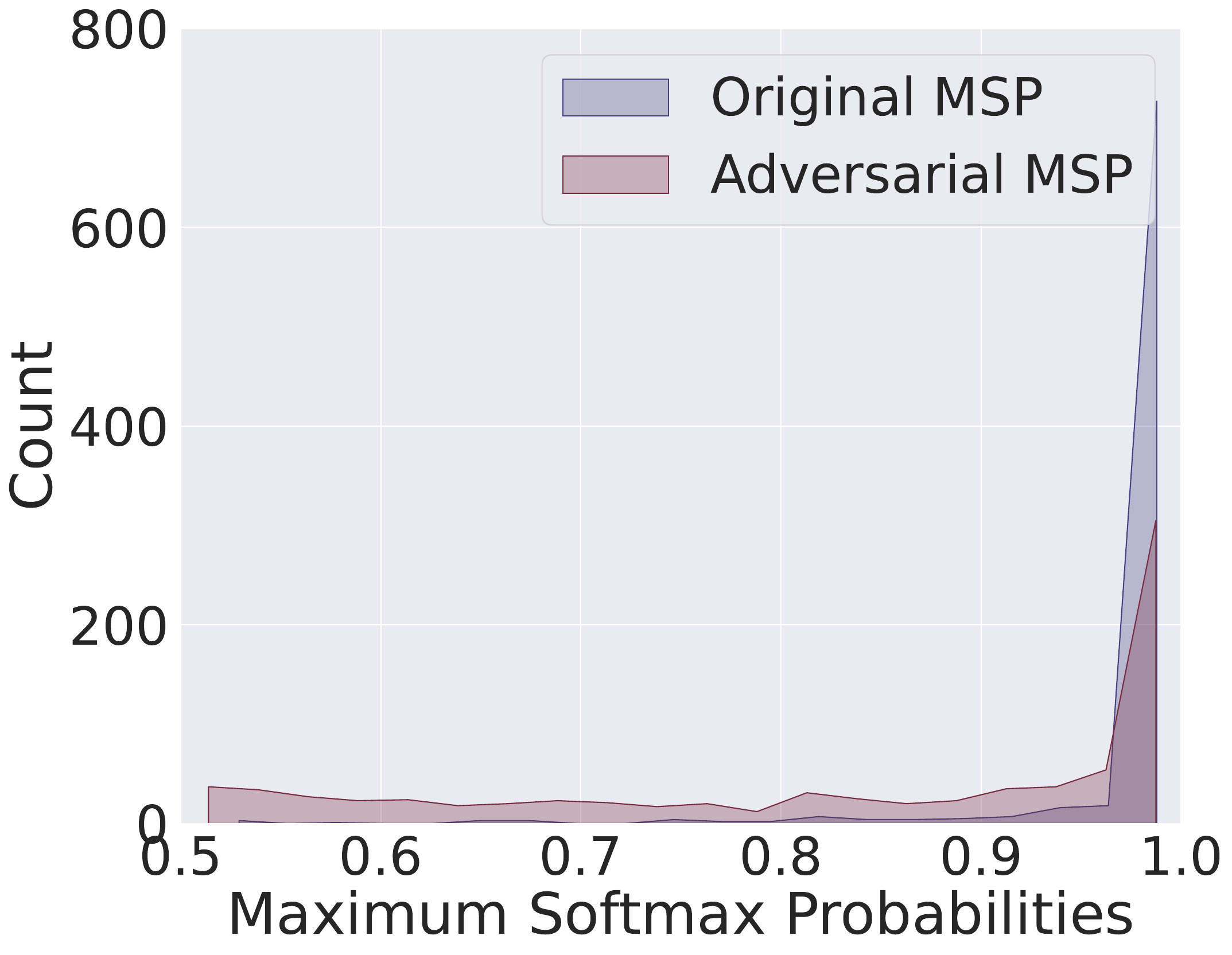}
         \caption{MSP on SST-2 dataset.}
     \end{subfigure}    
     \hfill
     \begin{subfigure}[t]{0.45\columnwidth}
         \centering
         \includegraphics[width=\textwidth]{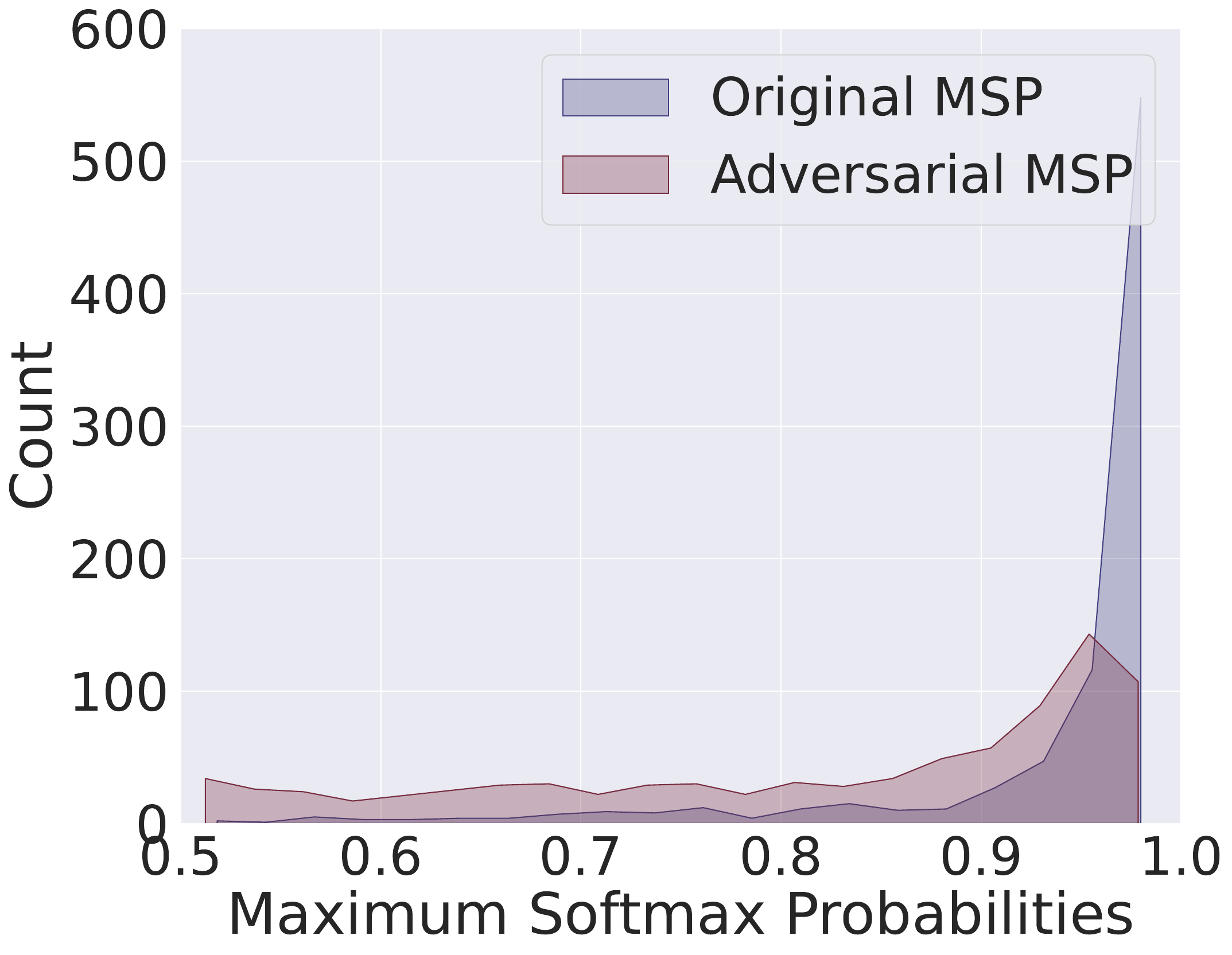}
         \caption{MSP on CoLA dataset.}
     \end{subfigure}    
     \hfill     
     \begin{subfigure}[t]{0.45\columnwidth}
         \centering
         \includegraphics[width=\textwidth]{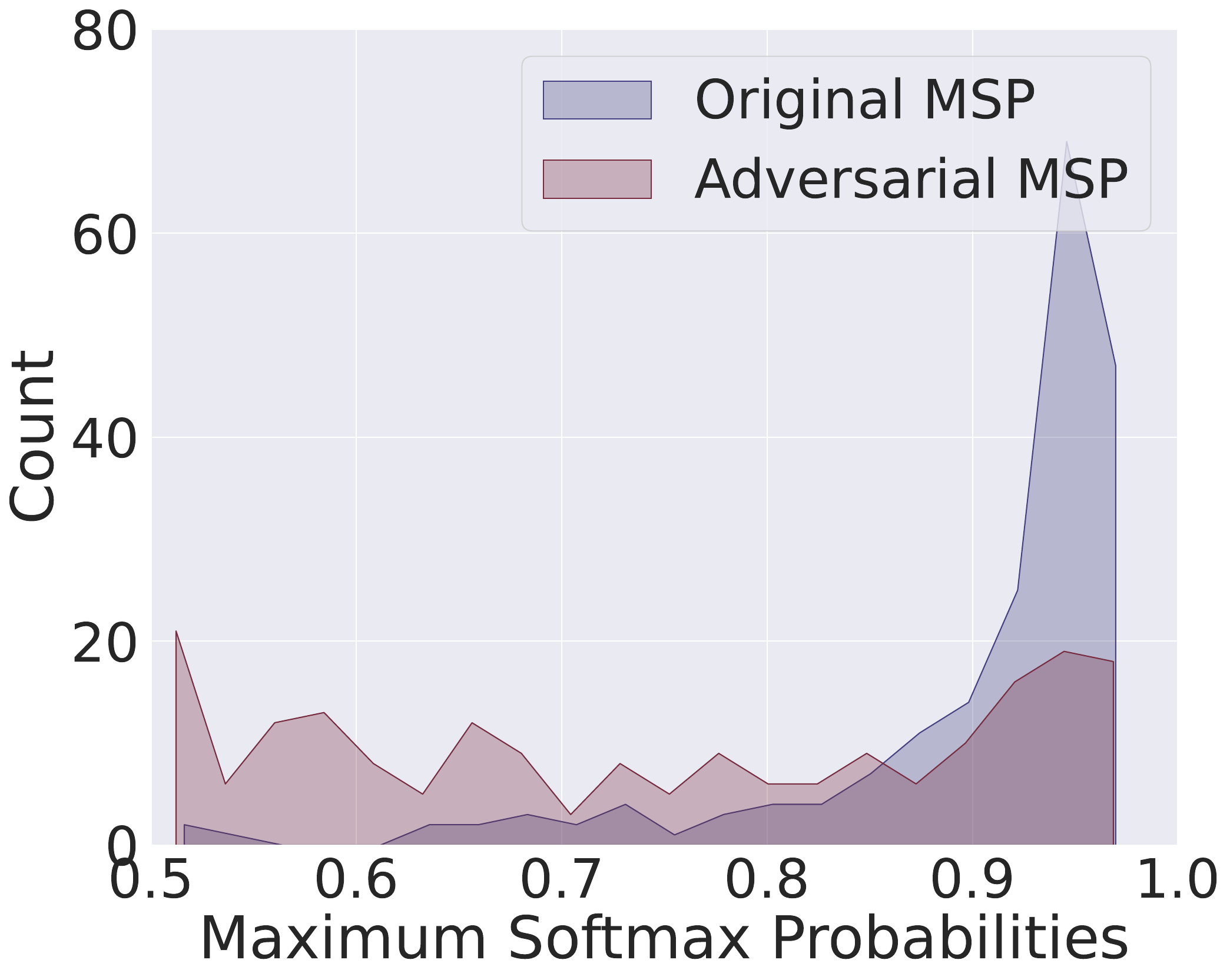}
         \caption{MSP on RTE dataset.}
     \end{subfigure}    
     \hfill
     \begin{subfigure}[t]{0.45\columnwidth}
         \centering
         \includegraphics[width=\textwidth]{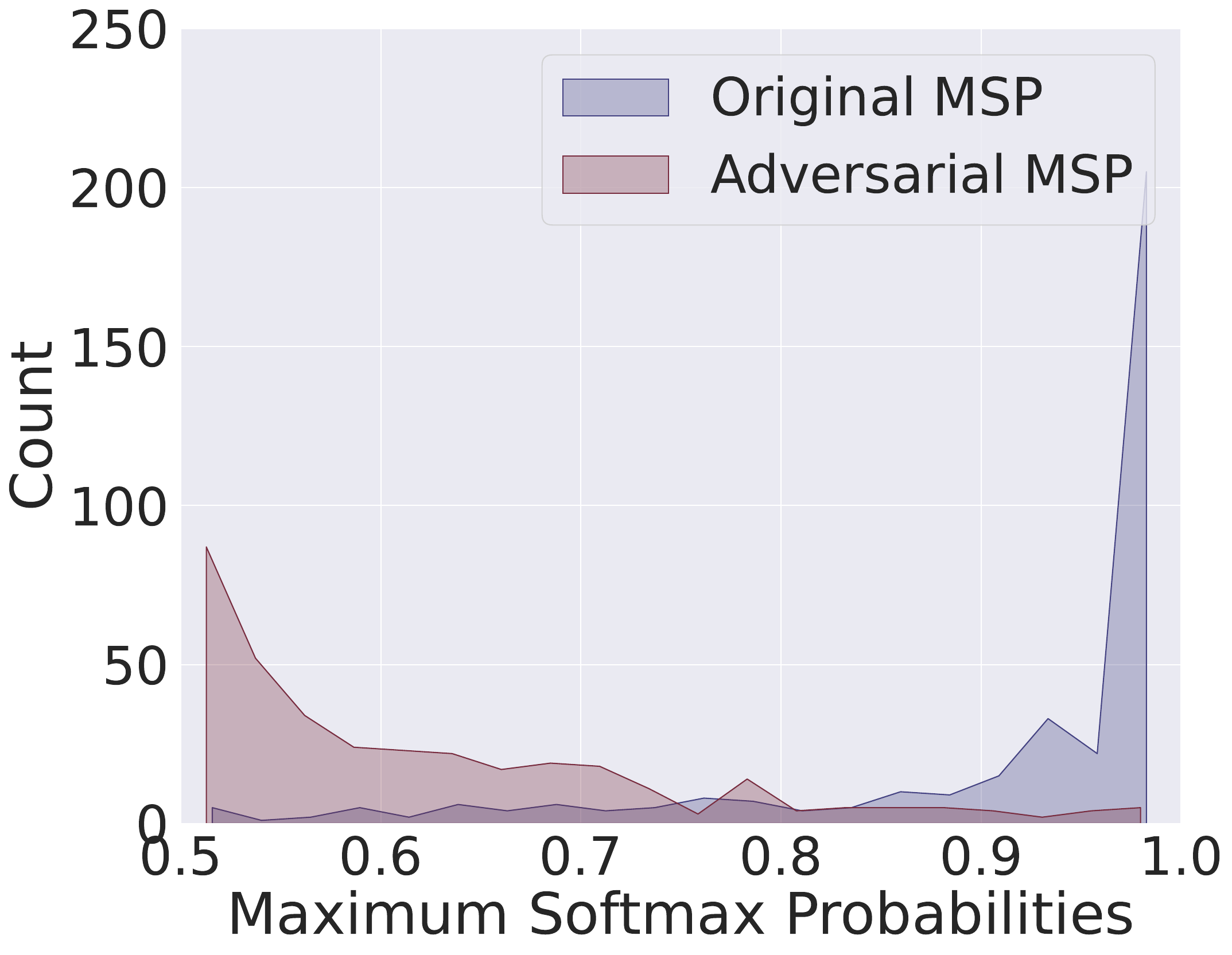}
         \caption{MSP on MRPC dataset.}
     \end{subfigure}   
    \caption{Visualization of the distribution shift between original data and adversarial data generated by BERT-Attack when attacking \textsc{BERT-base} regarding Maximum Softmax Probability.}
    \label{fig: bertattack_observe_MSP}
\end{figure*}

\begin{figure*}[t]
     \centering
     \begin{subfigure}[t]{0.45\columnwidth}
         \centering
         \includegraphics[width=\textwidth]{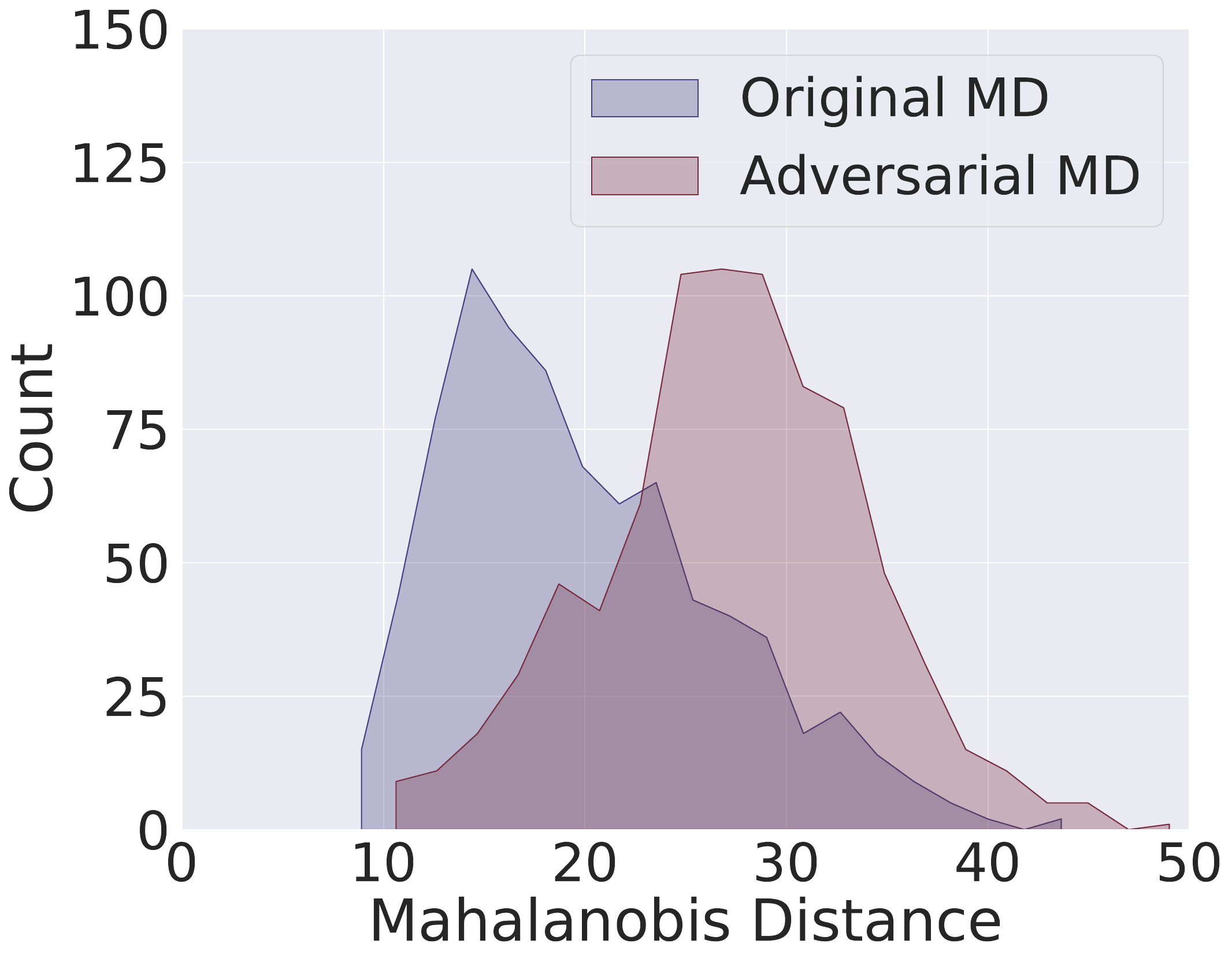}
         \caption{MD on SST-2 dataset.}
     \end{subfigure}
     \hfill
     \begin{subfigure}[t]{0.45\columnwidth}
         \centering
         \includegraphics[width=\textwidth]{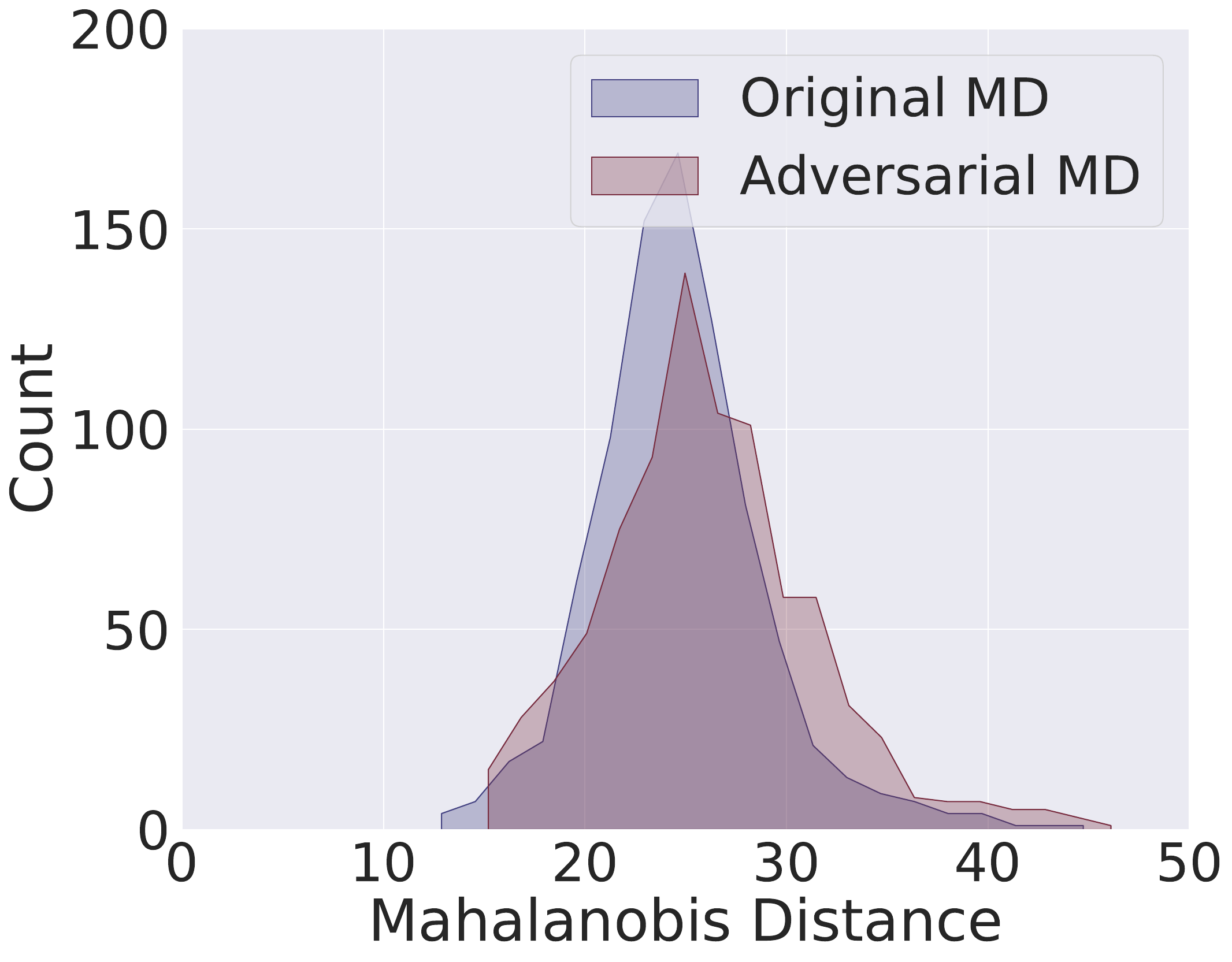}
         \caption{MD between on CoLA dataset.}
     \end{subfigure}
     \hfill
     \begin{subfigure}[t]{0.45\columnwidth}
         \centering
         \includegraphics[width=\textwidth]{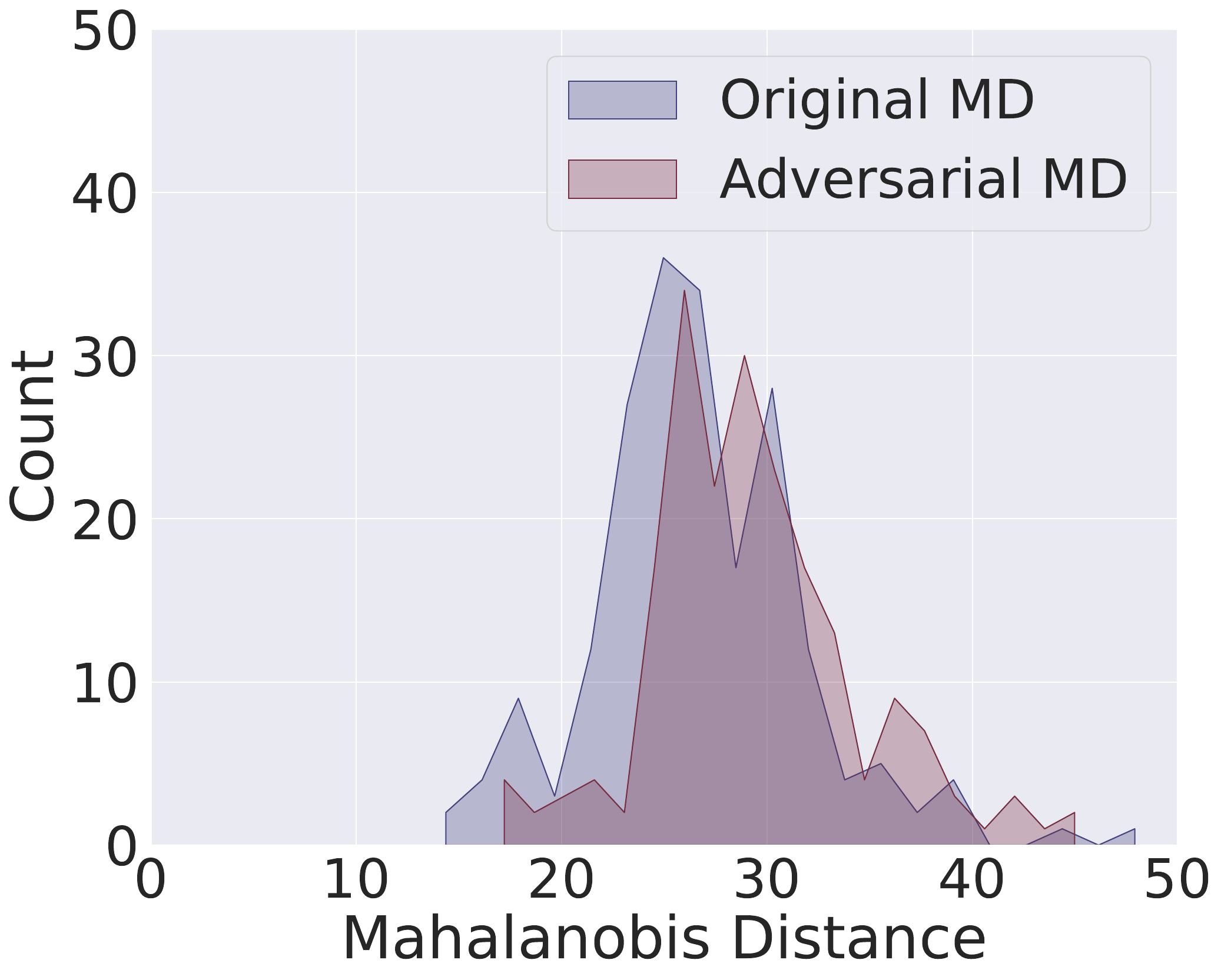}
         \caption{MD on RTE dataset.}
     \end{subfigure}
     \hfill
     \begin{subfigure}[t]{0.45\columnwidth}
         \centering
         \includegraphics[width=\textwidth]{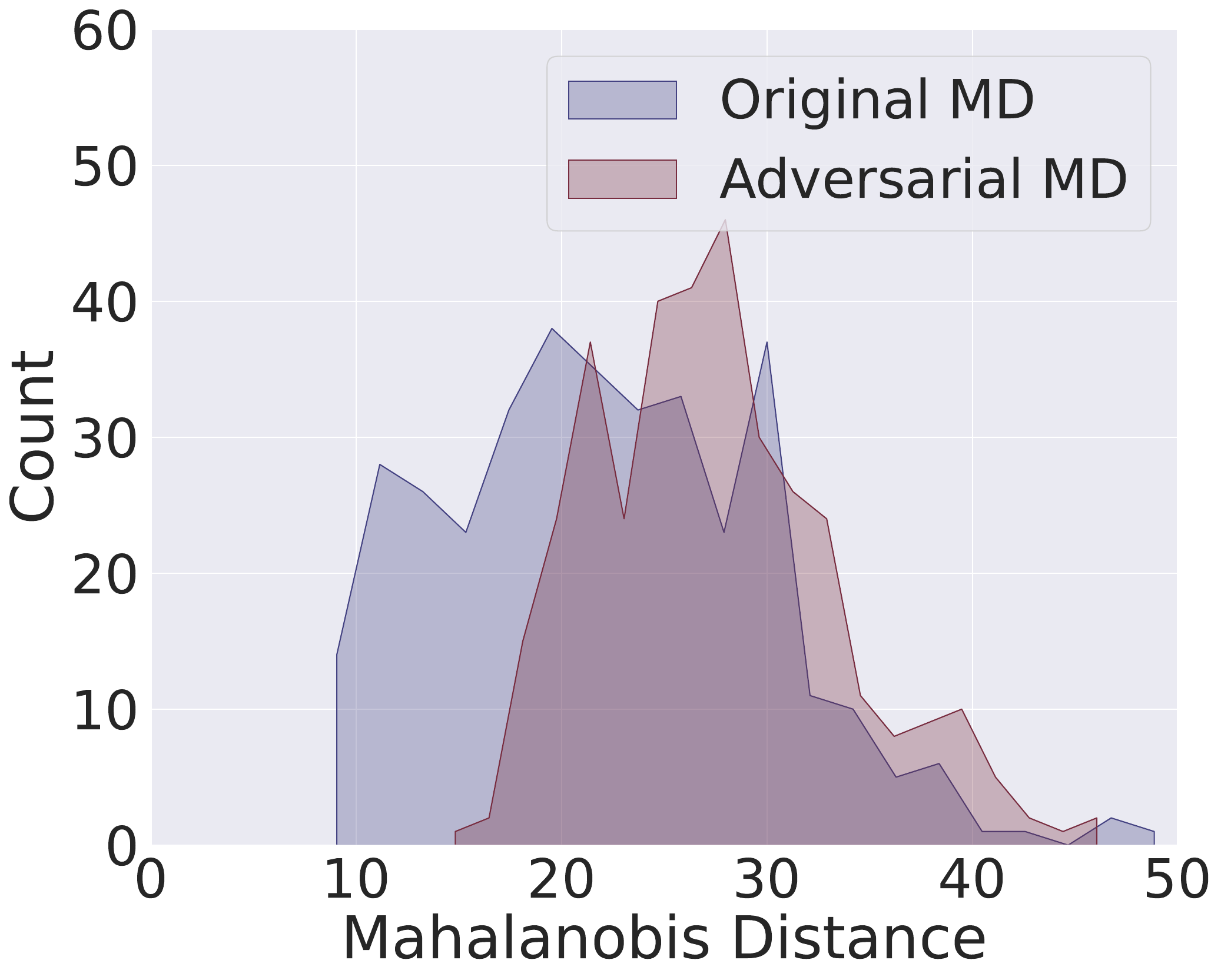}
         \caption{MD on MRPC dataset.}
     \end{subfigure}
    \caption{Visualization of the distribution shift between original data and adversarial data generated by BERT-Attack when attacking \textsc{BERT-base} regarding Mahalanobis Distance.}
    \label{fig: bertattack_observe_MD}
\end{figure*}

The observation experiments on previous attack methods TextFooler, TextBugger, DeepWordBug, and BERT-Attack are shown in Figure~\ref{fig: textfooler_observe_MSP}, Figure~\ref{fig: textfooler_observe_MD}, Figure~\ref{fig: textbugger_observe_MSP}, Figure~\ref{fig: textbugger_observe_MD}, Figure~\ref{fig: deepwordbug_observe_MSP}, Figure~\ref{fig: deepwordbug_observe_MD}, Figure~\ref{fig: bertattack_observe_MSP}, and Figure~\ref{fig: bertattack_observe_MD}.

The distribution shift between adversarial examples and original examples is more evident in terms of MSP across all the datasets. The distribution shift between adversarial examples and original examples in terms of MD is clear only on SST-2 dataset and MRPC dataset. Although this shift is not always present in terms of MD, it is imperative to address this issue given its presence in certain datasets.

\end{document}